\documentclass[letterpaper,10pt,journal,twoside]{IEEEtran}
\IEEEoverridecommandlockouts
\pdfoutput=1

\usepackage{cite}
\usepackage{amsmath,amssymb,amsfonts}
\usepackage{bm}
\usepackage{graphicx}
\usepackage{textcomp}
\usepackage{xcolor}
\usepackage{siunitx}
\usepackage{booktabs}
\usepackage{tabularx}
\usepackage{multirow}
\usepackage{array}
\usepackage{makecell}
\usepackage{algorithm}
\usepackage{algpseudocode}
\usepackage{amssymb}
\usepackage[caption=false,font=footnotesize,labelfont=rm,textfont=rm]{subfig}
\usepackage[hidelinks]{hyperref}

\algrenewcommand\algorithmicrequire{\textbf{Input:}}
\algrenewcommand\algorithmicensure{\textbf{Output:}}

\usepackage[font=footnotesize]{caption}
\def\BibTeX{{\rm B\kern-.05em{\sc i\kern-.025em b}\kern-.08em
    T\kern-.1667em\lower.7ex\hbox{E}\kern-.125emX}}

\begin{document}

\title{FATHOM: World-Model-Guided \\
Diffusion Action-Trajectory Selection \\
for Sparse-Marker AUV Tracking
\author{Xiaorui Wang\textsuperscript{$\dagger$}, Zeyu Sha\textsuperscript{$\dagger$}, Mingyang Yang, David Chen Wang and Feitian Zhang\textsuperscript{*}}

\thanks{\textsuperscript{$\dagger$} These authors contributed equally to this work.

The authors are with the Robotics and Control Laboratory, School of Advanced Manufacturing and Robotics, and the State Key Laboratory of Turbulence and Complex Systems,  Peking University, Beijing, 100871, China (email: \href{mailto:jnswxr@stu.pku.edu.cn}{jnswxr@stu.pku.edu.cn}; \href{mailto:schahzy@stu.pku.edu.cn}{schahzy@stu.pku.edu.cn}; \href{mailto:2201112083@stu.pku.edu.cn}{2201112083@stu.pku.edu.cn}; \href{mailto:6david035@gmail.com}{6david035@gmail.com}; \href{mailto:feitian@pku.edu.cn}{feitian@pku.edu.cn}).
}
}

\maketitle


\begin{abstract}
Reliable relative tracking is fundamental to cooperative autonomous underwater vehicles (AUVs), yet remains challenging under nonlinear vehicle dynamics and degraded underwater visual conditions. Active light markers improve robustness to appearance variation, but provide deliberately sparse observations that can become incomplete or corrupted by occlusion, reflections, and detection noise during dynamic motion. To address the resulting partial-observation control problem, this paper proposes \textbf{FATHOM}, a world-model-guided diffusion action-trajectory selection framework for sparse-marker AUV tracking. FATHOM first integrates structured marker observations, inertial measurements, action history, and temporal dynamics into a \textbf{predictive tracking belief} using a multimodal task-supervised recurrent state-space model. A diffusion action prior then generates multiple expert-like multi-thruster trajectories, whose future tracking consequences are predicted through the learned latent dynamics. Rather than directly executing a generated trajectory, FATHOM selects the most favorable candidate under an explicit receding-horizon objective and continually updates the decision from onboard observations. Extensive offline analyses verify the learned predictive belief, multi-step world-model prediction, diffusion trajectory generation, and predictive selection. More importantly, comprehensive real-AUV experiments, including strong pose-based comparisons and component ablations, demonstrate robust autonomous tracking across nominal and challenging motions with naturally occurring partial marker visibility, while maintaining real-time onboard operation without motion-capture feedback.
\end{abstract}


\section{Introduction}
\IEEEPARstart{A}{utonomous} underwater vehicles (AUVs) are increasingly moving from isolated operation toward cooperative autonomy, enabling persistent observation, distributed exploration, target following, inspection, and intervention over extended underwater environments \cite{paull2014auv,lin2017multiauv,chun2024moving,sha2025portable,weng2024link,sufan2025swim,cai2025event}. A fundamental capability in such systems is close-range relative tracking, where a follower must regulate its motion with respect to a moving leader. Unlike waypoint regulation against a static reference, relative tracking continuously couples perception, target motion, nonlinear vehicle dynamics, and multi-axis actuation. Underwater operation further complicates this loop through hydrodynamic coupling, external disturbances, and strongly variable visual conditions.

Optical sensing is attractive for close-range tracking because of its high spatial resolution and low latency. Its reliability, however, is affected by wavelength-dependent attenuation, scattering, turbidity, nonuniform illumination, and reflections \cite{kim2013visualslam,hover2012advanced,ancuti2018color,akkaynak2019seathru,xu2025aquaslam}. Active light markers provide a purposeful alternative to relying solely on natural appearance. Their controlled spectral and geometric structure produces distinctive observations and has been successfully used for underwater and robotic relative localization \cite{bosch2016beacons,ebmer2024led,sha2026amrpose}. The resulting sensing is robust but deliberately sparse. During dynamic relative motion, only a subset of markers may remain visible, while image noise, reflections, occlusion, and spurious detections further perturb individual observations. The controller therefore cannot always infer the interaction state reliably from the current frame alone.

\begin{figure*}[t]
    \centering
    \includegraphics[width=0.98\linewidth]{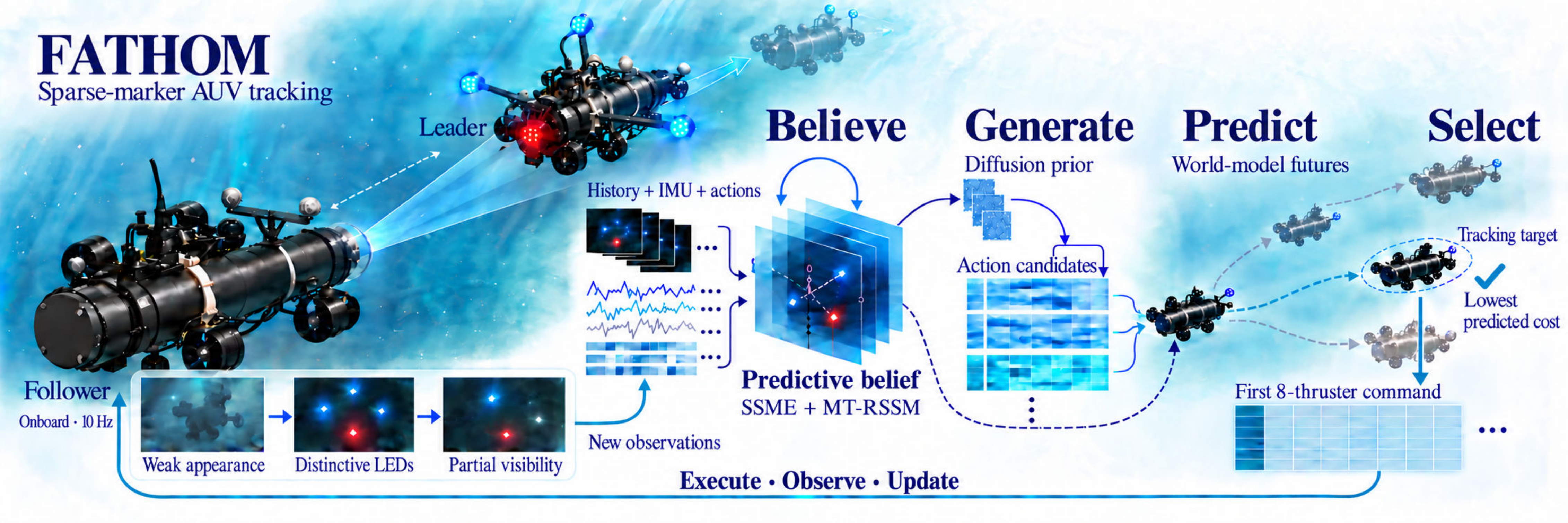}
    \caption{Conceptual overview of FATHOM for sparse-marker AUV tracking.
    Active LEDs provide distinctive but potentially incomplete observations.
    Visual--inertial histories and past actions form a predictive tracking belief.
    A diffusion prior generates candidate thruster trajectories, which the world model evaluates through their predicted tracking consequences.
    The lowest-cost candidate supplies the first command before the next observation updates the belief.}
    \label{fig:introduction}
\end{figure*}

A common solution is to reconstruct an explicit six-degree-of-freedom (6-DoF) relative pose and feed it to a conventional or learned controller. Modern active-marker estimators provide accurate and temporally consistent pose estimates even under partial observations \cite{bosch2016beacons,ebmer2024led,sha2026amrpose}. Yet pose estimation and feedback decision making are not identical problems. For dynamic tracking, information useful for predicting the next control consequence is distributed across recent marker observations, vehicle motion, and previously executed actions. Compressing this history immediately into a single geometric estimate risks discarding task-relevant temporal and dynamic information. This motivates maintaining a compact representation of how the relative interaction evolves rather than estimating only the leader's current pose.

Latent world models provide a natural basis for such reasoning. Recurrent state-space models summarize incomplete observation histories into latent states and predict their evolution under candidate actions \cite{hafner2019planet,hafner2020dreamer,hafner2025dreamerv3,hansen2022tdmpc,hansen2024tdmpc2,zhou2025dinowm,prasad2025latentdynamics}. This capability is particularly relevant to sparse underwater sensing because temporal memory and action-conditioned prediction address different aspects of the same partial-observation problem. A useful control representation should therefore not only explain the current tracking condition, but also support short-horizon prediction under hypothetical thruster commands.

Prediction alone, however, does not determine which commands should be considered. Direct regression to a single action sequence is restrictive when different multi-thruster behaviors are valid for the same tracking condition. Diffusion models provide an expressive way to represent high-dimensional and multimodal action distributions \cite{ho2020ddpm,song2021ddim,chi2024diffusionpolicy,ze2024dp3,wang2025hdp}. Recent work has extended this idea from direct imitation toward trajectory generation and model-guided planning \cite{carvalho2025mpd,power2024sampling,meng2026utracker,qi2026gpc}. This suggests a complementary view of generative control. Rather than committing immediately to one action, a learned policy generates multiple plausible trajectories whose predicted consequences are evaluated before execution.

These observations lead naturally to the framework proposed in this work. We introduce \textbf{FATHOM}, short for \textbf{F}eedback \textbf{A}ction \textbf{T}rajectory selection with \textbf{H}idden-state \textbf{O}utcome \textbf{M}odeling. The name also reflects the physical and conceptual setting of the problem. A \emph{fathom} is historically associated with measuring underwater depth, while to \emph{fathom} means to understand something that is not directly observable. In our setting, the controller must similarly infer a hidden tracking condition from sparse underwater observations and reason about the consequences of future actions.

FATHOM organizes closed-loop decision making as \emph{Believe--Generate--Predict--Select}, as illustrated in Fig.~\ref{fig:introduction}. A \textbf{Structured Sparse-Marker Encoder (SSME)} extracts a compact representation from unordered and partially missing LED observations. A \textbf{Multimodal Task-Supervised Recurrent State-Space Model (MT-RSSM)} then integrates this representation with inertial measurements, previous actions, and temporal dynamics to form a \textbf{predictive tracking belief}. Conditioned on this belief, a \textbf{Diffusion Action Prior (DAP)} generates multiple expert-like trajectories of eight-thruster commands. MT-RSSM predicts their future tracking consequences, and a \textbf{World-Model Predictive Selector (WMPS)} chooses the trajectory with the lowest receding-horizon control objective. Only its first command is executed before the belief is updated from the next onboard observation.

This design gives each learned component a distinct role. The recurrent world model captures what cannot be inferred reliably from a single sparse observation and predicts how the interaction may evolve. Diffusion represents multiple plausible control behaviors without collapsing them into a single conditional action. Predictive selection then converts this diversity into an online advantage by choosing among those behaviors according to their anticipated closed-loop consequences. The resulting controller operates from onboard sparse LED observations and inertial sensing, without requiring motion-capture feedback during deployment.

The main contributions of this paper are threefold. First, we develop a control-oriented representation for AUV tracking under deliberately sparse active visual sensing, where SSME and MT-RSSM integrate incomplete marker observations, inertial motion, action history, and temporal dynamics into a predictive tracking belief that supports both hidden-state inference and future prediction. Second, we propose FATHOM, a world-model-guided diffusion control framework that treats diffusion as an action-trajectory prior rather than the final decision maker. Multiple expert-like thruster trajectories are evaluated through learned latent dynamics, and a finite-error analysis relates world-model prediction accuracy to the quality of predictive trajectory selection. Third, we demonstrate fully onboard sparse visual--inertial tracking on real AUVs and systematically evaluate FATHOM against privileged and pose-based controllers, showing the benefits of predictive belief and trajectory selection under challenging motion and naturally occurring partial marker visibility while retaining real-time execution.

\section{Related Work}

\subsection{Vision-Based Relative Perception and Tracking for AUVs}

Underwater visual feedback is degraded by absorption, scattering, turbidity, illumination variation, and reflections \cite{kim2013visualslam,hover2012advanced,ancuti2018color,akkaynak2019seathru}. Image restoration and multimodal sensing address complementary aspects of these challenges \cite{ancuti2018color,akkaynak2019seathru,chun2024moving,yang2025leader,xu2025aquaslam}. In particular, AQUA-SLAM fuses acoustic, visual, and inertial measurements for robust underwater localization \cite{xu2025aquaslam}, while inertial-aided visual servoing incorporates vehicle dynamics into underwater image-based control \cite{krupinski2017visualservo}. For close-range cooperation, active light markers provide controlled geometric cues for relative localization \cite{bosch2016beacons,ebmer2024led}. AMR-Pose further combines probabilistic association, Lie-group propagation, and visibility-adaptive fusion for 6-DoF estimation from sparse and corrupted LED detections \cite{sha2026amrpose}. It provides a strong geometric perception reference, although accurate pose reconstruction and predictive control remain distinct objectives.

Underwater tracking research spans acoustic coordination, vision-based following, optical-link alignment, energy-aware control, and leader--follower prediction \cite{lin2017multiauv,sha2025portable,weng2024link,sufan2025swim,cai2025event}. Learning-based approaches include visual reinforcement learning \cite{norenko2026follower} and UTracker, which distills diffusion-based visuomotor control from simulated expert behavior \cite{meng2026utracker}. Beyond underwater applications, relative localization from partial measurements and belief-based control under occlusion have been studied in multirobot systems and robotic manipulation \cite{chen2025relativelocalizability,xu2025occlusioncontrol,zhong2025rumi}. FATHOM instead maintains a control-oriented belief from sparse visual--inertial history and previous actions, using it for both trajectory generation and prediction rather than relying solely on explicit pose feedback or direct visuomotor action prediction.

\subsection{Latent World Models for Partially Observed Robot Control}

Latent world models support action-conditioned prediction and planning under incomplete observations \cite{ha2018worldmodels,hafner2019planet}. RSSM-based approaches, including PlaNet, Plan2Explore, and Dreamer, learn recurrent dynamics for planning, exploration, and policy learning \cite{hafner2019planet,hafner2020plan2explore,hafner2020dreamer,hafner2021dreamerv2,hafner2025dreamerv3}. Related methods combine learned dynamics with search and predictive control, including MuZero, TD-MPC, and TD-MPC2 \cite{schrittwieser2020muzero,hansen2022tdmpc,hansen2024tdmpc2}. Learning-based dynamics control has also been demonstrated on physical robots \cite{wu2023daydreamer,saviolo2024active,duong2024porthamiltonian}. These approaches emphasize learning compact dynamics that preserve information needed to evaluate future actions.

Recent research also emphasizes task-relevant latent and feature-space prediction. DINO-WM, perceptive forward dynamics, and multimodal latent-dynamics models represent complementary approaches to predicting motion and interaction \cite{zhou2025dinowm,roth2025fdm,prasad2025latentdynamics}. Perception-aware predictive controllers also address incomplete geometric information \cite{grandia2023perceptive}. FLARE and LaDi-WM use predicted features for policy learning or refinement \cite{zheng2025flare,huang2025ladiwm}, while action-conditioned generative models extend prediction to planning and evaluation \cite{bar2025nwm,zhu2025irasim,quevedo2025evaluating}. Their sensing representations and downstream objectives differ from those of sparse-marker feedback control. FATHOM builds on recurrent latent prediction but addresses a distinct sensing-to-control interface. Its \textbf{Multimodal Task-Supervised RSSM (MT-RSSM)} integrates incomplete marker observations, IMU measurements, and action history into a \textbf{predictive tracking belief}. Unlike a standalone state estimator or Dreamer-style actor--critic, the same belief supports both posterior inference and prior prediction for online action selection.

\subsection{Diffusion Policies and World-Model-Guided Robot Control}

Deterministic imitation tends to average distinct valid actions, motivating expressive policy distributions such as energy-based imitation and Behavior Transformers \cite{florence2022ibc,shafiullah2022behavior}. Diffusion and score-based models provide multimodal action and trajectory generation \cite{ho2020ddpm,song2021score,song2021ddim,janner2022diffuser,ajay2023decisiondiffuser}. In robotics, goal-conditioned score policies, Diffusion Policy, and subsequent geometric or hierarchical variants model temporally coordinated behavior \cite{reuss2023goal,chi2024diffusionpolicy,ze2024dp3,yang2024equibot,ke2025diffuseractor,wang2025hdp}. Other advances focus on efficient action generation through consistency models and action chunking \cite{prasad2024consistency,zhao2023act}. These policies represent multiple plausible behaviors, but their samples need not minimize a task-specific predictive objective.

More closely related methods combine learned trajectory priors with predictive objectives. Motion Planning Diffusion uses a diffusion prior with optimization-based planning \cite{carvalho2025mpd}, while learned sampling distributions improve sampling-based model predictive control \cite{power2024sampling}. B-COD conditions diffusion planning on localization belief for marine navigation, and UTracker applies diffusion imitation to underwater tracking \cite{puthumanaillam2025bcod,meng2026utracker}. DiWA, FLARE, and LaDi-WM connect generative policies with learned dynamics or predicted latents \cite{chandra2025diwa,zheng2025flare,huang2025ladiwm}, while Navigation World Models and IRASim support action-conditioned planning and evaluation \cite{bar2025nwm,zhu2025irasim}. GPC is particularly relevant because it ranks or refines generative-policy candidates using world-model predictions at inference time \cite{qi2026gpc}. Actor--critic MPC further illustrates the integration of learned policies and online predictive optimization in real robot control \cite{romero2026actorcriticmpc}.

FATHOM shares this model-guided generative-control principle but addresses tracking under sparse and intermittent observations. Its recurrent world model first forms a tracking belief from visual--inertial and action histories, then predicts the effects of direct eight-thruster trajectories. DAP generates expert-like candidates, and WMPS selects among them using an explicit receding-horizon objective. This connects partial-observation belief inference with predictive control through the same learned dynamics.

\section{System Setup and Problem Formulation}
\label{sec:problem}

\subsection{Dual-AUV Experimental Platform}
\label{subsec:platform}

\begin{figure}[t]
    \centering
    \includegraphics[width=0.98\columnwidth]{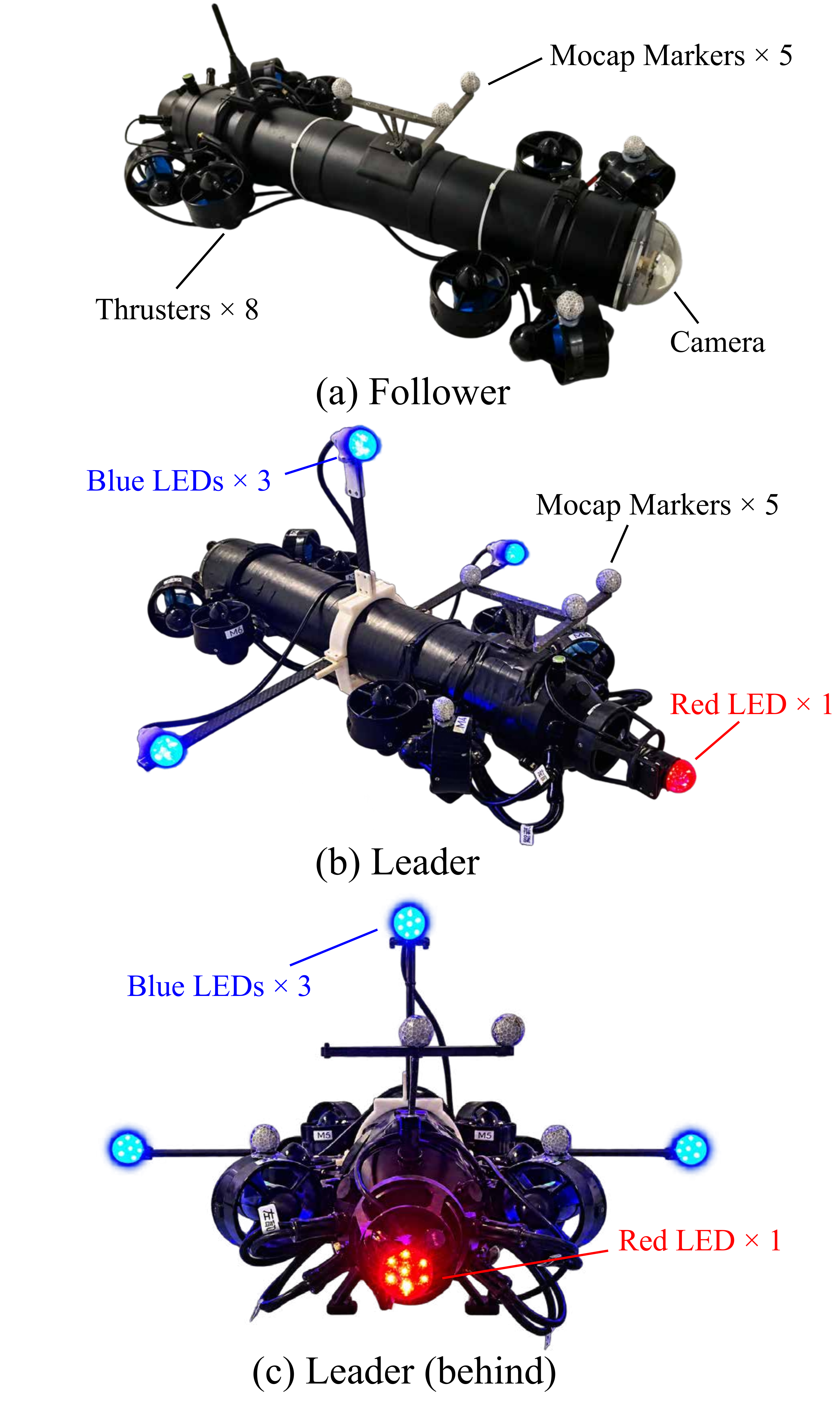}
    \caption{Dual-AUV hardware. (a) Follower with eight thrusters, a monocular camera, and five motion-capture markers. (b) Leader with a four-LED array and five motion-capture markers. (c) Rear view showing one red reference LED and three blue LEDs.}
    \label{fig:auv_hardware}
\end{figure}

Two OpenAUV vehicles form the leader--follower platform shown in Fig.~\ref{fig:auv_hardware}. The follower measures $920\times360\times305~\mathrm{mm}$, and the leader has comparable hull dimensions. Both vehicles are equipped with eight thrusters for multi-axis motion control. The follower carries a forward-looking monocular camera and an inertial measurement unit (IMU) providing attitude and angular-rate measurements. Its onboard electronics include a Pixhawk flight controller, a Raspberry Pi, and an NVIDIA Jetson AGX Orin with 64~GB of memory. The Jetson runs visual perception, world-model inference, diffusion sampling, and predictive selection, and FATHOM outputs commands to the eight follower thrusters.

The leader carries a rigid frame supporting the active LED array described below. Five motion-capture markers are mounted on each AUV for ground-truth measurements. Motion capture provides privileged feedback for expert-data collection and the reference controller, as well as labels for training and evaluation. It is never supplied to FATHOM during autonomous tracking.

\subsection{Sparse Active-Marker Sensing}
\label{subsec:marker_sensing}

The leader-mounted array follows the compact active-marker design in~\cite{sha2026amrpose}. A red reference LED and three blue LEDs are arranged on a rigid support, with the blue markers positioned to the left, above, and right of the red marker, as shown in Fig.~\ref{fig:auv_hardware}(b)--(c). Their distinct colors and calibrated spatial arrangement provide sparse geometric cues for relative perception. Each LED module integrates an emitter, driver circuit, diffusive lampshade, and pressure-resistant housing sealed with waterproof potting material. The modules support a 3--16~V supply and microcontroller-based pulse-width-modulation (PWM) brightness adjustment, allowing the illumination to be adapted to different observation conditions.

The follower extracts candidate LEDs from undistorted monocular images using HSV segmentation, morphological filtering, and connected-component analysis. The resulting observation is an unordered set
\begin{equation}
\mathcal{D}_{t}
=\left\{(\bm{z}_{m,t},c_{m,t})\right\}_{m=1}^{M_t},
\quad
\bm{z}_{m,t}=[u_{m,t},v_{m,t}]^{\top},
\label{eq:led_detection_set}
\end{equation}
where $c_{m,t}\in\{\mathrm{red},\mathrm{blue}\}$ and $M_t$ varies over time. During relative motion, occlusion, field-of-view changes, reflections, and spurious detections may remove or corrupt individual markers. FATHOM directly processes variable-cardinality detections without requiring complete visibility or frame-wise pose estimation.

\subsection{AUV Dynamics and Control Interface}
\label{subsec:auv_dynamics}

Let $\mathcal{I}$ denote the inertial frame and $\mathcal{B}_L$, $\mathcal{B}_F$ the leader and follower body frames. Both vehicles follow standard 6-DoF marine dynamics~\cite{fossen2011handbook}, with body-fixed velocity $\bm{\nu}_{i}=[\bm{v}_{i}^{\top},\bm{\omega}_{i}^{\top}]^{\top}$, pose $\bm{\eta}_{i}$, and inertia, Coriolis, damping, restoring, actuation, and disturbance terms for $i\in\{L,F\}$. For the follower, the control interface is
\begin{equation}
\bm{\tau}_{F,t}=\bm{B}_{F}\bm{f}_{t},\quad
\bm{f}_{t}=\mathcal{T}(\bm{a}_{t}),\quad
\bm{a}_{t}\in\mathcal{A}\subset\mathbb{R}^{8},
\label{eq:follower_control_interface}
\end{equation}
where $\bm{B}_{F}$ is the thruster allocation matrix and $\mathcal{T}(\cdot)$ maps normalized commands to thrust. FATHOM directly predicts $\bm{a}_{t}$ and does not require an identified hydrodynamic model for online selection.

Let ${}^{I}\bm{T}_{L},{}^{I}\bm{T}_{F}\in\mathrm{SE}(3)$ denote the vehicle poses. The leader pose in the follower frame is
\begin{equation}
{}^{F}\bm{T}_{L,t}
=({}^{I}\bm{T}_{F,t})^{-1}{}^{I}\bm{T}_{L,t}
=\begin{bmatrix}
\bm{R}_{FL,t}&\bm{p}_{FL,t}\\
\bm{0}^{\top}&1
\end{bmatrix}.
\label{eq:relative_pose}
\end{equation}
Its relative position evolves according to
\begin{equation}
\dot{\bm{p}}_{FL}
=-[\bm{\omega}_{F}]_{\times}\bm{p}_{FL}
-\bm{v}_{F}+\bm{R}_{FL}\bm{v}_{L}.
\label{eq:relative_position_dynamics}
\end{equation}
Thus, the tracking condition depends on the unmeasured leader motion as well as follower actuation and underwater disturbances. This coupling motivates temporal, action-conditioned state inference from onboard measurements.

\subsection{Sparse-Marker AUV Tracking Problem}
\label{subsec:tracking_problem}

The follower regulates the leader's relative position and yaw while stabilizing its own roll and pitch. The six-dimensional control-relevant state is
\begin{equation}
\bm{y}_{t}
=\begin{bmatrix}
\bm{p}_{FL,t}^{\top}&\phi_{F,t}&\theta_{F,t}&\psi_{FL,t}
\end{bmatrix}^{\top}\in\mathbb{R}^{6},
\label{eq:relative_tracking_state}
\end{equation}
where $\psi_{FL,t}$ is the relative yaw and $(\phi_{F,t},\theta_{F,t})$ are the follower's absolute roll and pitch. With a desired relative position $\bm{p}^{*}=[1.5,0,0]^{\top}\,\mathrm{m}$ and $\psi^{*}=0$, the target is
\begin{equation}
\bm{y}^{*}
=\begin{bmatrix}
{\bm{p}^{*}}^{\top}&0&0&\psi^{*}
\end{bmatrix}^{\top},
\label{eq:desired_relative_state}
\end{equation}
with tracking error
\begin{equation}
\bm{\varepsilon}_{t}
=\begin{bmatrix}
\bm{p}_{FL,t}-\bm{p}^{*}\\
\phi_{F,t}\\
\theta_{F,t}\\
\operatorname{wrap}(\psi_{FL,t}-\psi^{*})
\end{bmatrix}.
\label{eq:tracking_error}
\end{equation}
This separates relative heading regulation from follower attitude stabilization and supports a stable camera viewpoint.

The onboard observation comprises the detected LED set and follower IMU measurements,
\begin{equation}
\bm{o}_{t}=(\mathcal{D}_{t},\bm{s}^{\mathrm{imu}}_{t}),\quad
\bm{s}^{\mathrm{imu}}_{t}
=\begin{bmatrix}
\phi_{F,t}&\theta_{F,t}&p_{F,t}&q_{F,t}&r_{F,t}
\end{bmatrix}^{\top},
\label{eq:fathom_observation}
\end{equation}
where $(p_{F,t},q_{F,t},r_{F,t})$ are body angular rates. Since the current observation does not fully determine the tracking state, the controller uses the observation--action history $\mathcal{H}_{t}=(\bm{o}_{1},\bm{a}_{1},\ldots,\bm{o}_{t-1},\bm{a}_{t-1},\bm{o}_{t})$ and seeks to minimize
\begin{equation}
\min_{\pi}\;\mathbb{E}_{\pi}\!\left[
\sum_{t=0}^{T}
\ell\!\left(\bm{\varepsilon}_{t},\bm{a}_{t},
\bm{a}_{t}-\bm{a}_{t-1}\right)
\right],
\label{eq:tracking_control_objective}
\end{equation}
where $\bm{a}_{t}\sim\pi(\cdot\mid\mathcal{H}_{t})$ and $\ell(\cdot)$ penalizes tracking error, actuation effort, and command variation.

Rather than storing the complete history, FATHOM learns a compact predictive tracking belief,
\begin{equation}
\bm{c}_{t}=f_{\theta}(\mathcal{H}_{t}),
\label{eq:predictive_belief_definition}
\end{equation}
which should retain the information needed to evaluate a future thruster sequence $\bm{A}_{t}=(\bm{a}_{t},\ldots,\bm{a}_{t+H-1})$. Its desired action-conditioned predictive property is
\begin{equation}
\begin{aligned}
&p\!\left(\bm{y}_{t+1:t+H}\mid\mathcal{H}_{t},\bm{A}_{t}\right)\\
&\qquad\approx p_{\theta}\!\left(
\bm{y}_{t+1:t+H}\mid\bm{c}_{t},\bm{A}_{t}\right).
\end{aligned}
\label{eq:predictive_sufficiency}
\end{equation}
This control-oriented belief need not reconstruct the full physical state, but must support prediction under candidate actions (Section~\ref{sec:fathom}).

\section{FATHOM}
\label{sec:fathom}

\subsection{Framework Overview}
\label{subsec:fathom_overview}

\begin{figure*}[t]
    \centering
    \includegraphics[width=0.98\linewidth]{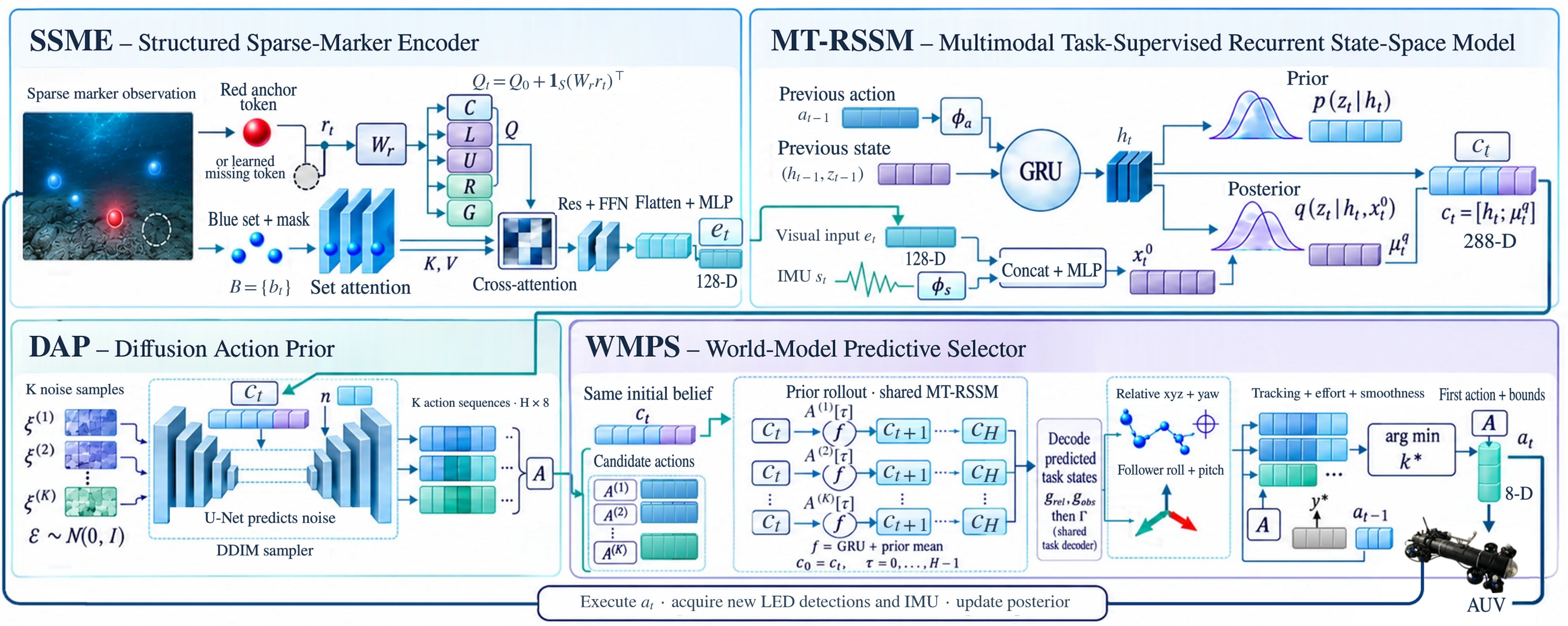}
    \caption{Overview of FATHOM. SSME encodes sparse LED observations. MT-RSSM combines visual features, IMU measurements, and action history to infer a tracking belief. DAP generates belief-conditioned action sequences. WMPS evaluates their predicted task states using tracking, effort, and smoothness costs, then executes the bounded first action of the lowest-cost sequence. New observations update the belief to close the loop.}
    \label{fig:fathom_framework}
\end{figure*}

FATHOM realizes the above formulation through a closed-loop
\emph{Believe--Generate--Predict--Select} process, as illustrated
in Fig.~\ref{fig:fathom_framework}. SSME converts sparse and
partially missing LED observations into a structured visual
representation. MT-RSSM integrates this representation with
IMU measurements, previous actions, and temporal dynamics
to update the predictive tracking belief $\bm{c}_{t}$.
DAP then generates $K$ expert-like multi-thruster trajectories.
Their future tracking consequences are predicted through the
MT-RSSM latent prior and ranked by WMPS under an explicit
receding-horizon objective. Only the first command is executed
before the next onboard observation corrects the belief.

The world model supports both \emph{state inference} and \emph{counterfactual prediction}. Diffusion generates plausible alternatives, while WMPS selects actions by predicted tracking performance rather than generation likelihood, without requiring online frame-wise pose estimation.

\subsection{Structured Sparse-Marker Encoder}
\label{subsec:ssme}

Sparse observations $\mathcal{D}_{t}$ are unordered and variable in cardinality, making coordinate concatenation sensitive to ordering and missing detections. SSME exploits asymmetric red--blue geometry while preserving permutation invariance among blue detections.

Let $m^{r}_{t}\in\{0,1\}$ indicate red-marker visibility and
$\bar{\bm{z}}^{r}_{t}
=[\bar{u}^{r}_{t},\bar{v}^{r}_{t}]^{\top}$
its normalized coordinate. The red-anchor token is
\begin{equation}
\bm{r}_{t}
=
\begin{cases}
\phi_r
\left(
[(\bar{\bm{z}}^{r}_{t})^{\top},m^{r}_{t}]^{\top}
\right)
+\bm{\tau}_{r},
& m^{r}_{t}=1,\\[1mm]
\bm{r}_{\mathrm{miss}}+\bm{\tau}_{r},
& m^{r}_{t}=0,
\end{cases}
\label{eq:ssme_red}
\end{equation}
where $\bm{\tau}_{r}$ is a marker-type embedding and
$\bm{r}_{\mathrm{miss}}$ is a learnable missing-marker token. Explicitly representing the missing-anchor case prevents temporary red-marker loss from collapsing the visual representation.

For blue detection $j$, the input feature and token are
\begin{align}
\bm{b}^{\mathrm{in}}_{j,t}
&=
[\bar{u}^{b}_{j,t},\bar{v}^{b}_{j,t},
\Delta\bar{u}_{j,t},\Delta\bar{v}_{j,t},m^{r}_{t}]^{\top},
\label{eq:ssme_blue_input}\\
\bm{b}_{j,t}
&=
\phi_b(\bm{b}^{\mathrm{in}}_{j,t})+\bm{\tau}_{b},
\label{eq:ssme_blue_embedding}
\end{align}
where $(\Delta\bar{u}_{j,t},\Delta\bar{v}_{j,t})$ are offsets from the red anchor when available. Invalid padded entries are removed by an attention mask.

The valid blue tokens are contextualized as an unordered set using a Transformer encoder \cite{vaswani2017attention},
\begin{equation}
\bm{B}_{t}
=
\operatorname{TransEnc}
\left(
\{\bm{b}_{j,t}\}_{j=1}^{M_t^b}
\right),
\label{eq:ssme_blue_transformer}
\end{equation}
where $M_t^b$ is the number of valid blue detections. A zero-valued dummy token is used when $M_t^b=0$ to avoid empty attention inputs. This avoids imposing a hard correspondence on partially visible blue markers.

SSME uses five learnable role-inspired queries,
\begin{equation}
\bm{Q}_{0}
=
[\bm{q}_{C},\bm{q}_{L},\bm{q}_{U},
\bm{q}_{R},\bm{q}_{G}]^{\top},
\label{eq:ssme_queries}
\end{equation}
for center, left, upper, right, and global context. These structural slots are not explicitly supervised marker identities. The red anchor conditions the queries,
\begin{equation}
\bm{Q}_{t}
=
\bm{Q}_{0}
+\bm{1}_{5}(\bm{W}_{r}\bm{r}_{t})^{\top},
\label{eq:ssme_query_condition}
\end{equation}
The red token and encoded blue features form the
cross-attention memory,
\begin{equation}
\bm{M}_{t}
=
\begin{bmatrix}
\bm{B}_{t}\\
\bm{r}_{t}^{\top}
\end{bmatrix}.
\label{eq:ssme_memory}
\end{equation}
Cross-attention then aggregates the marker representations,
\begin{equation}
\widetilde{\bm{Q}}_{t}
=
\operatorname{FFN}
\left[
\operatorname{MHA}(\bm{Q}_{t},\bm{M}_{t},\bm{M}_{t})
+\bm{Q}_{t}
\right].
\label{eq:ssme_cross_attention}
\end{equation}
The resulting sparse-marker representation is
\begin{equation}
\bm{e}_{t}
=
\phi_e
\left(
\operatorname{vec}(\widetilde{\bm{Q}}_{t})
\right)
\in\mathbb{R}^{128}.
\label{eq:ssme_output}
\end{equation}
SSME thus preserves geometry and visibility cues without requiring all four markers.

\subsection{Multimodal Task-Supervised RSSM}
\label{subsec:mtrssm}

SSME summarizes the current marker geometry but cannot recover the hidden coupled dynamics from one frame. MT-RSSM therefore combines sparse visual features, inertial sensing, previous actuation, and recurrent dynamics within an RSSM formulation \cite{hafner2019planet}.

The IMU and action embeddings are $\bm{u}^{s}_{t}=\phi_s(\bm{s}^{\mathrm{imu}}_{t})\in\mathbb{R}^{64}$ and $\bm{u}^{a}_{t}=\phi_a(\bm{a}_{t})\in\mathbb{R}^{64}$. Visual--inertial fusion gives
\begin{equation}
\bm{x}^{o}_{t}
=
\phi_o
\left(
\begin{bmatrix}
\bm{e}_{t}\\
\bm{u}^{s}_{t}
\end{bmatrix}
\right)
\in\mathbb{R}^{128}.
\label{eq:mtrssm_obsfusion}
\end{equation}

MT-RSSM maintains a deterministic state
$\bm{h}_{t}\in\mathbb{R}^{256}$
and a stochastic state
$\bm{z}_{t}\in\mathbb{R}^{32}$.
The action-conditioned recurrent transition is
\begin{equation}
\bm{h}_{t}
=
\operatorname{GRU}
\left(
\bm{h}_{t-1},
\begin{bmatrix}
\bm{z}_{t-1}\\
\bm{u}^{a}_{t-1}
\end{bmatrix}
\right).
\label{eq:mtrssm_transition}
\end{equation}
Before observing the current frame, the latent prior is
\begin{equation}
p_{\theta_W}(\bm{z}_{t}\mid\bm{h}_{t})
=
\mathcal{N}
\left(
\bm{\mu}^{p}_{t},
\operatorname{diag}[(\bm{\sigma}^{p}_{t})^{2}]
\right),
\label{eq:mtrssm_prior}
\end{equation}
with
$(\bm{\mu}^{p}_{t},\bm{\sigma}^{p}_{t})
=f_p(\bm{h}_{t})$.
After the sparse visual--inertial observation arrives, the posterior becomes
\begin{equation}
q_{\theta_W}(\bm{z}_{t}\mid\bm{h}_{t},\bm{x}^{o}_{t})
=
\mathcal{N}
\left(
\bm{\mu}^{q}_{t},
\operatorname{diag}[(\bm{\sigma}^{q}_{t})^{2}]
\right),
\label{eq:mtrssm_posterior}
\end{equation}
where
$(\bm{\mu}^{q}_{t},\bm{\sigma}^{q}_{t})
=f_q([\bm{h}_{t}^{\top},{\bm{x}^{o}_{t}}^{\top}]^{\top})$.

Online control uses the posterior mean,
\begin{equation}
\bm{z}_{t}=\bm{\mu}^{q}_{t},
\qquad
\bm{c}_{t}
=
\begin{bmatrix}
\bm{h}_{t}\\
\bm{z}_{t}
\end{bmatrix}
\in\mathbb{R}^{288},
\label{eq:predictive_tracking_belief}
\end{equation}
which instantiates the predictive tracking belief in~\eqref{eq:predictive_belief_definition}. The recurrent state retains temporal context, while the posterior incorporates new visual--inertial observations.

MT-RSSM also decodes task-relevant geometry and multimodal observations,
\begin{align}
\hat{\bm{\rho}}^{\,\mathrm{rel}}_{t}
&=
g_{\mathrm{rel}}(\bm{c}_{t})
\in\mathbb{R}^{6},
\label{eq:mtrssm_pose_head}\\
(\hat{\bm{e}}_{t},\hat{\bm{s}}^{\mathrm{imu}}_{t})
&=
g_{\mathrm{obs}}(\bm{c}_{t}).
\label{eq:mtrssm_obs_head}
\end{align}
The control-relevant state is reconstructed as
\begin{equation}
\hat{\bm{y}}_{t}
=
\Gamma
\left(
\hat{\bm{\rho}}^{\,\mathrm{rel}}_{t},
\hat{\bm{s}}^{\mathrm{imu}}_{t}
\right),
\label{eq:mtrssm_task_state}
\end{equation}
where $\Gamma(\cdot)$ selects relative position and yaw together with follower roll and pitch.

MT-RSSM uses the posterior for inference under incomplete sensing and the action-conditioned prior for predicting candidate action outcomes. Its task-supervised objectives are detailed in Section~\ref{sec:learning}.

\subsection{Diffusion Action Prior}
\label{subsec:dap}

A predictive belief can admit several reasonable multi-thruster responses. Rather than collapsing these behaviors into one deterministic command, DAP models a distribution over complete action trajectories using diffusion \cite{ho2020ddpm,song2021ddim,chi2024diffusionpolicy}. For horizon $H$,
\begin{equation}
\bm{A}_{t}
=
\begin{bmatrix}
\bm{a}^{\top}_{t}\\
\vdots\\
\bm{a}^{\top}_{t+H-1}
\end{bmatrix}
\in\mathbb{R}^{H\times8},
\qquad
p_{\theta_D}(\bm{A}_{t}\mid\bm{c}_{t}).
\label{eq:dap_distribution}
\end{equation}
With $H=10$, DAP models an $80$-dimensional temporally correlated trajectory rather than a single eight-dimensional command.

Let $n\in\{1,\ldots,N_d\}$ denote the diffusion step. The forward process is
\begin{equation}
\bm{A}^{[n]}_{t}
=
\sqrt{\bar{\alpha}_{n}}\bm{A}^{[0]}_{t}
+
\sqrt{1-\bar{\alpha}_{n}}\bm{\epsilon},
\qquad
\bm{\epsilon}\sim\mathcal{N}(\bm{0},\bm{I}),
\label{eq:dap_forward}
\end{equation}
and the temporal U-Net predicts the injected noise conditioned on the current belief,
\begin{equation}
\hat{\bm{\epsilon}}
=
\epsilon_{\theta_D}
(\bm{A}^{[n]}_{t},n,\bm{c}_{t}).
\label{eq:dap_noise_prediction}
\end{equation}
Belief conditioning and trajectory-level generation capture state dependence and inter-thruster coordination.

Online sampling begins with $\bm{A}^{[N_d]}_{t}\sim\mathcal{N}(\bm{0},\bm{I})$ and applies deterministic DDIM updates~\cite{song2021ddim} using $\epsilon_{\theta_D}(\bm{A}^{[n]}_{t},n,\bm{c}_{t})$ to obtain a clean action trajectory. Independent initial noise samples generate
\begin{equation}
\bm{A}^{(k)}_{t}
=
\mathcal{G}_{\theta_D}
(\bm{c}_{t},\bm{\epsilon}^{(k)}),
\qquad
k=1,\ldots,K,
\label{eq:dap_candidates}
\end{equation}
where
$\bm{\epsilon}^{(k)}
\overset{\mathrm{i.i.d.}}{\sim}
\mathcal{N}(\bm{0},\bm{I})$.

\textbf{DAP provides an action-trajectory prior, not the final decision}. It generates expert-like, temporally coherent candidates whose predicted consequences are compared before execution.

\subsection{World-Model Predictive Selector}
\label{subsec:wmps}

WMPS converts DAP diversity into an online predictive decision. Each candidate is initialized from the current posterior belief and propagated through the learned MT-RSSM prior, avoiding reliance on an identified analytical AUV model. For candidate $k$, let
$\bm{h}^{(k)}_{t|t}=\bm{h}_{t}$ and
$\bm{z}^{(k)}_{t|t}=\bm{z}_{t}$.
For $\tau=0,\ldots,H-1$,
\begin{equation}
\bm{h}^{(k)}_{t+\tau+1|t}
=
\operatorname{GRU}
\left(
\bm{h}^{(k)}_{t+\tau|t},
\begin{bmatrix}
\bm{z}^{(k)}_{t+\tau|t}\\
\phi_a(\bm{a}^{(k)}_{t+\tau})
\end{bmatrix}
\right).
\label{eq:wmps_hidden_rollout}
\end{equation}
To avoid sampling variance, online ranking propagates the prior mean,
\begin{equation}
\bm{z}^{(k)}_{t+\tau+1|t}
=
\bm{\mu}^{p,(k)}_{t+\tau+1|t}
=
\left[f_p(\bm{h}^{(k)}_{t+\tau+1|t})\right]_{\mu},
\label{eq:wmps_mean_rollout}
\end{equation}
where $[\cdot]_{\mu}$ extracts the mean of the prior distribution.

The resulting future belief is
$\bm{c}^{(k)}_{t+\tau|t}
=
[{\bm{h}^{(k)}_{t+\tau|t}}^{\top},
 {\bm{z}^{(k)}_{t+\tau|t}}^{\top}]^{\top}$.
Its control-relevant prediction is
\begin{equation}
\hat{\bm{y}}^{(k)}_{t+\tau|t}
=
\Gamma
\left(
g_{\mathrm{rel}}(\bm{c}^{(k)}_{t+\tau|t}),
\left[
g_{\mathrm{obs}}(\bm{c}^{(k)}_{t+\tau|t})
\right]_{\mathrm{imu}}
\right).
\label{eq:wmps_predicted_state}
\end{equation}
Here, $[\cdot]_{\mathrm{imu}}$ extracts the predicted IMU state. The error $\hat{\bm{\varepsilon}}^{(k)}_{t+\tau|t}$ follows~\eqref{eq:tracking_error} with $\bm{y}_t$ replaced by $\hat{\bm{y}}^{(k)}_{t+\tau|t}$. The receding-horizon tracking cost is
\begin{equation}
\widehat{J}_{\mathrm{trk}}^{(k)}
=
\sum_{\tau=1}^{H}
\gamma^{\tau-1}
\left\|
\hat{\bm{\varepsilon}}^{(k)}_{t+\tau|t}
\right\|_{\bm{Q}}^{2},
\label{eq:wmps_tracking_cost}
\end{equation}
where $\bm{Q}\succeq0$ weights relative tracking and follower attitude stabilization. Control effort and temporal variation are penalized by
\begin{align}
J_{u}^{(k)}
&=
\lambda_u
\sum_{\tau=0}^{H-1}
\|\bm{a}^{(k)}_{t+\tau}\|_2^2,
\label{eq:wmps_control_cost}\\
J_{\Delta u}^{(k)}
&=
\lambda_{\Delta u}
\sum_{\tau=1}^{H-1}
\|\bm{a}^{(k)}_{t+\tau}
-\bm{a}^{(k)}_{t+\tau-1}\|_2^2,
\label{eq:wmps_smoothness_cost}\\
J_{\Delta u,0}^{(k)}
&=
\lambda_{\Delta u,0}
\|\bm{a}^{(k)}_{t}-\bm{a}_{t-1}\|_2^2.
\label{eq:wmps_initial_smoothness_cost}
\end{align}
The total predicted cost is
\begin{equation}
\widehat{J}_{t}(\bm{A}^{(k)}_{t})
=
\widehat{J}_{\mathrm{trk}}^{(k)}
+J_{u}^{(k)}
+J_{\Delta u}^{(k)}
+J_{\Delta u,0}^{(k)}.
\label{eq:wmps_total_cost}
\end{equation}
WMPS selects
\begin{equation}
\hat{k}_{t}
=
\arg\min_{k\in\{1,\ldots,K\}}
\widehat{J}_{t}(\bm{A}^{(k)}_{t}),
\label{eq:wmps_selection}
\end{equation}
and executes only
\begin{equation}
\bm{a}_{t}
=
\Pi_{\mathcal{A}}
(\bm{a}^{(\hat{k}_{t})}_{t}),
\label{eq:wmps_execute}
\end{equation}
where $\Pi_{\mathcal{A}}(\cdot)$ enforces admissible thruster bounds. New observations correct the posterior belief at the next step, closing the predictive control loop.

\subsection{Predictive Selection Analysis}
\label{subsec:selection_theory}

The quality of WMPS depends on how accurately MT-RSSM ranks the sampled DAP candidates. Let $J_t(\bm{A}^{(k)}_{t})$ denote the true finite-horizon cost and $\widehat{J}_t(\bm{A}^{(k)}_{t})$ its world-model estimate. Assume a uniform prediction error over the sampled set,
\begin{equation}
\left|
J_t(\bm{A}^{(k)}_{t})
-
\widehat{J}_t(\bm{A}^{(k)}_{t})
\right|
\le
\epsilon_J,
\quad
k=1,\ldots,K.
\label{eq:cost_error_bound}
\end{equation}
Define the predicted and oracle selections as
\begin{equation}
\hat{k}_{t}
=
\arg\min_k\widehat{J}_t(\bm{A}^{(k)}_{t}),
\qquad
k^{\star}_{t}
=
\arg\min_k J_t(\bm{A}^{(k)}_{t}).
\label{eq:predicted_oracle_selection}
\end{equation}

\noindent\textbf{Proposition 1.}
Under~\eqref{eq:cost_error_bound}, the true cost of the WMPS-selected trajectory satisfies
\begin{equation}
J_t(\bm{A}^{(\hat{k}_{t})}_{t})
\le
J_t(\bm{A}^{(k^{\star}_{t})}_{t})
+2\epsilon_J.
\label{eq:selection_regret_bound}
\end{equation}

\noindent\emph{Proof.}
By~\eqref{eq:cost_error_bound} and optimality of $\hat{k}_{t}$ under the predicted cost,
\begin{align}
J_t(\bm{A}^{(\hat{k}_{t})}_{t})
&\le
\widehat{J}_t(\bm{A}^{(\hat{k}_{t})}_{t})
+\epsilon_J
\nonumber\\
&\le
\widehat{J}_t(\bm{A}^{(k^{\star}_{t})}_{t})
+\epsilon_J
\nonumber\\
&\le
J_t(\bm{A}^{(k^{\star}_{t})}_{t})
+2\epsilon_J.
\label{eq:selection_regret_proof}
\end{align}
\hfill$\square$

This bound links world-model accuracy directly to online selection quality. If the state-dependent stage cost is $L_{\ell}$-Lipschitz over the operating region and the $\tau$-step prediction error satisfies
\begin{equation}
\left\|
\bm{y}_{t+\tau}
-
\hat{\bm{y}}_{t+\tau|t}
\right\|_2
\le
\epsilon_{\tau},
\qquad
\tau=1,\ldots,H,
\label{eq:state_prediction_bound}
\end{equation}
then, because the action-dependent penalties are evaluated exactly from the known candidate sequence,
\begin{equation}
\epsilon_J
\le
L_{\ell}
\sum_{\tau=1}^{H}
\gamma^{\tau-1}\epsilon_{\tau}.
\label{eq:cost_state_error_relation}
\end{equation}
Combining the two bounds gives
\begin{equation}
J_t(\bm{A}^{(\hat{k}_{t})}_{t})
-
J_t(\bm{A}^{(k^{\star}_{t})}_{t})
\le
2L_{\ell}
\sum_{\tau=1}^{H}
\gamma^{\tau-1}\epsilon_{\tau}.
\label{eq:selection_state_bound}
\end{equation}
Thus, multi-step prediction accuracy bounds selection degradation under the stated assumptions.

For nested candidate sets, the minimum true cost cannot increase with $K$, although model errors may cause incorrect rankings. Longer horizons provide greater look-ahead but accumulate prediction errors. These tradeoffs motivate the $K$ and horizon analyses
in Section~\ref{sec:offline_eval}.

\subsection{Overall Online Algorithm}
\label{subsec:online_algorithm}

Algorithm~\ref{alg:fathom_online} summarizes the online loop. The current belief is corrected from real onboard observations, while hypothetical candidates are evaluated using deterministic prior-mean rollouts. Diffusion generation and world-model prediction are batched across all $K$ candidates.

\begin{algorithm}[t]
\caption{Online FATHOM Control}
\label{alg:fathom_online}
\begin{algorithmic}[1]
\Require Trained SSME, MT-RSSM, and DAP
\Require Target $\bm{y}^{*}$, candidate number $K$, horizon $H$
\State Initialize $\bm{h}_{-1}$, $\bm{z}_{-1}$,
and $\bm{a}_{-1}$ to zero

\For{$t=0,1,2,\ldots$}
    \State Acquire sparse markers $\mathcal{D}_{t}$ and IMU state $\bm{s}^{\mathrm{imu}}_{t}$
    \State Encode $\mathcal{D}_{t}$ with SSME to obtain $\bm{e}_{t}$
    \State Fuse $\bm{e}_{t}$ and $\bm{s}^{\mathrm{imu}}_{t}$ into $\bm{x}^{o}_{t}$
    
    \State Update recurrent state $\bm{h}_{t}$ using $\bm{z}_{t-1}$ and $\bm{a}_{t-1}$
    \State Compute MT-RSSM posterior $q(\bm{z}_{t}\mid\bm{h}_{t},\bm{x}^{o}_{t})$
    \State Set $\bm{z}_{t}\gets\bm{\mu}^{q}_{t}$ and form belief $\bm{c}_{t}$
    
    \State Generate $\{\bm{A}^{(k)}_{t}\}_{k=1}^{K}$ from DAP conditioned on $\bm{c}_{t}$
    \State Initialize all candidate rollouts from $(\bm{h}_{t},\bm{z}_{t})$
    
    \ForAll{$k=1,\ldots,K$ \textbf{ in parallel}}
        \State Roll out $\bm{A}^{(k)}_{t}$ using MT-RSSM prior means
        \State Decode $\{\hat{\bm{y}}^{(k)}_{t+\tau|t}\}_{\tau=1}^{H}$
        \State Evaluate $\widehat{J}_{t}(\bm{A}^{(k)}_{t})$
    \EndFor
    
    \State Select
    $\hat{k}_{t}\gets
    \arg\min_{k}\widehat{J}_{t}(\bm{A}^{(k)}_{t})$
    \State Execute
    $\bm{a}_{t}\gets
    \Pi_{\mathcal{A}}(\bm{a}^{(\hat{k}_{t})}_{t})$
    \State Store the executed $\bm{a}_{t}$ for the next belief update
\EndFor
\end{algorithmic}
\end{algorithm}

\begin{figure}[t]
    \centering
    \includegraphics[width=0.98\columnwidth]{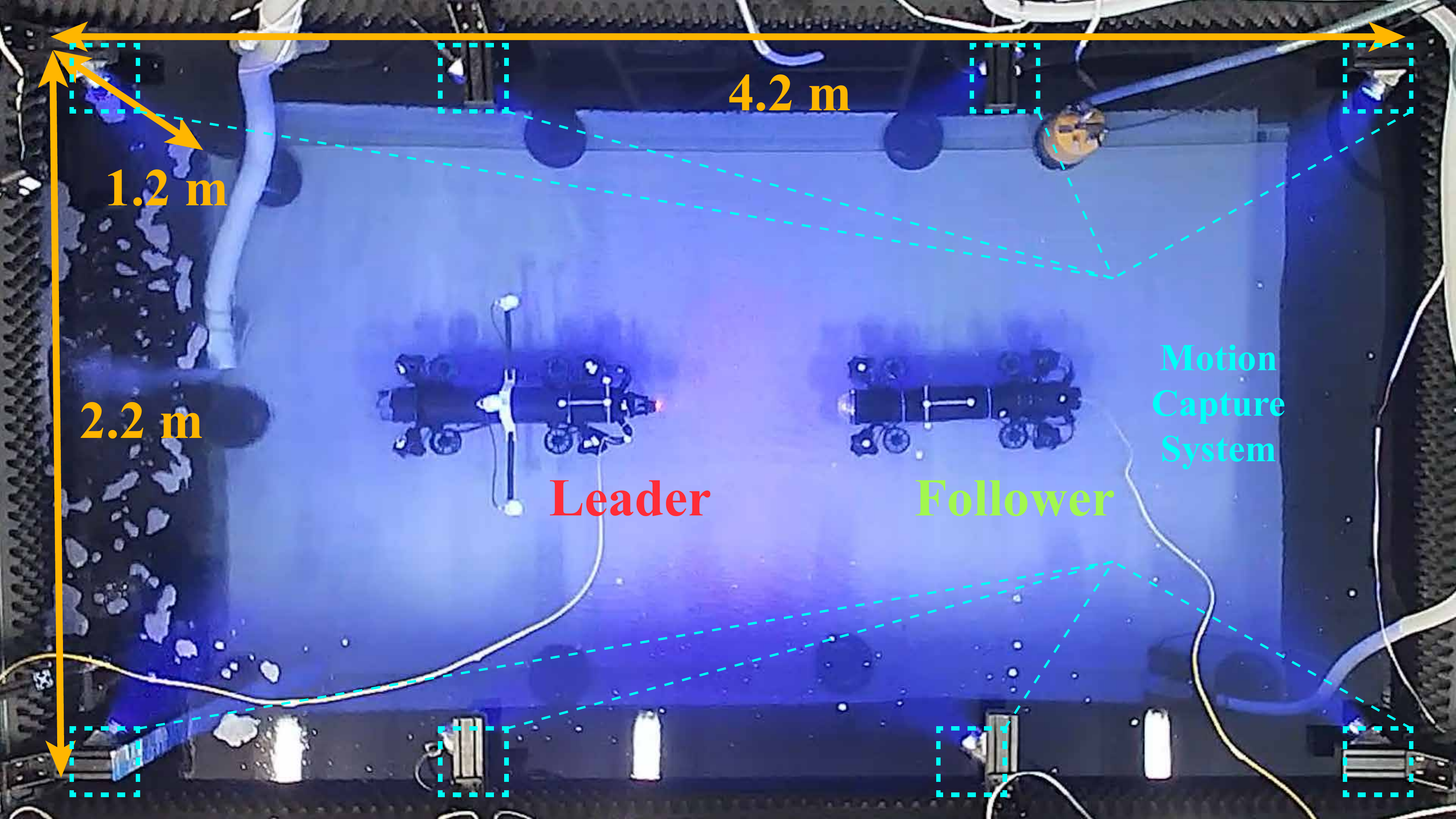}
    \caption{Real-world leader--follower AUV experimental setup in a $4.2\times2.2\times1.2$~m water tank, showing the two vehicles and surrounding motion-capture system.}
    \label{fig:experimental_setup}
\end{figure}

\begin{figure*}[t]
    \centering
    \includegraphics[width=0.98\linewidth]{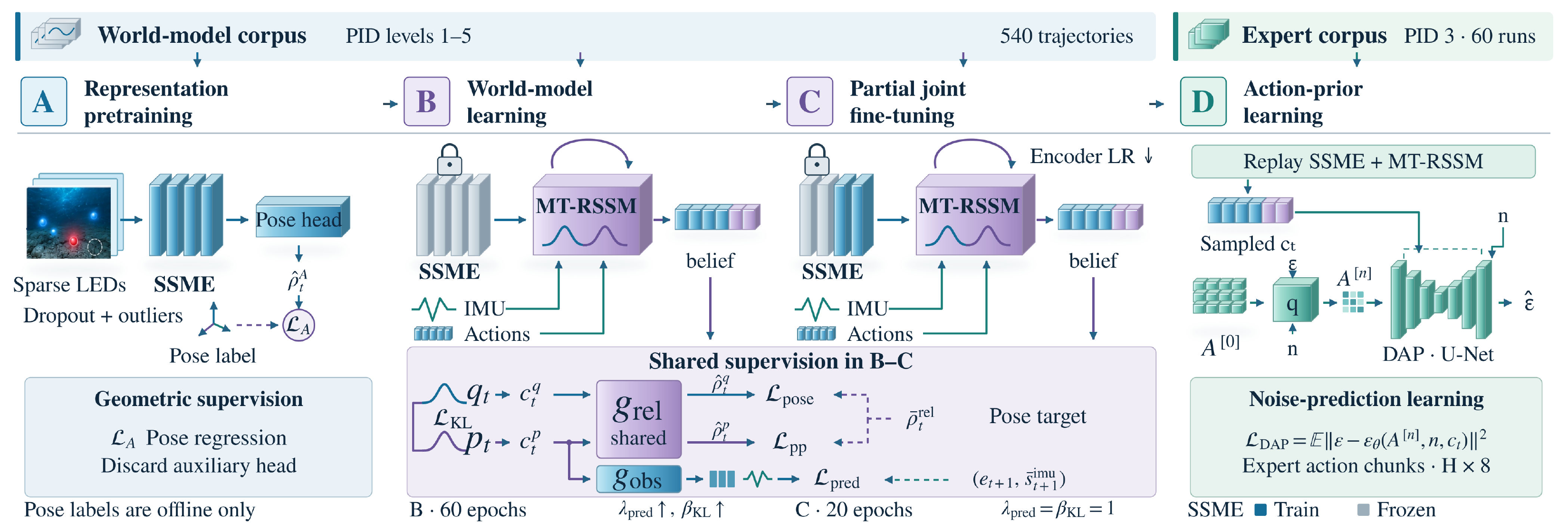}
    \caption{Four-stage training pipeline of FATHOM.
    (A) SSME representation pretraining with an auxiliary pose head.
    (B) MT-RSSM learning with frozen SSME.
    (C) Partial joint fine-tuning with a reduced encoder learning rate.
    Stages B and C share posterior and prior pose supervision, multimodal prediction, and KL regularization.
    (D) DAP learning from expert action trajectories.
    Stages A--C use the 540-trajectory world-model corpus, whereas Stage D uses 60 expert trajectories.}
    \label{fig:fathom_training}
\end{figure*}

\section{Physical Data Collection and Learning}
\label{sec:learning}

FATHOM uses a broad physical corpus for marker representation and dynamics learning, and a smaller expert corpus for DAP. The four-stage training procedure is shown in Fig.~\ref{fig:fathom_training}.

\subsection{Physical Data Collection}
\label{subsec:data_collection}

All data are collected using the dual-AUV platform in a
$4.2\times2.2\times1.2~\mathrm{m}$ water tank with an operating depth of approximately $0.9~\mathrm{m}$ (Fig.~\ref{fig:experimental_setup}). During collection, motion capture provides the relative pose ${}^{F}\bm{T}_{L,t}$ to a PID controller on the follower, while the leader is manually operated through a joystick. The leader motion includes longitudinal and lateral translation, depth variation, yaw turns, and coupled maneuvers. Consequently, the recordings naturally contain rapid relative motion, partial marker visibility, temporary LED loss and recovery, reflections, and visual interference from the underwater motion-capture illumination.

Two physical datasets are constructed, as summarized in Table~\ref{tab:dataset_composition}. The first is the \emph{world-model corpus} used for SSME and MT-RSSM. Training only on well-tracked trajectories would concentrate the data near the nominal tracking manifold and provide insufficient coverage of unfavorable state--action transitions. We therefore use five PID gain levels $g\in\{1,\ldots,5\}$. Level $3$ is the tuned expert controller, levels $1$--$2$ produce increasingly under-responsive tracking, and levels $4$--$5$ yield progressively aggressive responses and overshoot.

For each PID level, nine desired relative positions are used,
\begin{equation}
\begin{aligned}
\mathcal{P}_{\mathrm{WM}}=\{&
(1.3,-0.2),(1.3,0),(1.3,0.2),\\
&(1.5,-0.35),(1.5,0),(1.5,0.35),\\
&(1.7,-0.5),(1.7,0),(1.7,0.5)\}\ {\rm m},
\end{aligned}
\label{eq:wm_target_set}
\end{equation}
where each pair denotes $(x_{\mathrm{rel}}^{*},y_{\mathrm{rel}}^{*})$, with
$z_{\mathrm{rel}}^{*}=\psi_{\mathrm{rel}}^{*}=0$.
For every PID--target combination, six nominal-motion and six challenging-motion runs are recorded, resulting in
$5\times9\times2\times6=540$ physical trajectories. This combination provides diverse tracking quality, relative configurations, and marker-visibility conditions, allowing the learned world model to distinguish favorable and unfavorable future evolution.

The second dataset is the \emph{expert corpus} used by DAP. The motion-capture-based level-$3$ PID provides privileged and accurately tuned relative tracking during data collection. Its desired relative configuration is fixed at
$\bm{p}^{*}=[1.5,0,0]^{\top}\,\mathrm{m}$ and $\psi^{*}=0$.
Thirty nominal-motion and thirty challenging-motion trajectories are collected, giving 60 expert runs in total. The trajectories cover slow tracking, rapid translation, depth changes, turns, and coupled maneuvers.

Thus, the 540-trajectory corpus captures diverse dynamics and observations, whereas the 60 expert trajectories provide coherent behavior for DAP.

\begin{table*}[t]
\centering
\caption{Composition and Roles of the Physical Datasets}
\label{tab:dataset_composition}
\begin{tabular}{lcccccc}
\hline
\textbf{Dataset} &
\textbf{Controller} &
\textbf{Targets} &
\textbf{Motion Regimes} &
\textbf{Runs/Condition} &
\textbf{Total} &
\textbf{Role} \\
\hline
World-model corpus &
PID levels 1--5 &
9 &
Nominal / Challenging &
6 &
540 &
SSME, MT-RSSM \\
Expert corpus &
PID level 3 &
$(1.5,0)$ &
Nominal / Challenging &
30 &
60 &
DAP \\
\hline
\end{tabular}
\end{table*}

\subsection{Data Processing and Normalization}
\label{subsec:data_processing}

Each trajectory synchronizes LED detections, IMU measurements, eight-thruster commands, and motion-capture poses. Motion capture serves only offline supervision and evaluation, not autonomous feedback.

The full relative pose used for representation and world-model supervision is
\begin{equation}
\bm{\rho}^{\mathrm{rel}}_t
=
[x_{FL,t},y_{FL,t},z_{FL,t},
\phi_{FL,t},\theta_{FL,t},\psi_{FL,t}]^{\top}.
\label{eq:training_relative_pose}
\end{equation}
Motion-capture pose signals are smoothed offline using a second-order zero-phase Butterworth filter with a $1.5~\mathrm{Hz}$ cutoff. Angular signals are unwrapped before filtering and wrapped afterward. This noncausal filtering is applied only to training labels.

The onboard IMU follows the causal preprocessing used during deployment,
\begin{equation}
\widetilde{\bm{s}}^{\mathrm{imu}}_t
=
\alpha_{\mathrm{imu}}\bm{s}^{\mathrm{imu}}_t
+
(1-\alpha_{\mathrm{imu}})
\widetilde{\bm{s}}^{\mathrm{imu}}_{t-1},
\qquad
\alpha_{\mathrm{imu}}=0.35.
\label{eq:imu_ema}
\end{equation}
LED detections are converted into the structured red-anchor and blue-set inputs of SSME, with validity masks used for missing and padded observations. Pose, IMU, and thruster variables are then mapped to fixed normalized ranges using affine transformations without clipping, so that off-nominal physical states remain available to the learning process.

Splitting is performed at the trajectory level to prevent temporal leakage. For each six-run condition, four runs are assigned to training, one to validation, and one to testing. This yields 360/90/90 trajectories for the world-model corpus and 40/10/10 for the expert corpus while preserving controller, target, and motion-regime distributions.

\subsection{SSME Representation Pretraining}
\label{subsec:ssme_training}

SSME is first pretrained so that the sparse marker representation retains relative geometric information before recurrent dynamics are introduced. A temporary pose head $g_A(\cdot)$ maps the encoded feature $\bm{e}_t$ to the normalized relative pose,
\begin{equation}
\hat{\bm{\rho}}^{A}_t=g_A(\bm{e}_t),
\qquad
\mathcal{L}_{A}
=
\frac{1}{N_A}
\sum_t
\left\|
\hat{\bm{\rho}}^{A}_t-
\bar{\bm{\rho}}^{\mathrm{rel}}_t
\right\|_2^2.
\label{eq:ssme_pretrain_loss}
\end{equation}
This provides geometric supervision without requiring a deployed pose estimator.

To expose SSME to the partial observations encountered underwater, marker-level augmentation randomly removes red and blue detections and injects a small number of spurious blue candidates. The probabilities are $0.05$, $0.15$, and $0.10$ for red dropout, blue dropout, and spurious-marker injection, respectively. The auxiliary pose head is discarded after Stage A.

\subsection{MT-RSSM Learning}
\label{subsec:mtrssm_training}

MT-RSSM is trained on physical sequences for observation-conditioned inference and action-conditioned prediction. Tracking-related pose supervision further shapes its latent state for predictive selection.

During training, the posterior latent is sampled using
\begin{equation}
\bm{z}^{q}_t
=
\bm{\mu}^{q}_t+
\bm{\sigma}^{q}_t\odot\bm{\xi}_t,
\qquad
\bm{\xi}_t\sim\mathcal{N}(\bm{0},\bm{I}),
\label{eq:rssm_reparameterization}
\end{equation}
forming the posterior belief
$\bm{c}^{q}_t=[\bm{h}_t^{\top},{\bm{z}^{q}_t}^{\top}]^{\top}$.
The corresponding prior belief is
$\bm{c}^{p}_t=[\bm{h}_t^{\top},{\bm{\mu}^{p}_t}^{\top}]^{\top}$.

To impose task relevance, the posterior is supervised by the full relative pose,
\begin{equation}
\mathcal{L}_{\mathrm{pose}}
=
\frac{1}{N_s}
\sum_t
\left\|
g_{\mathrm{rel}}(\bm{c}^{q}_t)
-
\bar{\bm{\rho}}^{\mathrm{rel}}_t
\right\|_2^2.
\label{eq:rssm_pose_loss}
\end{equation}
Offline pose supervision preserves leader--follower geometry in the belief without requiring pose measurements in the online control loop.

To learn action-conditioned predictive dynamics, the prior reconstructs the next structured visual feature and IMU state,
\begin{equation}
\mathcal{L}_{\mathrm{pred}}
=
\frac{1}{N_s}
\sum_t
\left(
\left\|
\hat{\bm{e}}^{p}_{t+1}-\bm{e}_{t+1}
\right\|_2^2
+
\left\|
\hat{\bm{s}}^{p}_{t+1}
-
\bar{\bm{s}}^{\mathrm{imu}}_{t+1}
\right\|_2^2
\right).
\label{eq:rssm_prediction_loss}
\end{equation}

Posterior and prior distributions are regularized using a free-bits KL objective,
\begin{equation}
\begin{aligned}
\mathcal{L}_{\mathrm{KL}}
=
\frac{1}{N_s}\sum_t\sum_{j=1}^{d_z}
\max\Big\{
&D_{\mathrm{KL}}\big[
q(z_{t,j}\mid\bm{h}_t,\bm{x}^{o}_t)
\\[-1mm]
&\Vert p(z_{t,j}\mid\bm{h}_t)
\big],
\lambda_{\mathrm{fb}}
\Big\},
\end{aligned}
\label{eq:rssm_kl_loss}
\end{equation}
where $d_z=32$ and $\lambda_{\mathrm{fb}}=1$. Since WMPS evaluates candidate actions through prior rollout, the prior is additionally constrained to preserve relative-pose information,
\begin{equation}
\mathcal{L}_{\mathrm{pp}}
=
\frac{1}{N_s}
\sum_t
\left\|
g_{\mathrm{rel}}(\bm{c}^{p}_t)
-
\bar{\bm{\rho}}^{\mathrm{rel}}_t
\right\|_2^2.
\label{eq:rssm_prior_pose_loss}
\end{equation}

The complete world-model objective is
\begin{equation}
\mathcal{L}_{\mathrm{WM}}
=
\mathcal{L}_{\mathrm{pose}}
+
\lambda_{\mathrm{pred}}\mathcal{L}_{\mathrm{pred}}
+
\beta_{\mathrm{KL}}\mathcal{L}_{\mathrm{KL}}
+
\lambda_{\mathrm{pp}}\mathcal{L}_{\mathrm{pp}}.
\label{eq:wm_training_loss}
\end{equation}
The pose, prediction, prior-pose, and KL terms jointly enforce task relevance, multimodal dynamics, and regularized belief inference.

Training uses a $10$-step burn-in followed by a $50$-step recurrent learning interval with one-step predictive supervision. The burn-in updates the recurrent state without contributing to the loss, allowing temporal context to form before supervision begins. In Stage B, SSME is frozen while MT-RSSM is trained, with $\lambda_{\mathrm{pred}}$ increased from $0.1$ to $1$ and $\beta_{\mathrm{KL}}$ from $0$ to $1$ during the first 20 epochs; $\lambda_{\mathrm{pp}}=0.2$ remains fixed. Stage C starts from the Stage-B model and jointly fine-tunes MT-RSSM and the task-facing SSME layers using a lower encoder learning rate, while fixing $\lambda_{\mathrm{pred}}=\beta_{\mathrm{KL}}=1$ and $\lambda_{\mathrm{pp}}=0.2$.

\subsection{DAP Learning}
\label{subsec:dap_training}

After world-model training, DAP learns from the 60 expert trajectories. Each sequence is replayed through SSME and MT-RSSM to construct belief--action pairs $\mathcal{D}_{\mathrm{DAP}}=\{(\bm{c}_t,\bm{A}^{[0]}_t)\}$, where $\bm{A}^{[0]}_t=[\bm{a}_t,\ldots,\bm{a}_{t+H-1}]$. Using the forward diffusion process in~\eqref{eq:dap_forward}, DAP minimizes the standard noise-prediction objective \cite{ho2020ddpm},
\begin{equation}
\mathcal{L}_{\mathrm{DAP}}
=
\mathbb{E}
\left[
\left\|
\bm{\epsilon}
-
\epsilon_{\theta_D}
\left(
\bm{A}^{[n]}_t,n,\bm{c}_t
\right)
\right\|_2^2
\right],
\label{eq:dap_training_loss}
\end{equation}
where the expectation is taken over expert action chunks, diffusion steps, and Gaussian noise.

The loss spans the full $H\times8$ trajectory to capture temporal and inter-thruster coordination. Training uses posterior-sampled latent conditions, whereas deployment uses the posterior mean in~\eqref{eq:predictive_tracking_belief} for deterministic belief estimation. The diffusion process contains $1000$ training steps with a cosine schedule, while online generation uses a 10-step DDIM sampler \cite{song2021ddim}.

\subsection{Training Configuration and Implementation Details}
\label{subsec:training_config}

Table~\ref{tab:training_config} summarizes optimization settings; architecture dimensions are given in Section~\ref{sec:fathom}.

\begin{table*}[t]
\centering
\caption{Training Configuration of FATHOM}
\label{tab:training_config}
\begin{tabular}{lccccccc}
\hline
\textbf{Stage} &
\textbf{Epochs} &
\textbf{Batch} &
\textbf{Learning Rate} &
\textbf{Weight Decay} &
\textbf{Grad. Clip} &
\textbf{Sequence} &
\textbf{Setting} \\
\hline
SSME A
& 80
& 256
& $1\times10^{-3}$
& $1\times10^{-4}$
& 1
& Frame-wise
& Pose pretraining \\

MT-RSSM B
& 60
& 32
& $3\times10^{-4}$
& $1\times10^{-5}$
& 10
& $10+50$
& SSME frozen \\

MT-RSSM C
& 20
& 32
& $3\times10^{-4}$ / $5\times10^{-5}$
& $1\times10^{-5}$
& 10
& $10+50$
& Partial fine-tuning \\

DAP
& 250
& 256
& $2\times10^{-4}$
& $1\times10^{-6}$
& 1
& $H=10$
& $N_d=1000$, DDIM $=10$ \\
\hline
\end{tabular}
\end{table*}

Training uses AdamW and automatic mixed precision, with validation-based model selection. Section~\ref{sec:offline_eval} evaluates held-out representation, prediction, action generation, and selection performance.

\section{Offline Evaluation and Algorithmic Analysis}
\label{sec:offline_eval}

\subsection{Offline Evaluation Protocol}
\label{subsec:offline_protocol}

Validation data are used only for checkpoint selection, and all results are evaluated on held-out physical trajectories. We assess predictive belief and latent dynamics, diffusion action generation, and world-model-guided selection. In the normalized learning space, NRMSE for a $d$-dimensional sequence is
$\sqrt{\frac{1}{Td}\sum_{t=1}^{T}\|\hat{\bar{\bm{x}}}_{t}-\bar{\bm{x}}_{t}\|_2^2}$.
Metrics are averaged within each trajectory and then across trajectories. Posterior inference uses the current visual--inertial observation, whereas prior prediction propagates the belief under recorded actions without future observation correction. The latter therefore tests whether task-relevant information persists without subsequent measurement updates.

\subsection{Predictive Belief and Multimodal World-Model Prediction}
\label{subsec:belief_worldmodel_eval}

We compare Full MT-RSSM with three ablations. \emph{Generic Set} replaces SSME with a permutation-invariant encoder of marker coordinates and colors, without red-relative geometry or role-based aggregation. \emph{No Temporal} removes observation--action history from belief inference but retains action-conditioned prediction. \emph{No Task Sup.} removes $\mathcal{L}_{\mathrm{pose}}$ and $\mathcal{L}_{\mathrm{pp}}$, while retaining predictive and KL losses. A common lightweight probe trained on frozen beliefs evaluates relative pose, while the learned decoders provide IMU and visual predictions. This separates the predictive representation from any advantage due to a variant-specific pose head.

Table~\ref{tab:worldmodel_offline} compares posterior inference and prior prediction. Rel.~Pose measures full 6-DoF relative pose, IMU State the five-dimensional inertial state, and Visual Feat. the structured visual representation. Task State @$k$ evaluates the control-relevant state in~\eqref{eq:relative_tracking_state}.

\begin{table*}[t]
\centering
\caption{Held-out predictive-belief and multimodal world-model evaluation.
All entries are trajectory-averaged NRMSE; lower is better.
\emph{Task State @$k$} denotes the $k$-step prediction NRMSE
of the six-dimensional control-relevant task state, while
\emph{Multimodal @10} is the equal-weight mean of the
relative-pose, IMU-state, and visual-feature NRMSEs at step 10.}
\label{tab:worldmodel_offline}

\renewcommand{\arraystretch}{1.12}
\setlength{\tabcolsep}{5.0pt}

\begin{tabular}{lccc!{\vrule width 0.7pt}ccccc}
\hline
& \multicolumn{3}{c}{Posterior Inference}
& \multicolumn{5}{c}{Prior Prediction} \\
\cline{2-4} \cline{5-9}

Method
& Rel. Pose
& IMU State
& Visual Feat.
& \makecell[c]{Task State @1}
& \makecell[c]{Task State @3}
& \makecell[c]{Task State @5}
& \makecell[c]{Task State @10}
& Multimodal @10 \\
\hline

Generic Set
& 0.090
& 0.082
& 0.417
& \textbf{0.264}
& 0.273
& 0.321
& 0.371
& 0.496 \\

No Temporal
& 0.092
& 0.384
& 2.024
& 0.335
& 0.734
& 0.903
& 1.617
& 6.163 \\

No Task Sup.
& 0.227
& \textbf{0.041}
& 0.710
& 0.618
& 0.918
& 1.025
& 1.072
& 0.759 \\

Full MT-RSSM
& \textbf{0.084}
& 0.059
& \textbf{0.399}
& \textbf{0.264}
& \textbf{0.261}
& \textbf{0.282}
& \textbf{0.314}
& \textbf{0.457} \\

\hline
\end{tabular}
\end{table*}

Full MT-RSSM achieves the lowest Task State @10 NRMSE of $0.314$, a $15.4\%$ reduction from Generic Set ($0.371$), while Multimodal @10 improves from $0.496$ to $0.457$. Both models obtain $0.264$ at Task State @1, indicating that structured marker encoding becomes more beneficial over longer predictions. The full model also yields the lowest posterior relative-pose NRMSE ($0.084$) and visual-feature NRMSE ($0.399$).

Temporal memory and task supervision are more consequential. Removing temporal history raises Task State @10 from $0.314$ to $1.617$, over five times the full-model error. No Task Sup. yields the lowest IMU posterior error ($0.041$), yet its Task State @10 error reaches $1.072$. Although the IMU modality is reconstructed accurately, the degraded task-state prediction shows that modality-level reconstruction alone does not preserve the geometry and motion information required for control.

\begin{figure}[t]
    \centering
    \includegraphics[width=\columnwidth]{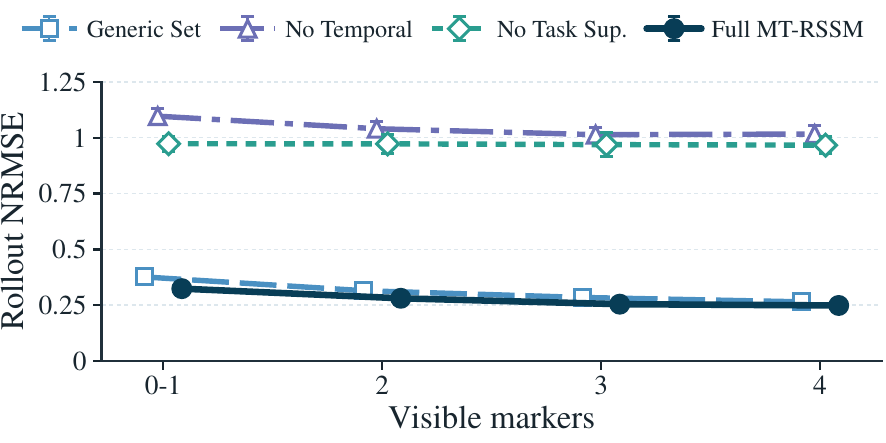}
    \caption{Task-state rollout error under different marker-visibility conditions.
    The 10-step rollout NRMSE is conditioned on the validated marker count at the rollout start.
    Error bars indicate 95\% confidence intervals across held-out test trajectories.}
    \label{fig:rssm_visibility}
\end{figure}

Figure~\ref{fig:rssm_visibility} examines 10-step task-state trajectory error across marker-visibility levels, grouped by the validated marker count at prediction start. With four visible markers, Full MT-RSSM and Generic Set achieve approximately $0.25$ and $0.27$ NRMSE, respectively. In the $0$--$1$ marker bin, these errors rise to about $0.35$ and $0.39$, widening the gap as geometric evidence becomes sparse. No Temporal and No Task Sup. remain near or above $0.9$ even at higher visibility. These results show that marker completeness alone does not replace temporal memory and task supervision for multi-step prediction.

\subsection{Diffusion Action Prior Analysis}
\label{subsec:dap_eval}

We compare DAP with deterministic behavior cloning (BC) using the same 60-trajectory expert corpus, predictive-belief condition, and $H=10$ horizon. BC directly regresses one $10\times8$ action trajectory, whereas DAP provides $K=16$ samples. Coverage NRMSE uses the closest DAP sample to the expert versus BC's single prediction. SWD evaluates distribution matching, while the remaining metrics quantify action magnitude, temporal variation, and sample diversity.

\begin{table}[t]
\centering
\caption{Held-out action-trajectory modeling performance.
BC denotes deterministic behavior cloning.
Lower is better for the first four metrics.
Coverage NRMSE uses the best of $K=16$ DAP samples and the single BC output.
SWD denotes the sliced Wasserstein distance to the expert distribution;
the gap metrics measure deviations from expert action RMS and temporal variation,
while Pairwise Div. measures trajectory diversity.}
\label{tab:dap_bc}

\renewcommand{\arraystretch}{1.10}
\setlength{\tabcolsep}{2.0pt}

\begin{tabular*}{0.98\columnwidth}
{@{\extracolsep{\fill}}lccccc@{}}
\hline

Method
& Coverage
& SWD
& Action RMS
& Temp. Var.
& Pairwise \\
& NRMSE
&
& Gap
& Gap
& Div. \\

\hline

BC
& 0.0703
& 0.0189
& 0.0170
& 0.0203
& 0.0000 \\

DAP
& \textbf{0.0570}
& \textbf{0.0151}
& \textbf{0.0016}
& \textbf{0.0081}
& 0.0876 \\

\hline
\end{tabular*}
\end{table}

As shown in Table~\ref{tab:dap_bc}, DAP reduces Coverage NRMSE from $0.0703$ to $0.0570$ ($18.9\%$) and SWD by $20.1\%$. Action RMS and Temporal Variation gaps decrease by $90.6\%$ and $60.1\%$, respectively, while Pairwise Diversity reaches $0.0876$ versus zero for BC. These improvements jointly indicate that the increased coverage is not obtained by introducing arbitrary action variation. DAP maintains an expert-like action distribution and produces multiple temporally coherent responses.

\begin{figure}[t]
    \centering
    \includegraphics[width=\columnwidth]{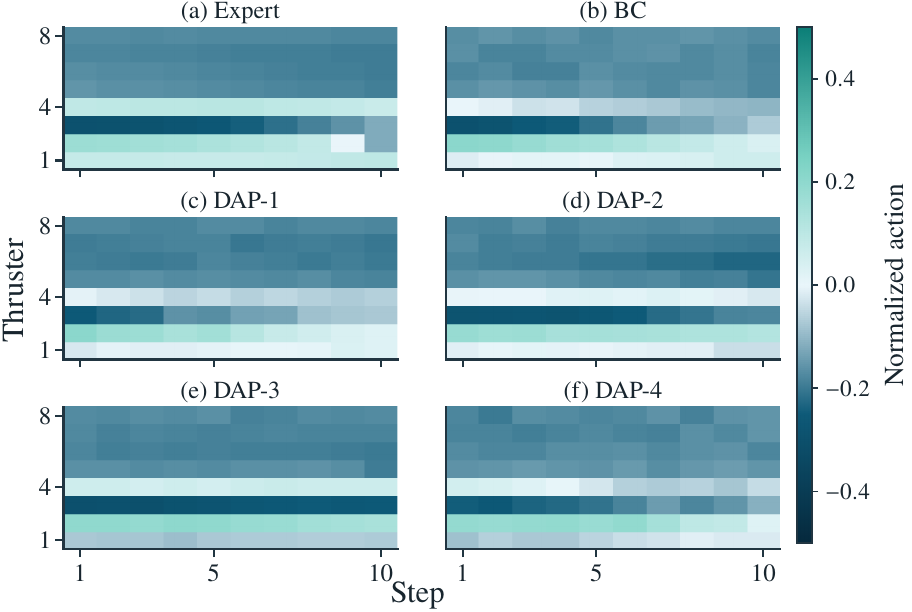}
    \caption{Representative action-trajectory comparison on the held-out test set.
    (a) Expert action trajectory and (b) deterministic behavior-cloning prediction.
    (c)--(f) Four independently sampled DAP trajectories conditioned on the same predictive tracking belief.
    All panels show the normalized commands of the eight thrusters over the $H=10$ prediction horizon and share the same color scale.}
    \label{fig:dap_action_heatmaps}
\end{figure}

Figure~\ref{fig:dap_action_heatmaps} compares an expert action sequence, BC's deterministic output, and four DAP samples for the same belief. BC approximates one action pattern, whereas DAP retains the expert's coordinated cross-thruster structure across distinct samples with varying local amplitudes and temporal profiles. Together with Table~\ref{tab:dap_bc}, this shows that DAP offers WMPS diverse, expert-like alternatives rather than unstructured thruster noise.

\subsection{World-Model Predictive Action-Trajectory Selection}
\label{subsec:wmps_eval}

We next assess whether predictive selection improves the use of DAP candidates. DAP-only selects the first trajectory from the same candidate set as FATHOM, isolating selection from sampling differences. FATHOM evaluates $K=16$ trajectories using MT-RSSM and the WMPS objective. CEM uses the same predictor and cost but replaces DAP with a Gaussian proposal, evaluating eight samples over two updates for the same 16-candidate budget.

\begin{table}[t]
\centering
\caption{Offline comparison of trajectory proposal and predictive selection.
Pred. Cost uses the common MT-RSSM/WMPS objective, and Cost Gain is relative
to DAP-only. SWD denotes the sliced Wasserstein distance to expert trajectories.
Runtime is the batched offline per-window computation time.}
\label{tab:selection_offline}

\renewcommand{\arraystretch}{1.12}
\setlength{\tabcolsep}{2.0pt}
\footnotesize

\begin{tabular*}{0.98\columnwidth}
{@{\extracolsep{\fill}}lcccccc@{}}
\hline

Method
& Learned
& Predictive
& Pred.
& Cost Gain
& SWD
& Runtime \\
&
& Prior
& Selection
& Cost
& (\%)
& (ms) \\

\hline

DAP-only
& $\checkmark$
& $\times$
& 1.209
& --
& \textbf{0.016}
& 1.550 \\

CEM
& $\times$
& $\checkmark$
& 2.616
& -116.3
& 0.134
& \textbf{0.607} \\

FATHOM
& $\checkmark$
& $\checkmark$
& \textbf{0.990}
& \textbf{+18.1}
& 0.036
& 1.915 \\

\hline
\end{tabular*}
\end{table}

Table~\ref{tab:selection_offline} shows that FATHOM reduces predicted cost from $1.209$ to $0.990$ relative to DAP-only, an $18.1\%$ model-space gain. Under the same evaluation budget, CEM yields a higher predicted cost ($2.616$) and SWD ($0.134$) than FATHOM ($0.036$). DAP-only has the lowest SWD ($0.016$), whereas FATHOM accepts a modest deviation from expert behavior to obtain lower predicted cost. This supports the complementary roles of the learned action prior and predictive selection. The cost improvement is measured within the learned model and is distinguished from realized closed-loop tracking performance evaluated in the real-AUV experiments.

Table~\ref{tab:kh_sensitivity} examines candidate number and prediction horizon. Pred. Gain is measured relative to the first DAP candidate. The $K$ sweep uses nested subsets of a common $K=32$ sample pool. In the horizon sweep, DAP always generates $H=10$ actions with $K=16$, but WMPS varies the prediction depth. Task Pred. NRMSE is the endpoint prediction error under recorded actions. Unlike Table~\ref{tab:worldmodel_offline}, it uses the trained pose head and expert test set rather than a diagnostic probe and world-model test set. Runtime measures batched offline computation per window.

\begin{table}[t]
\centering
\caption{Sensitivity of WMPS to candidate number and rollout horizon.
Pred. Gain denotes the reduction in model-predicted cost relative to the first
DAP candidate. Task Pred. NRMSE is the endpoint task-state rollout error under
recorded actions. The candidate-number sweep uses nested candidate sets, while
the horizon sweep fixes $K=16$.}
\label{tab:kh_sensitivity}

\renewcommand{\arraystretch}{1.10}
\setlength{\tabcolsep}{2.0pt}
\footnotesize

\begin{tabular*}{0.98\columnwidth}
{@{\extracolsep{\fill}}lccccc@{}}
\hline
\multicolumn{6}{l}{\textit{(a) Candidate number}} \\
\hline

Setting
& $K=1$
& $K=4$
& $K=8$
& $K=16$
& $K=32$ \\

\hline

Pred. Gain (\%)
& 0.000
& 12.050
& 16.585
& 20.384
& 23.289 \\

Runtime (ms)
& 1.995
& 1.966
& 1.992
& 1.814
& 1.816 \\

\hline
\end{tabular*}

\vspace{0.5em}

\begin{tabular*}{0.98\columnwidth}
{@{\extracolsep{\fill}}lcccc@{}}
\hline
\multicolumn{5}{l}{\textit{(b) Rollout horizon}} \\
\hline

Setting
& $H=1$
& $H=3$
& $H=5$
& $H=10$ \\

\hline

Task Pred. NRMSE
& 0.080
& 0.102
& 0.124
& 0.145 \\

Pred. Gain (\%)
& 7.340
& 11.594
& 14.978
& 20.259 \\

Runtime (ms)
& 1.559
& 1.629
& 1.690
& 1.830 \\

\hline
\end{tabular*}
\end{table}

Increasing $K$ improves predicted selection with diminishing returns. Pred. Gain rises from $12.05\%$ at $K=4$ to $20.38\%$ at $K=16$ and $23.29\%$ at $K=32$, while the incremental improvement falls from $4.54$ to $2.91$ percentage points for the $4\!\rightarrow\!8$ and $16\!\rightarrow\!32$ increases, respectively. The $K=16$ setting captures most of the benefit available at $K=32$, with similar batched offline runtimes.

Extending the prediction horizon from $H=1$ to $H=10$ raises Pred. Gain from $7.34\%$ to $20.26\%$, but Task Pred. NRMSE increases from $0.080$ to $0.145$ ($81.3\%$). These results support the selection tradeoff in Section~\ref{subsec:selection_theory}. More candidates increase the chance of a favorable choice, whereas longer predictions improve discrimination at the cost of accumulated model error.

\section{Real-World AUV Experiments}
\label{sec:real_experiments}

\subsection{Experimental Setup, Tracking Scenarios, Baselines, and Metrics}
\label{subsec:real_setup}

Real-AUV experiments used two vehicles in an indoor water tank. The leader followed prescribed maneuvers, while the follower tracked autonomously using its onboard camera, IMU, and Jetson AGX Orin. Motion capture supported repeatable leader motion and ground-truth evaluation but provided no feedback to autonomous follower controllers. A tether enabled monitoring and data transmission without supplying external state measurements.

\begin{figure*}[t]
    \centering

    \subfloat[Longitudinal Double Shuttle\label{fig:snapshot_t1}]{
        \begin{minipage}{0.98\textwidth}
            \centering
            \includegraphics[width=0.160\linewidth]{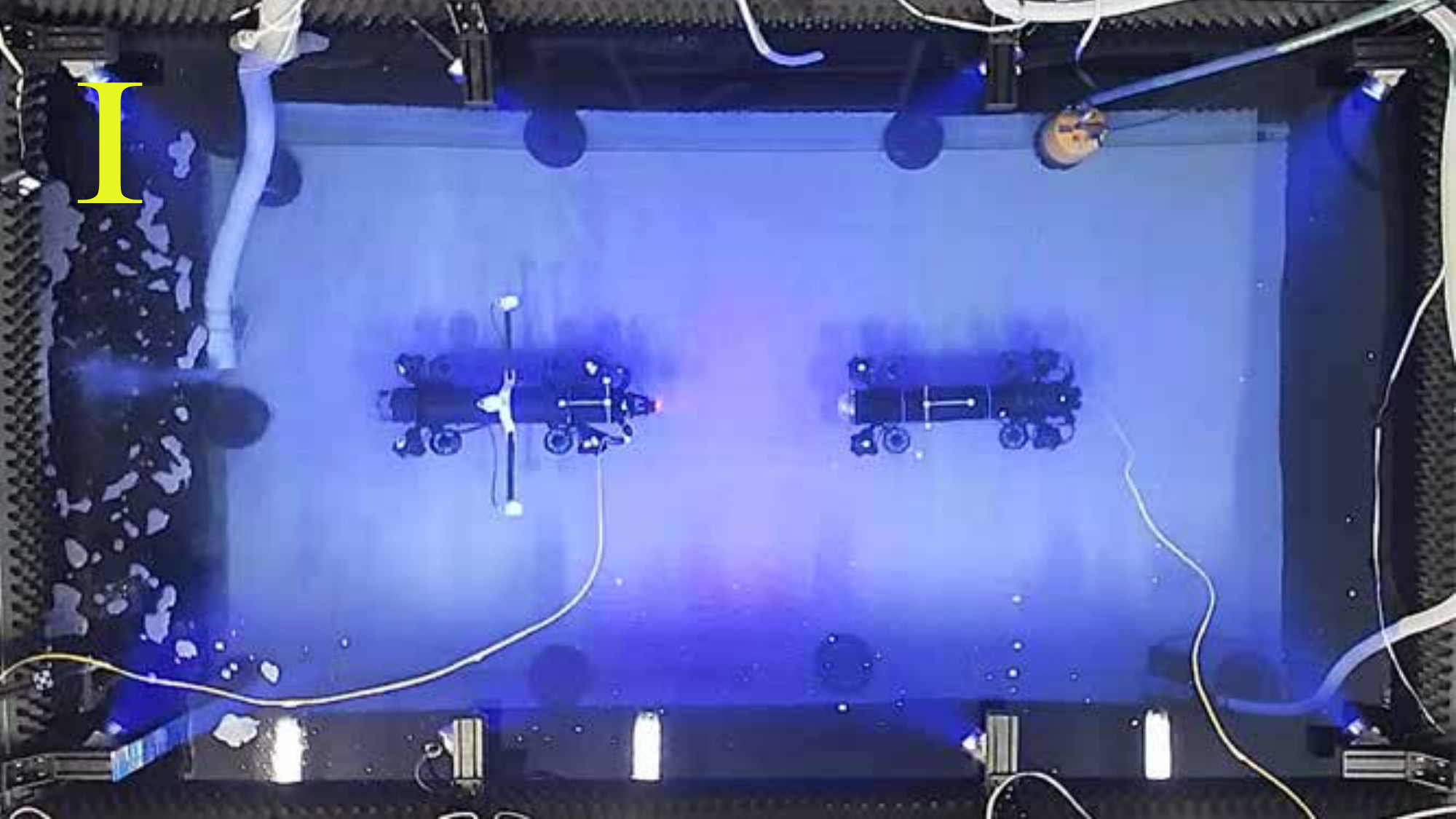}
            \hfill
            \includegraphics[width=0.160\linewidth]{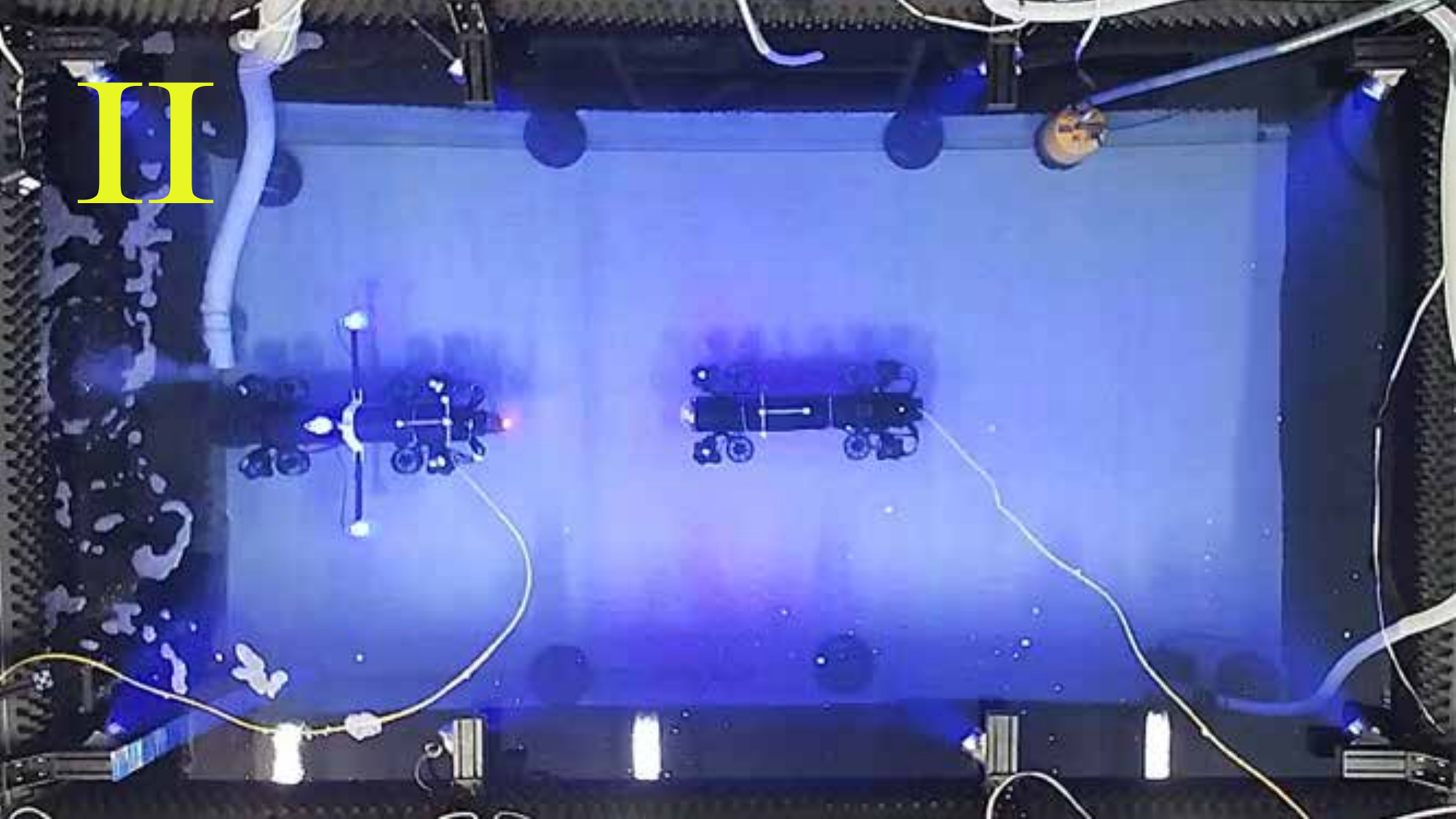}
            \hfill
            \includegraphics[width=0.160\linewidth]{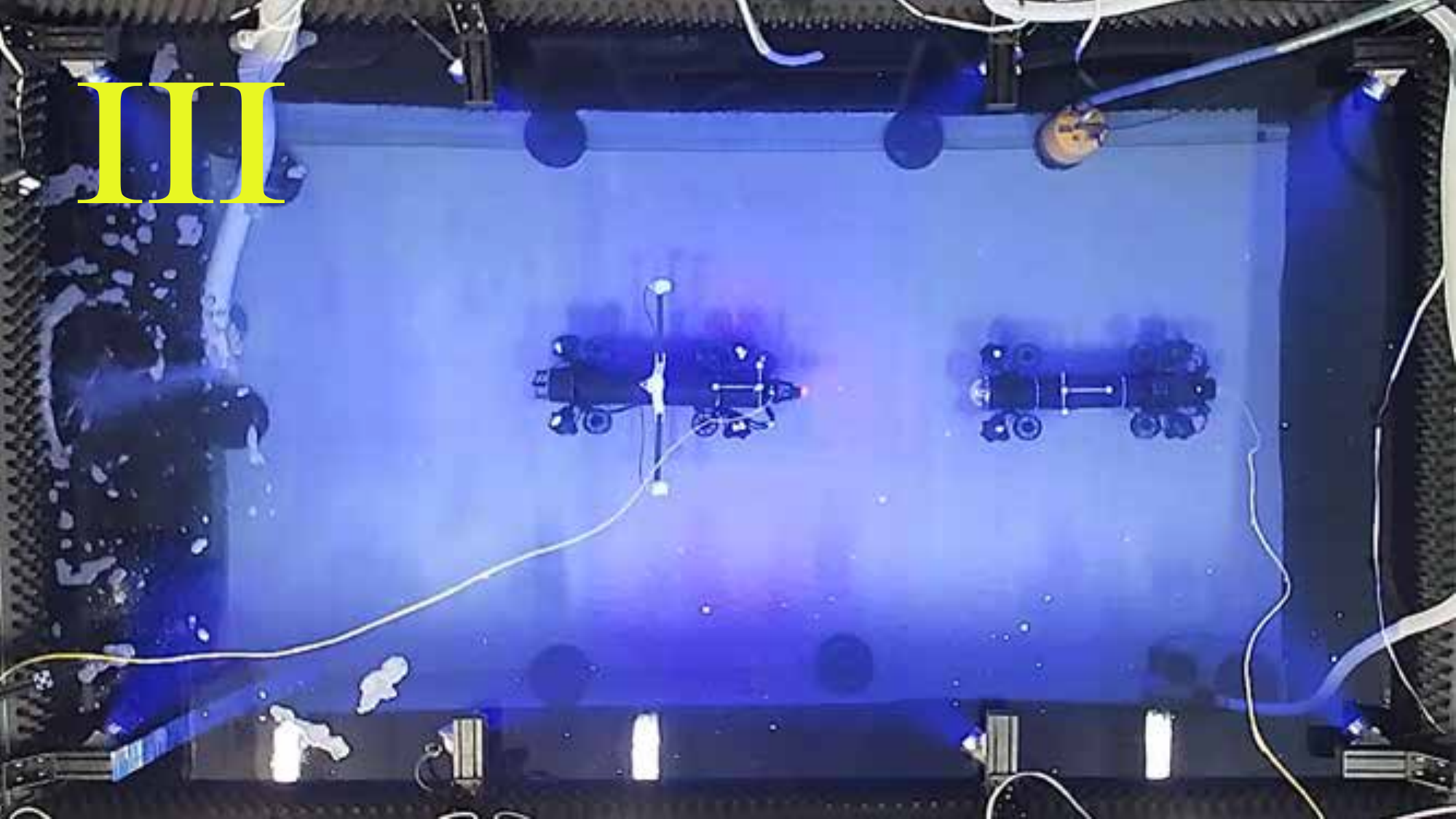}
            \hfill
            \includegraphics[width=0.160\linewidth]{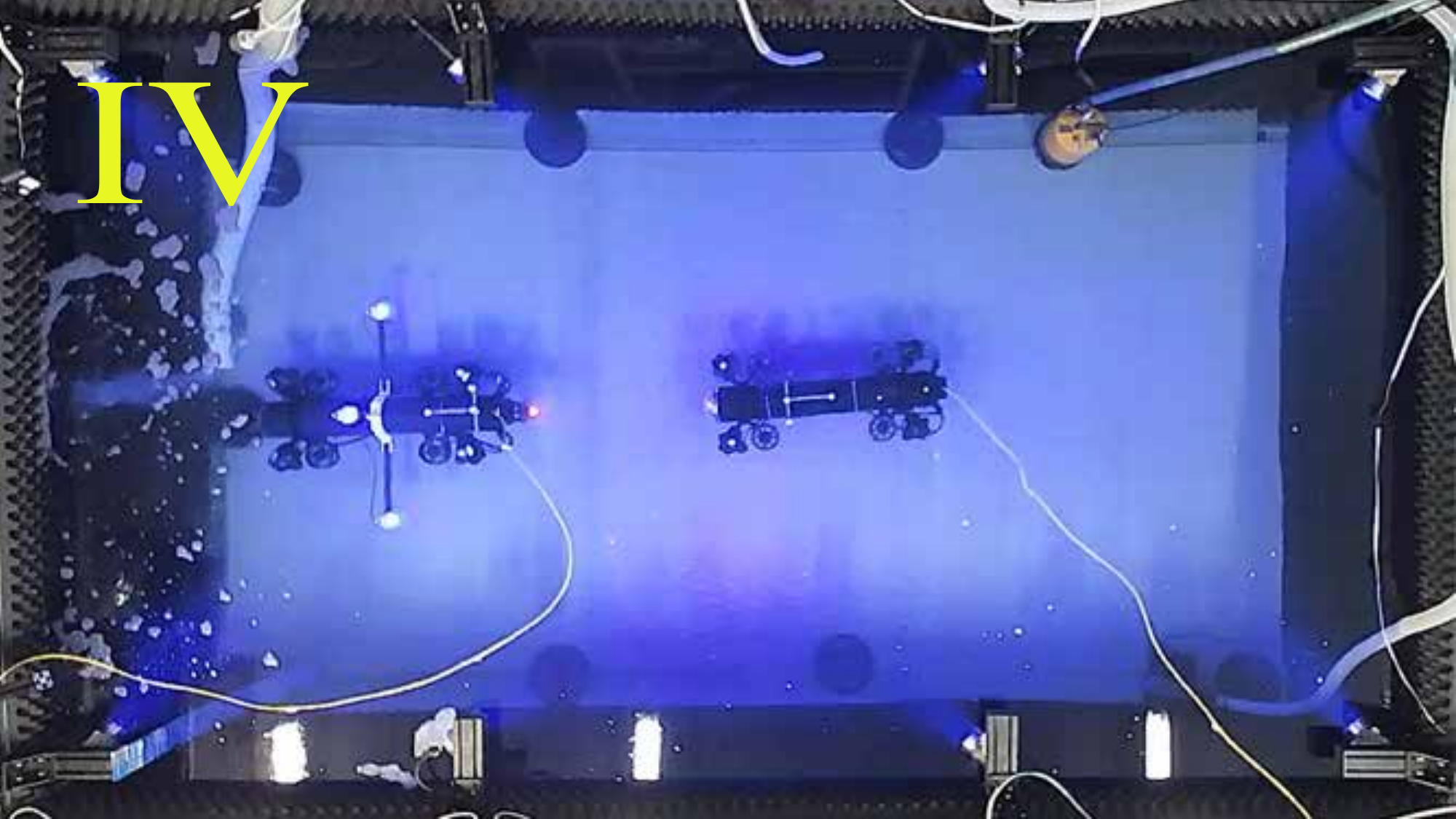}
            \hfill
            \includegraphics[width=0.160\linewidth]{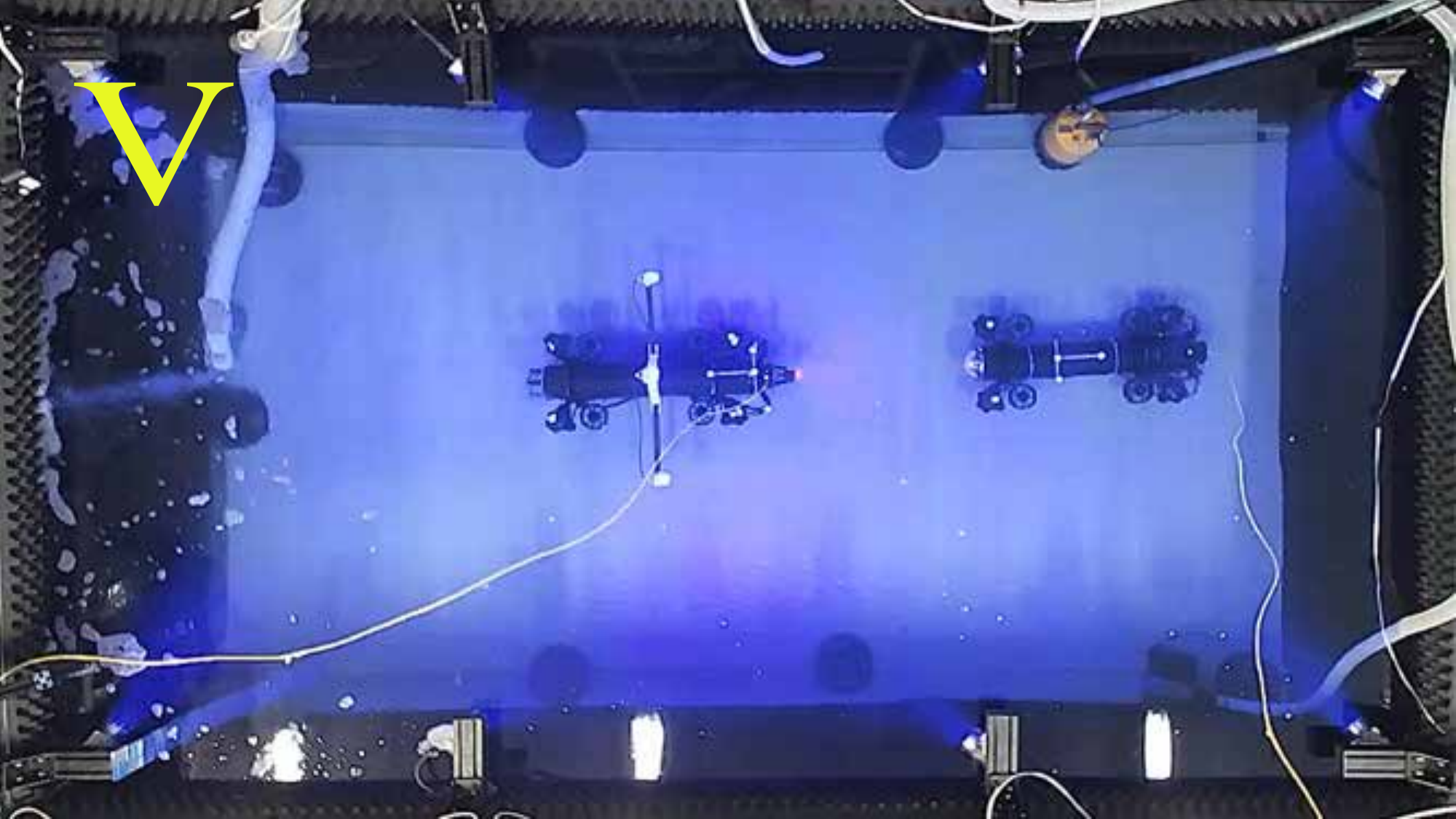}
            \hfill
            \includegraphics[width=0.160\linewidth]{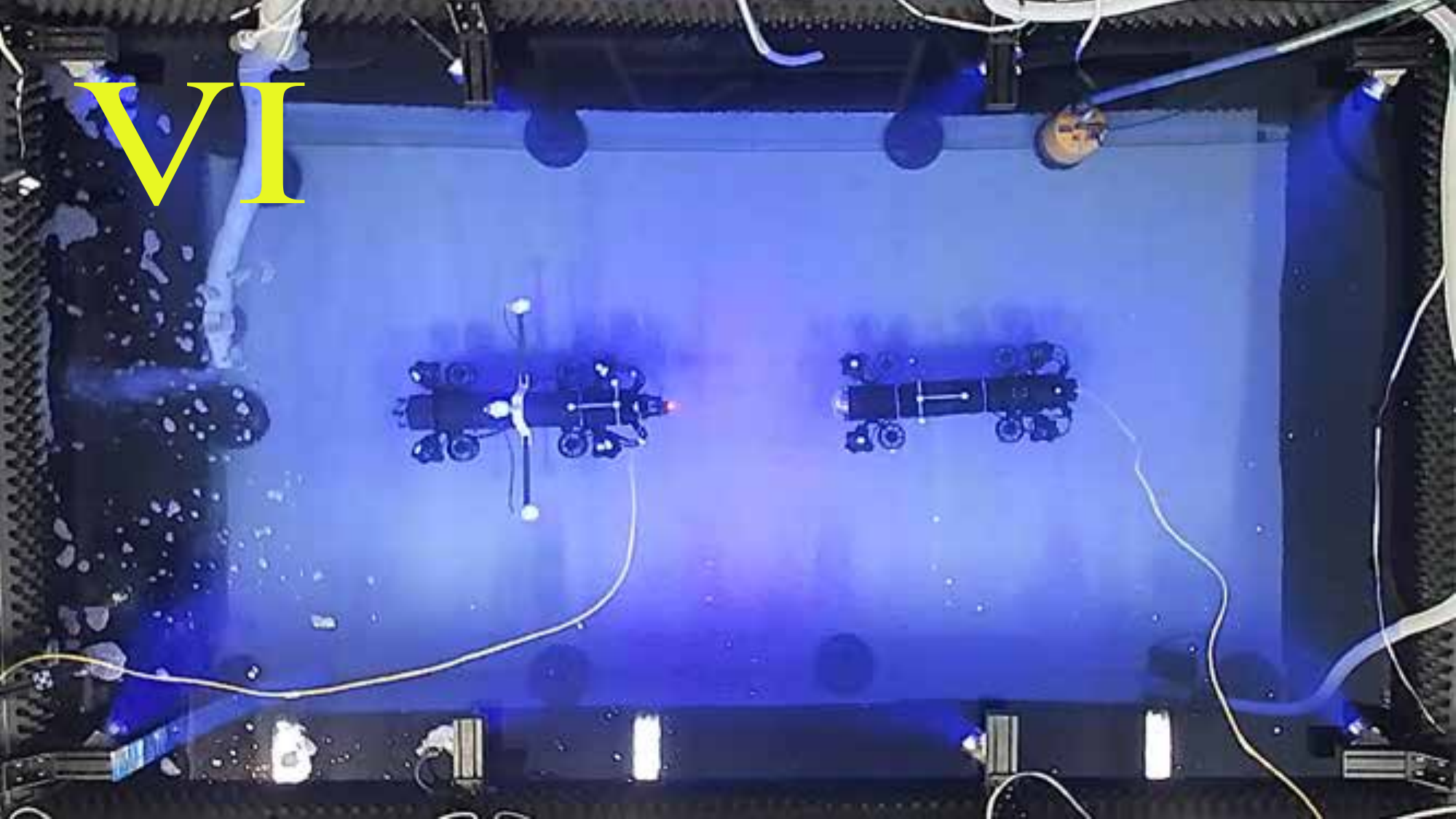}
        \end{minipage}
    }

    \vspace{-0.6em}

    \subfloat[Lateral Double Shuttle\label{fig:snapshot_t2}]{
        \begin{minipage}{0.98\textwidth}
            \centering
            \includegraphics[width=0.160\linewidth]{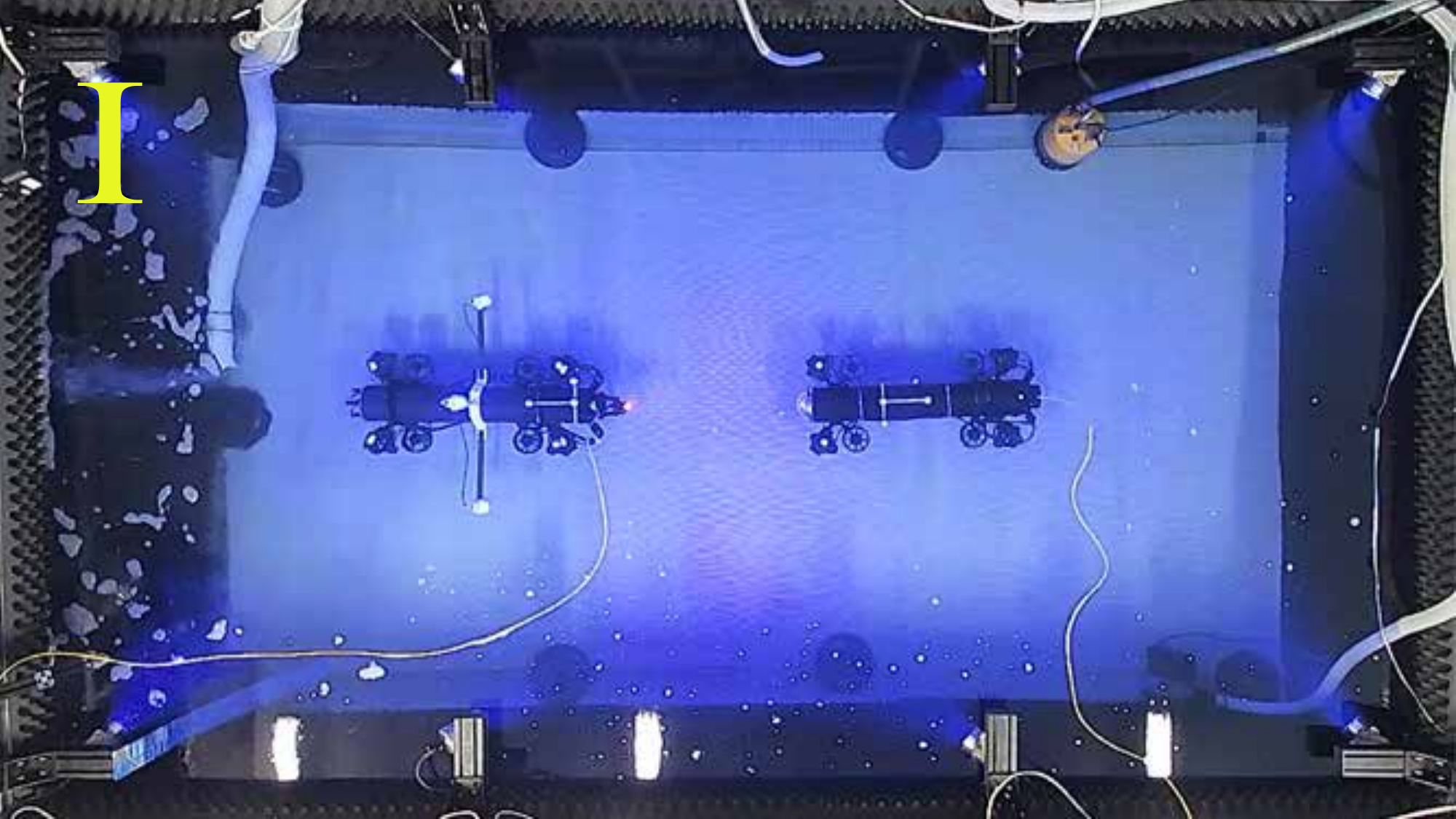}
            \hfill
            \includegraphics[width=0.160\linewidth]{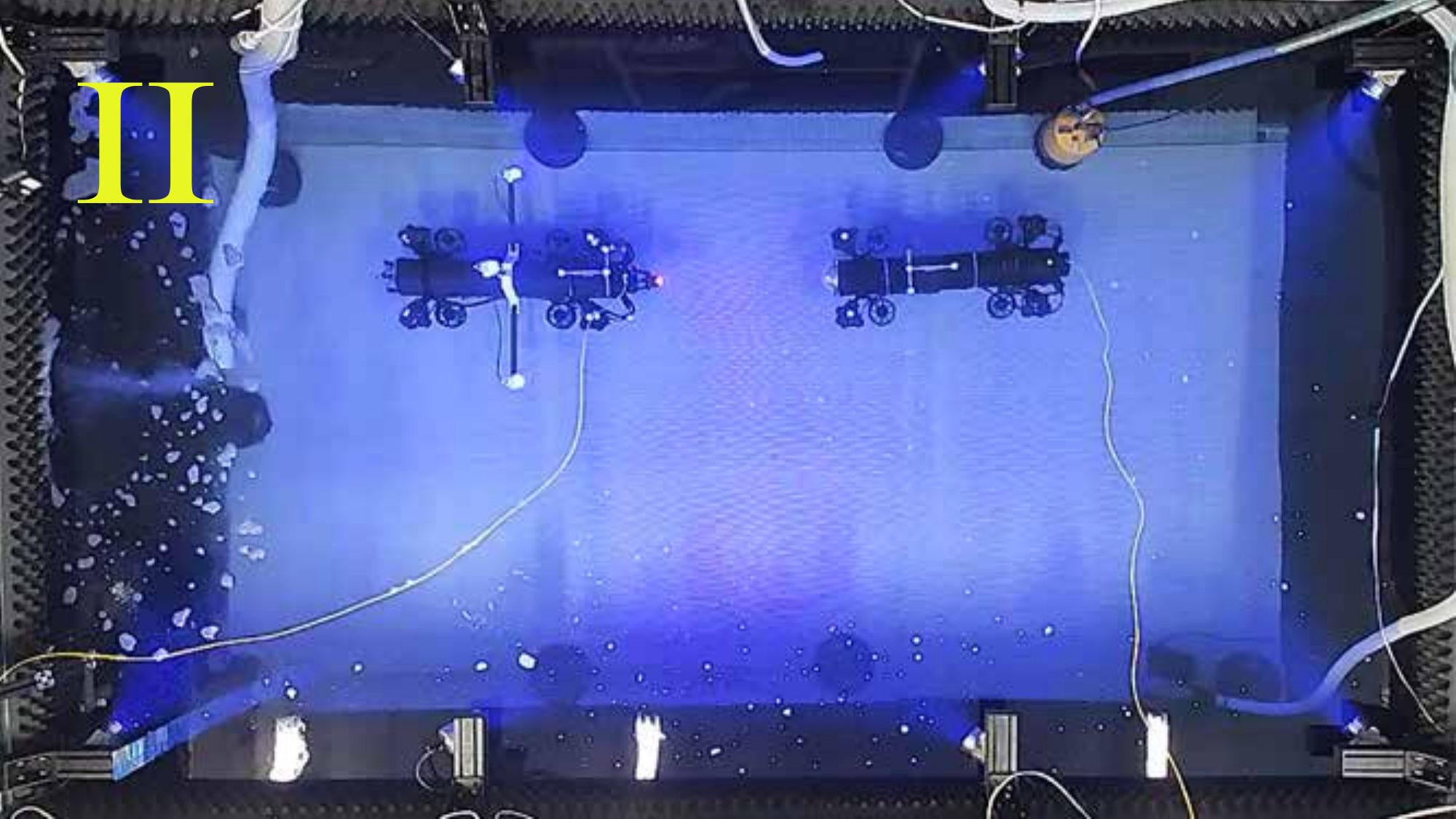}
            \hfill
            \includegraphics[width=0.160\linewidth]{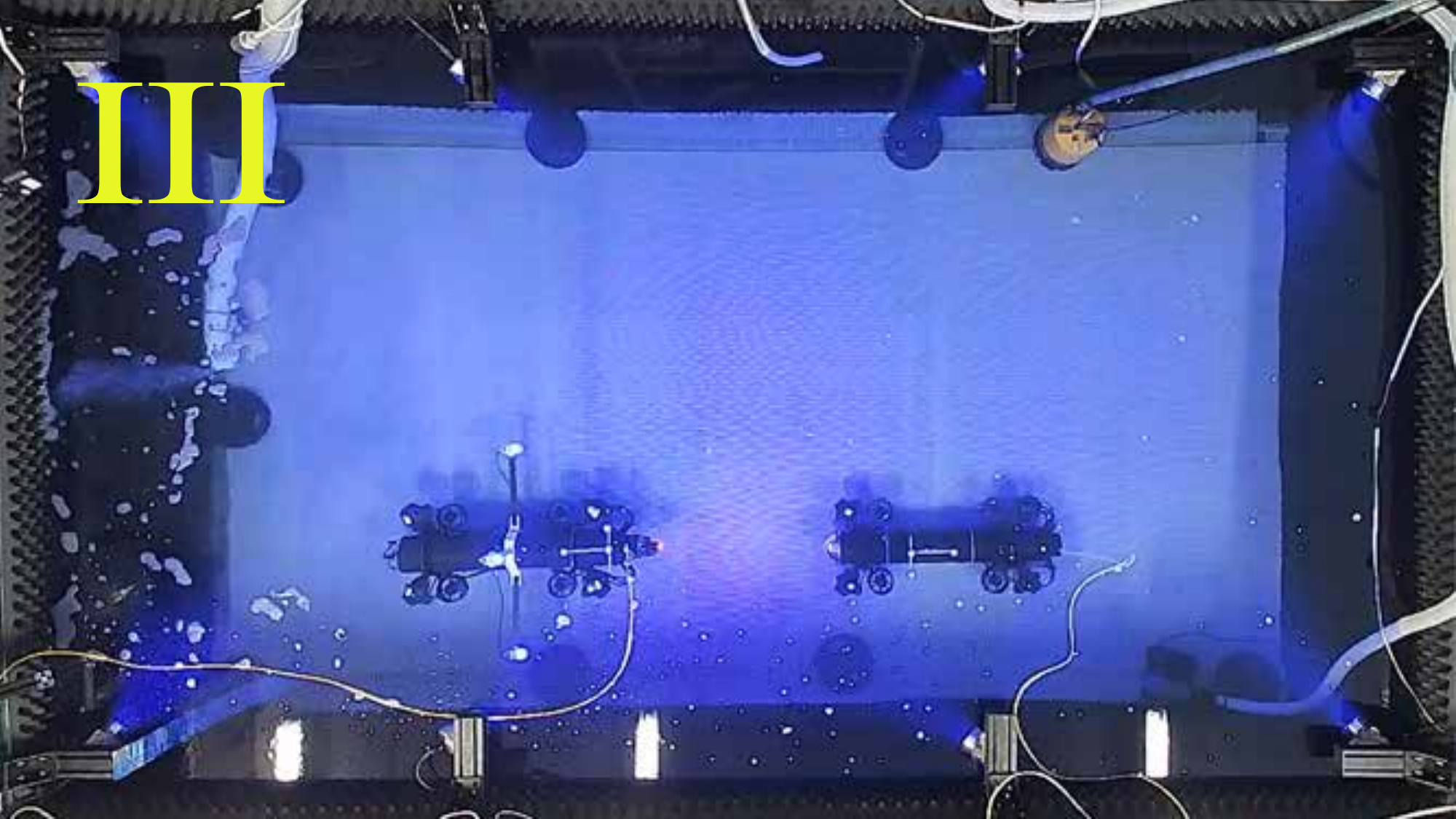}
            \hfill
            \includegraphics[width=0.160\linewidth]{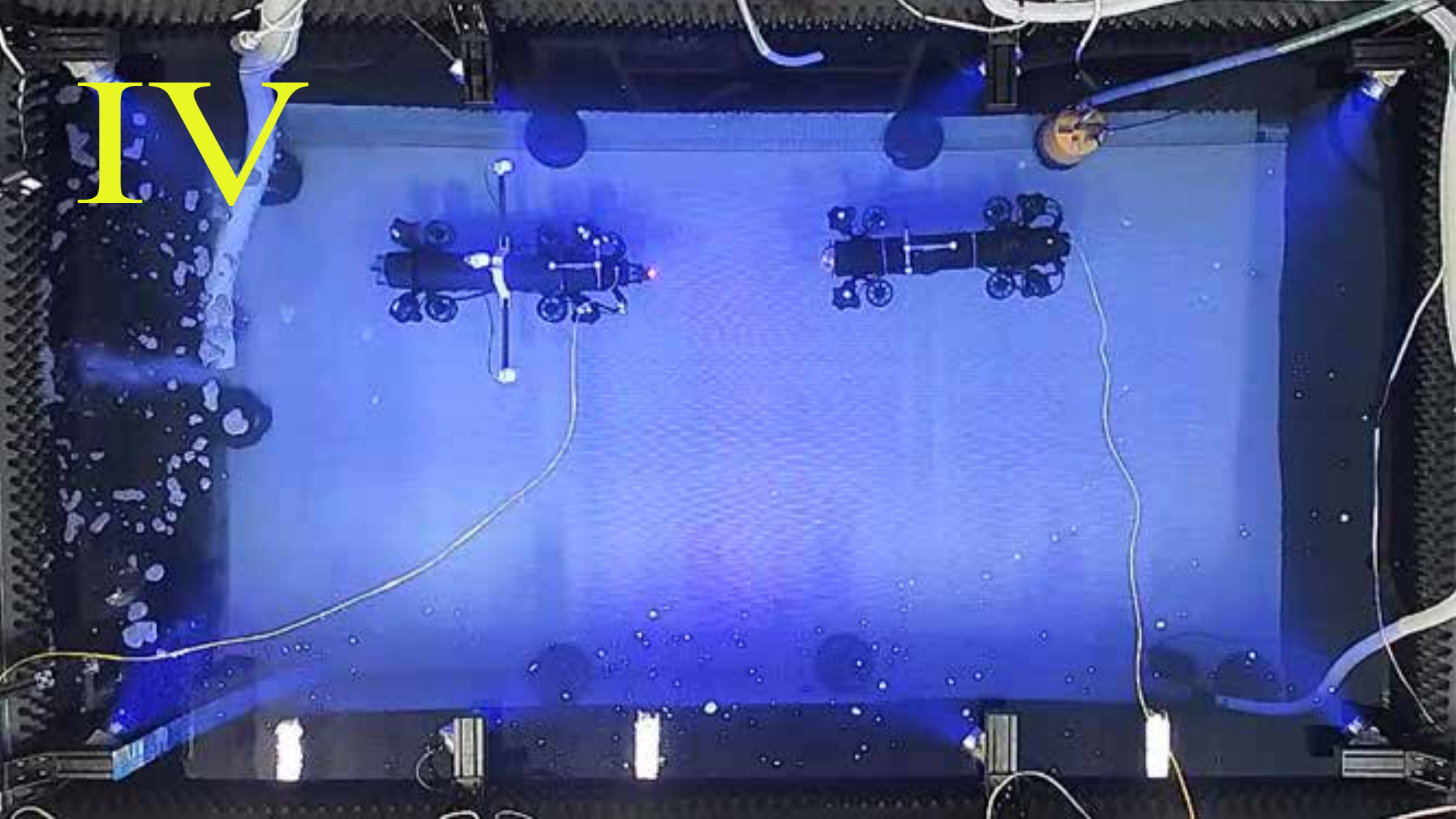}
            \hfill
            \includegraphics[width=0.160\linewidth]{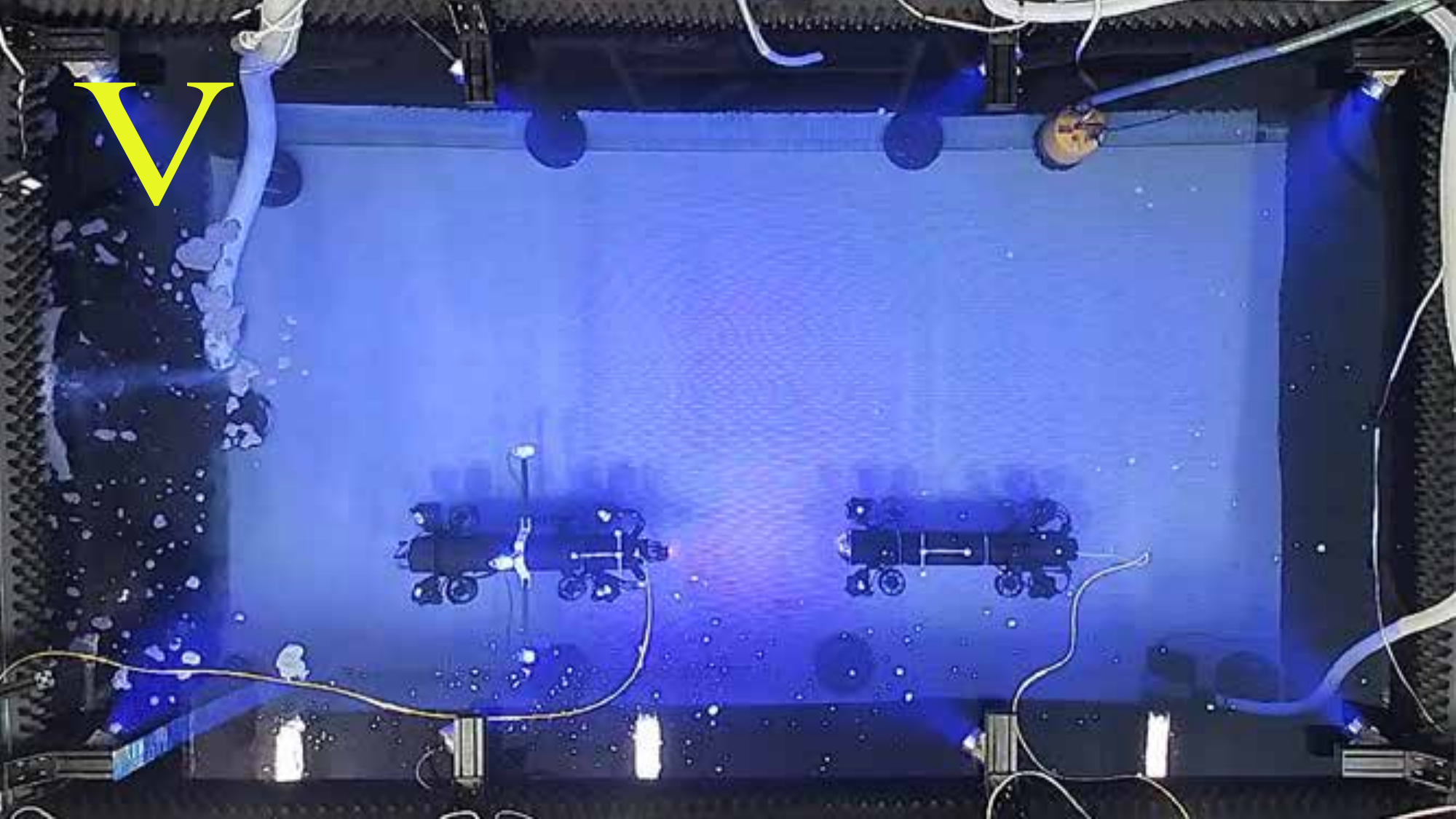}
            \hfill
            \includegraphics[width=0.160\linewidth]{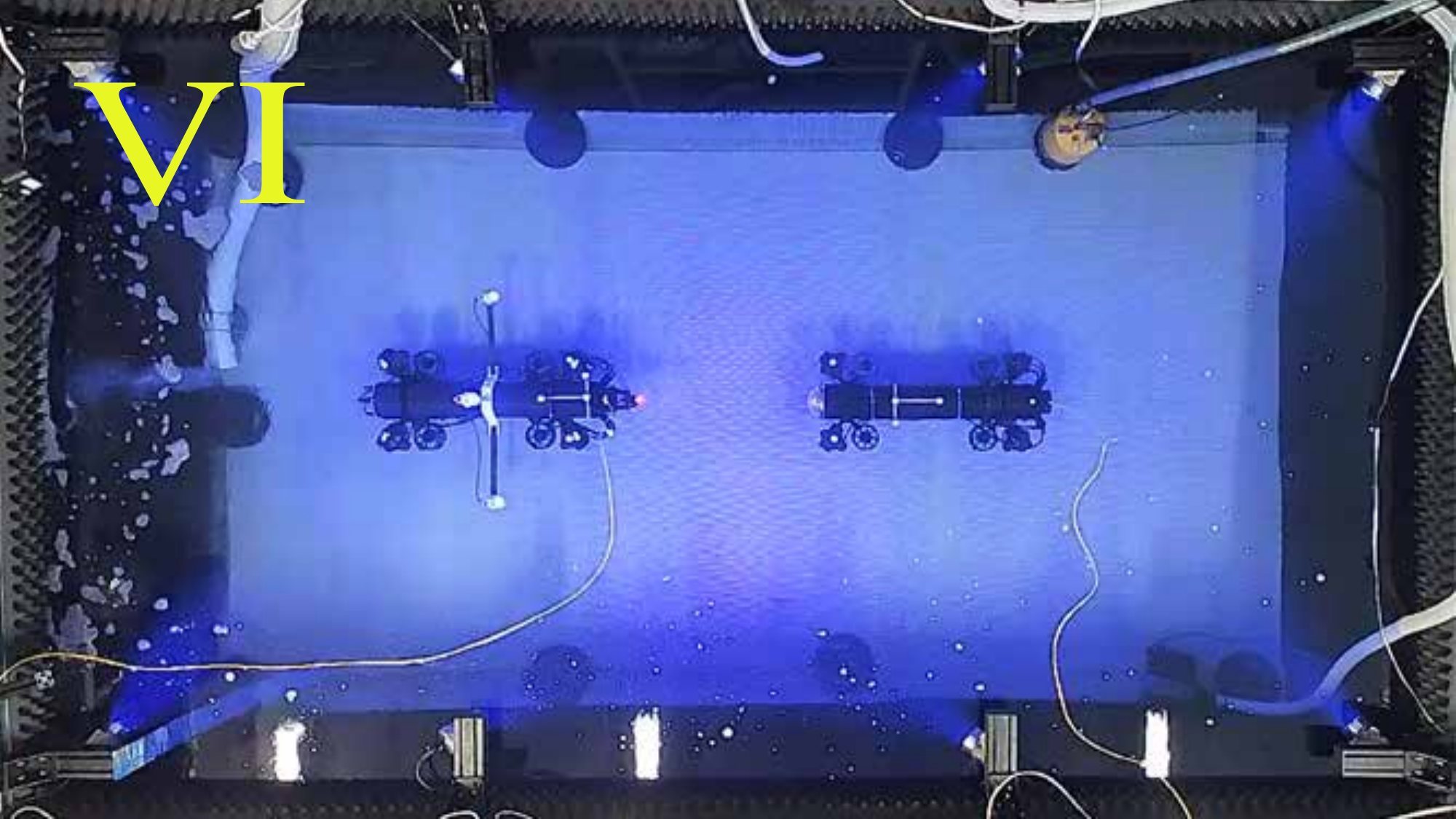}
        \end{minipage}
    }

    \vspace{-0.6em}

    \subfloat[Yaw Double Sweep\label{fig:snapshot_t3}]{
        \begin{minipage}{0.98\textwidth}
            \centering
            \includegraphics[width=0.160\linewidth]{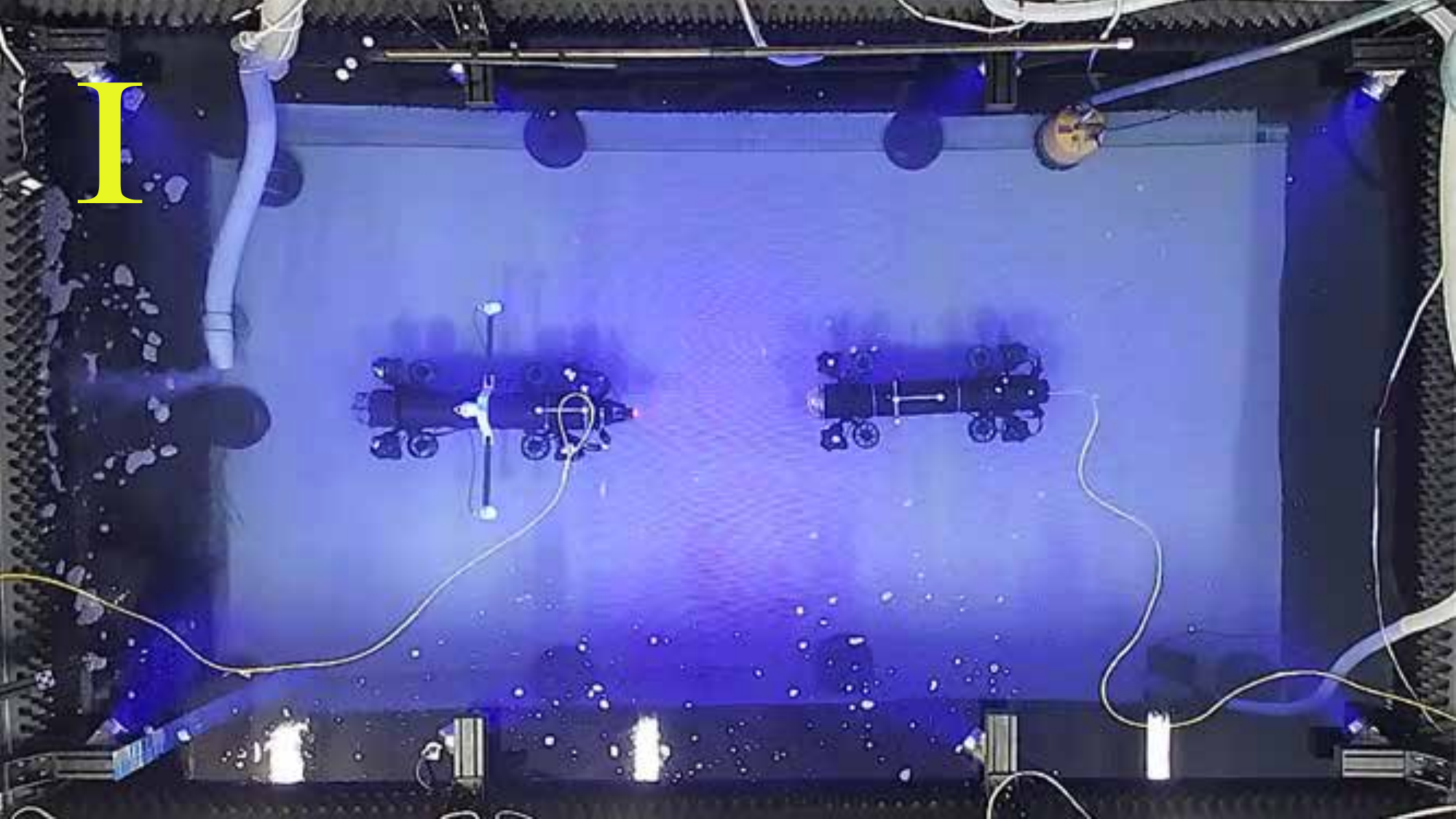}
            \hfill
            \includegraphics[width=0.160\linewidth]{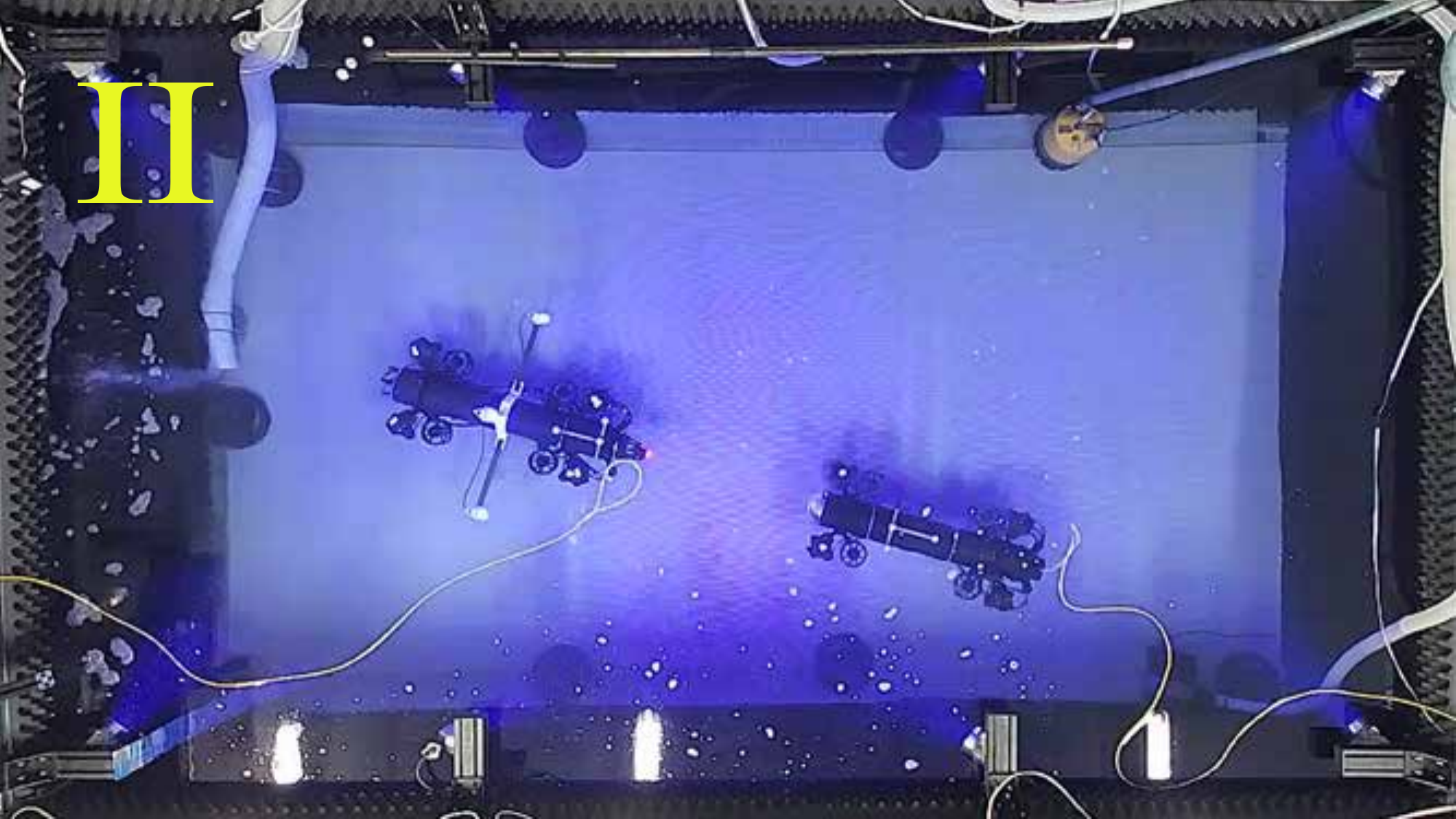}
            \hfill
            \includegraphics[width=0.160\linewidth]{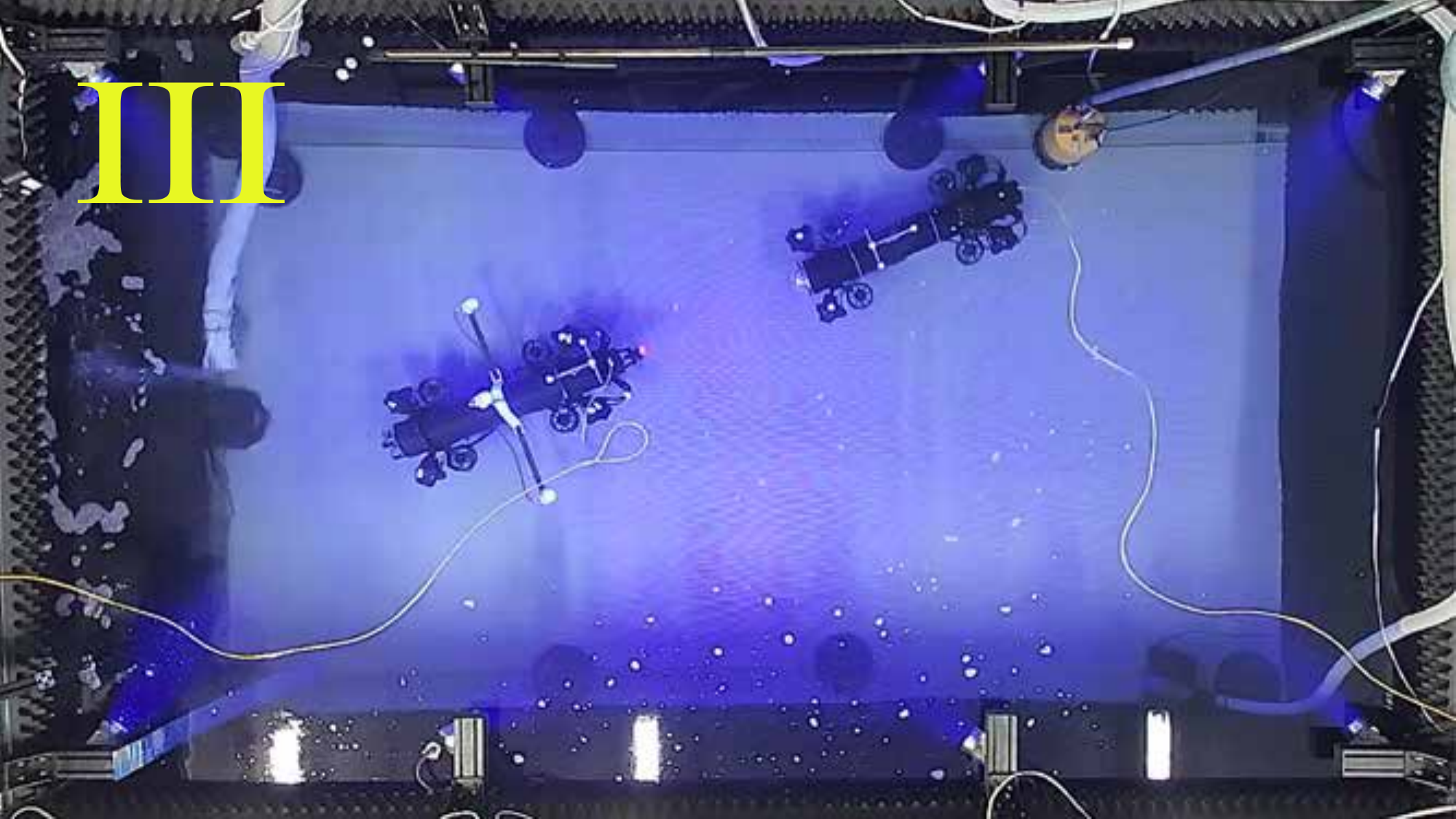}
            \hfill
            \includegraphics[width=0.160\linewidth]{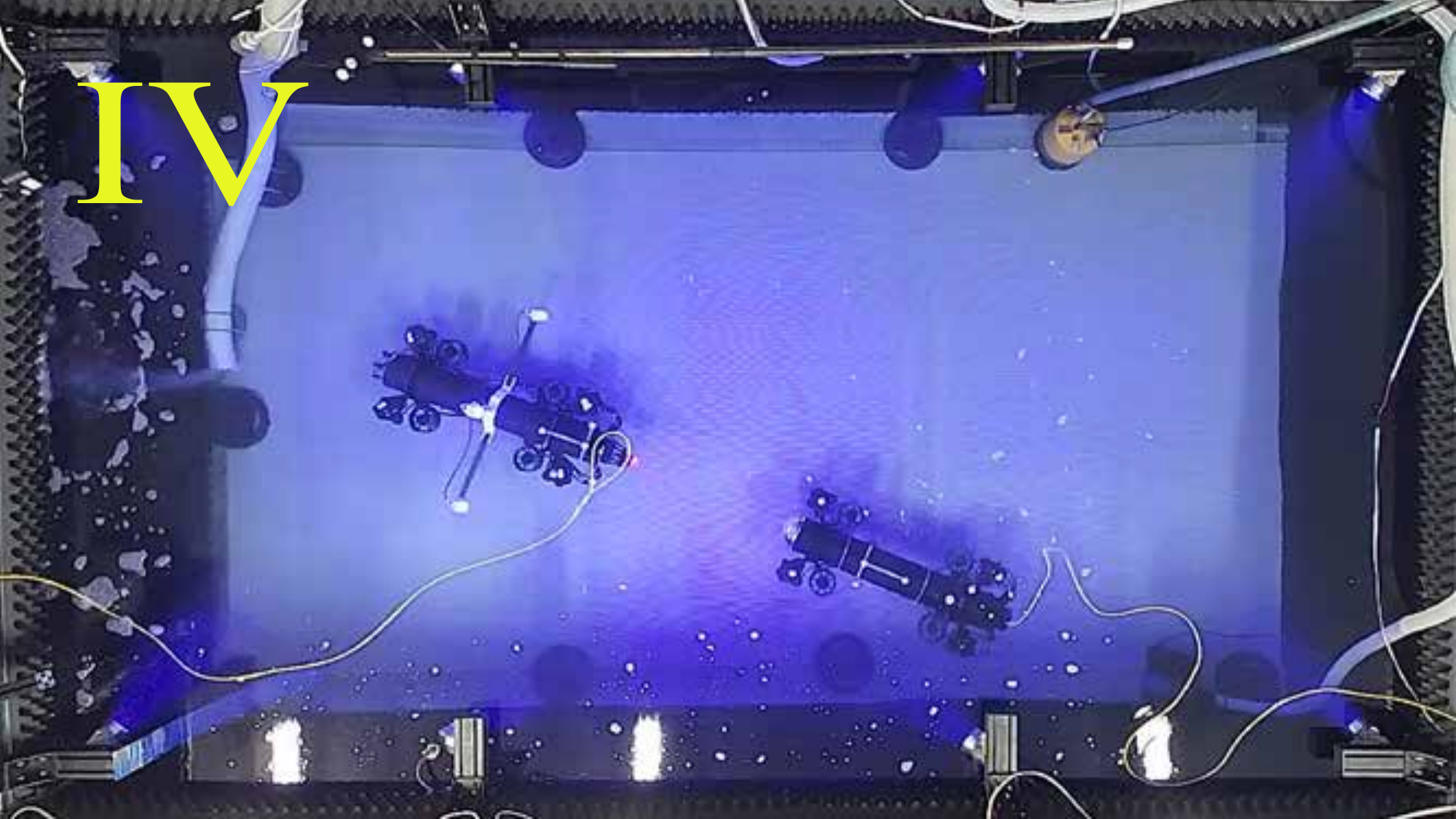}
            \hfill
            \includegraphics[width=0.160\linewidth]{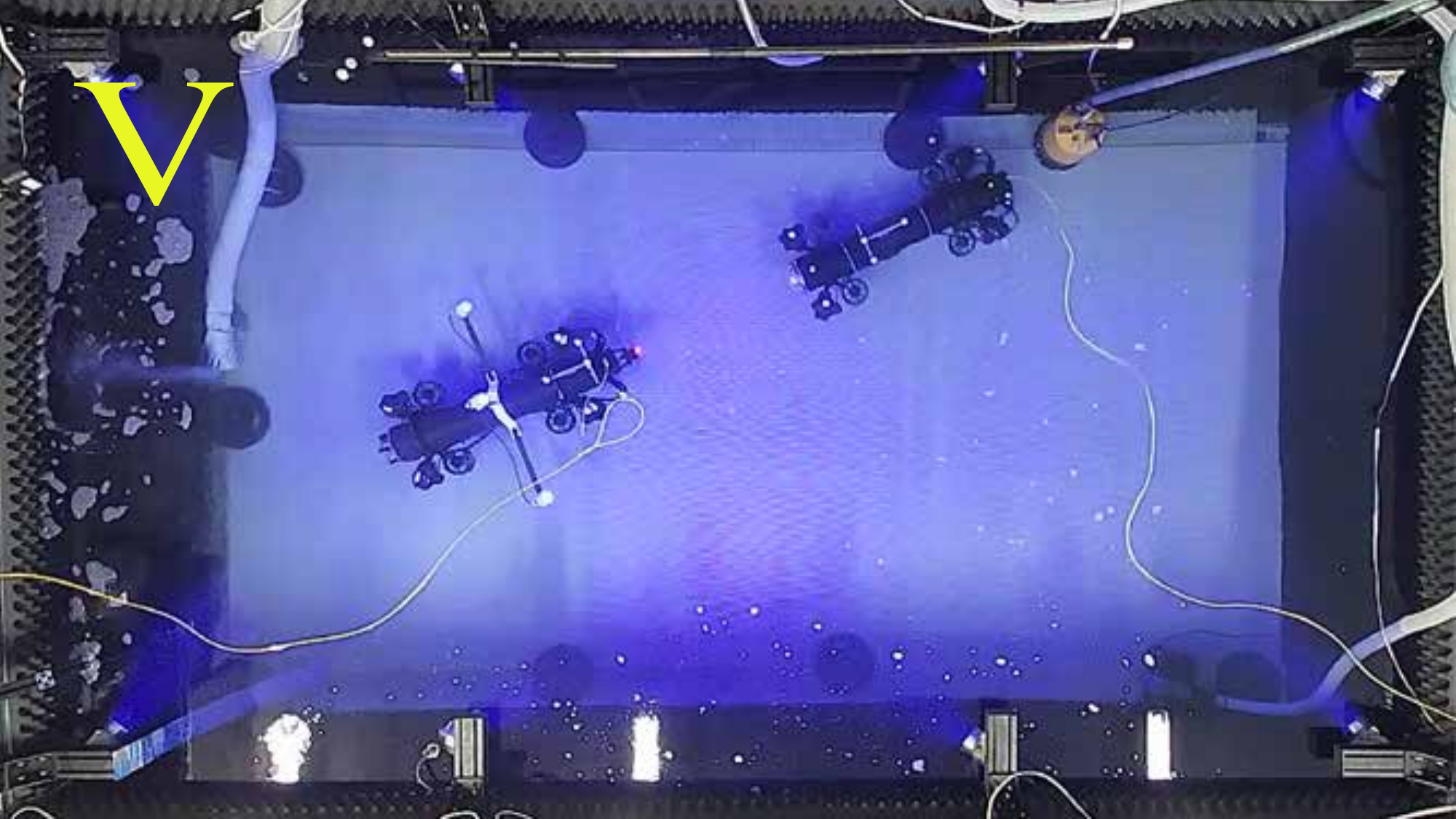}
            \hfill
            \includegraphics[width=0.160\linewidth]{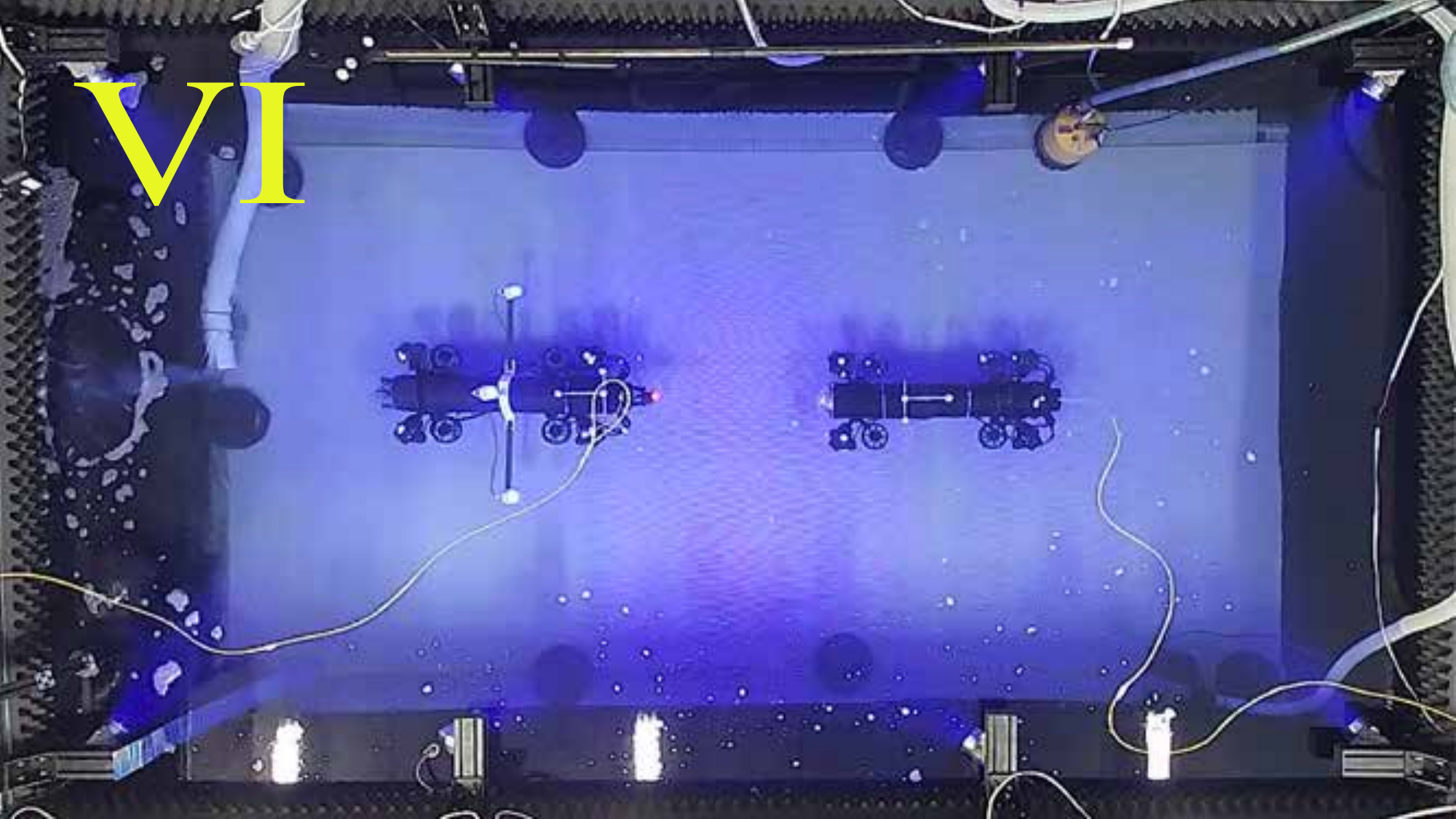}
        \end{minipage}
    }

    \vspace{-0.6em}

    \subfloat[Four-Corner Coupled\label{fig:snapshot_t4}]{
        \begin{minipage}{0.98\textwidth}
            \centering
            \includegraphics[width=0.160\linewidth]{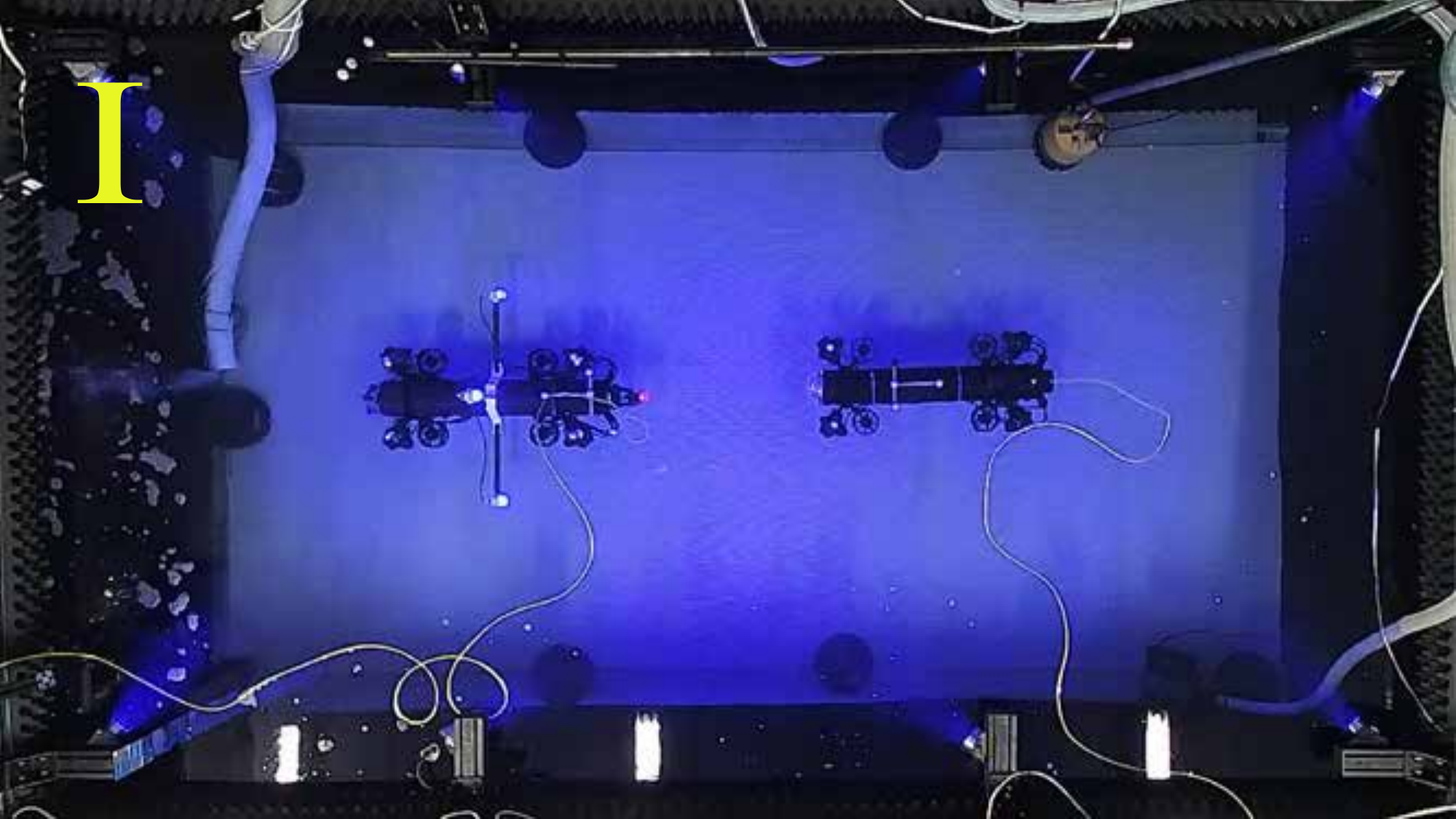}
            \hfill
            \includegraphics[width=0.160\linewidth]{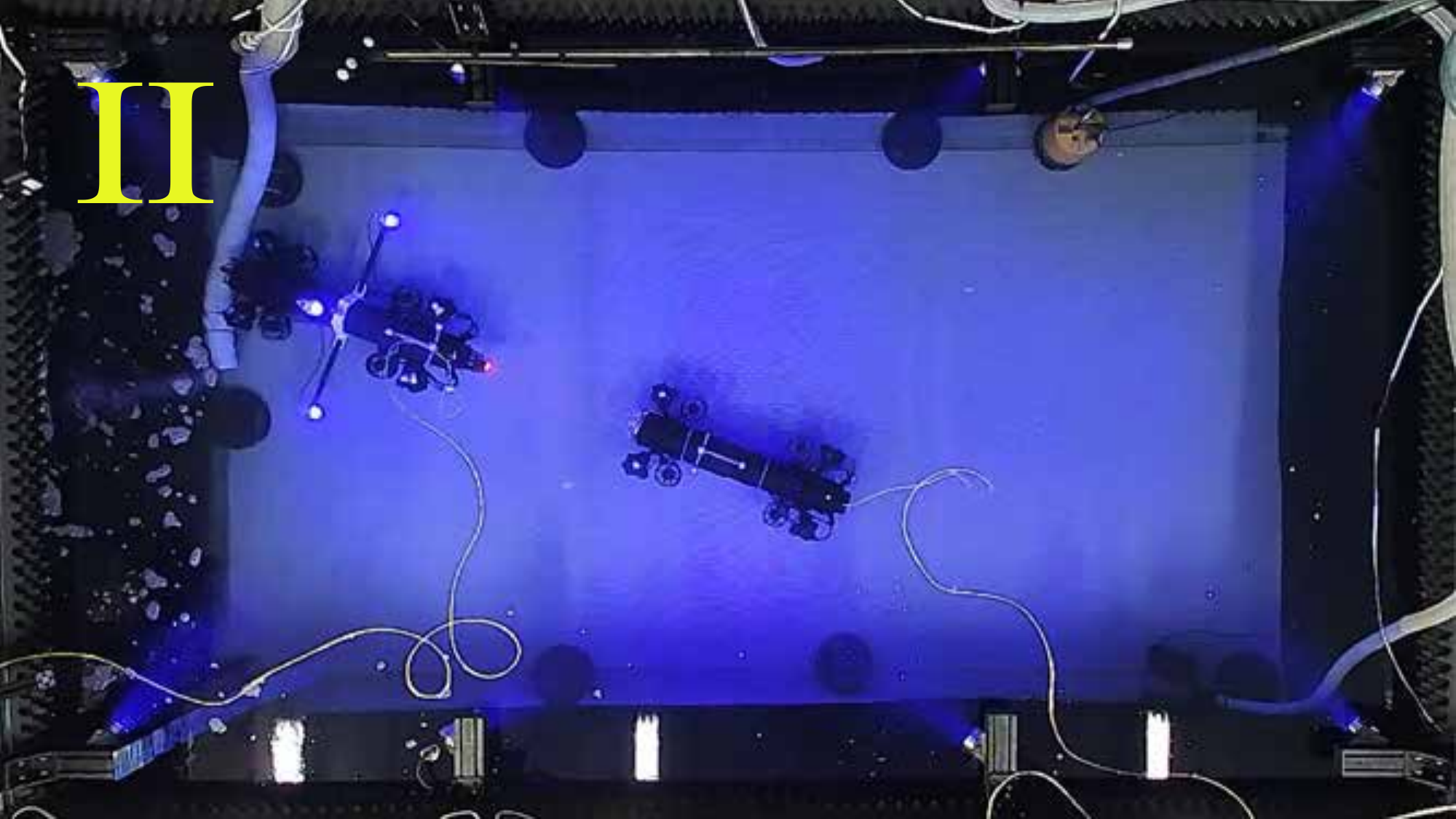}
            \hfill
            \includegraphics[width=0.160\linewidth]{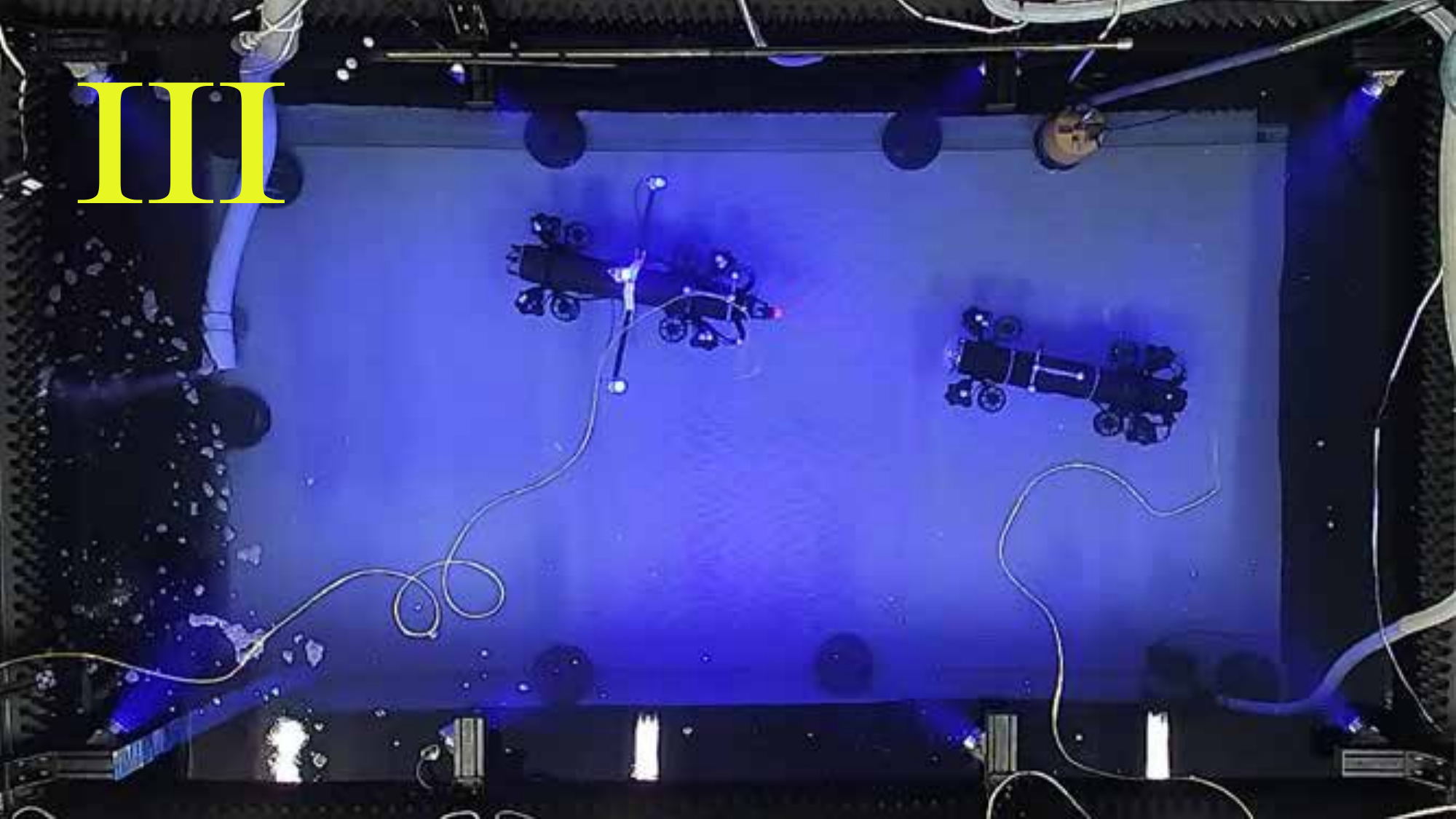}
            \hfill
            \includegraphics[width=0.160\linewidth]{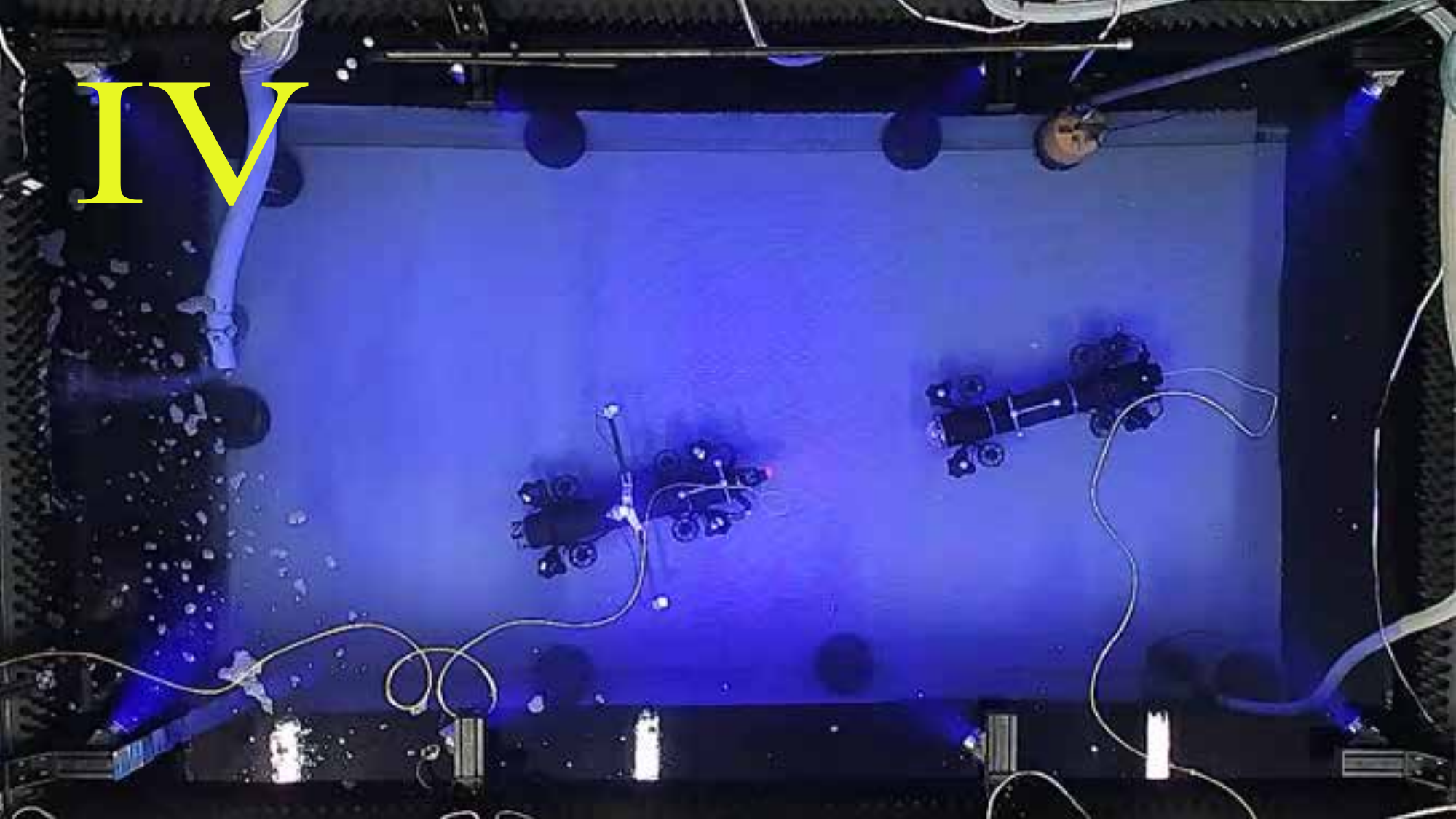}
            \hfill
            \includegraphics[width=0.160\linewidth]{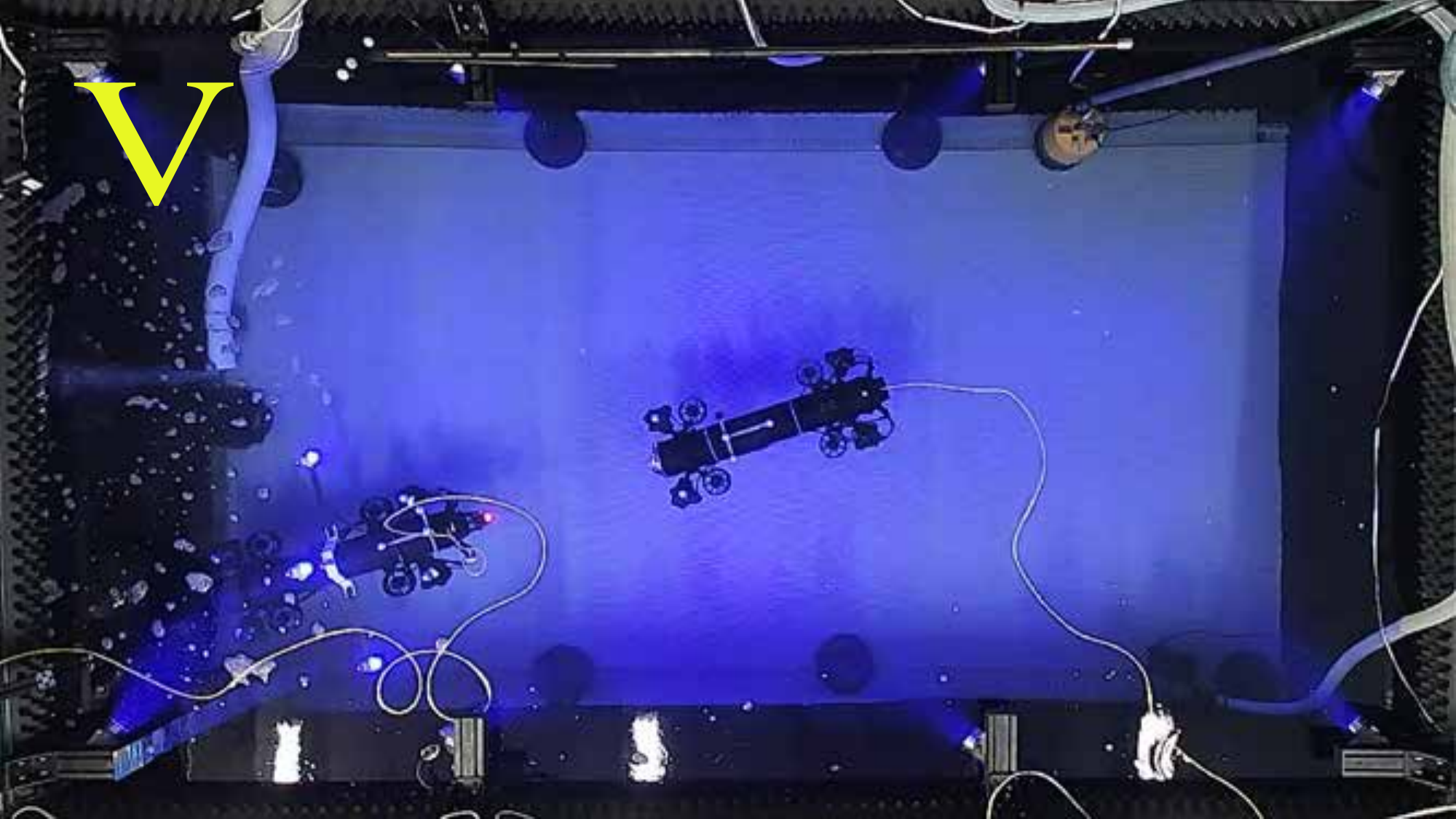}
            \hfill
            \includegraphics[width=0.160\linewidth]{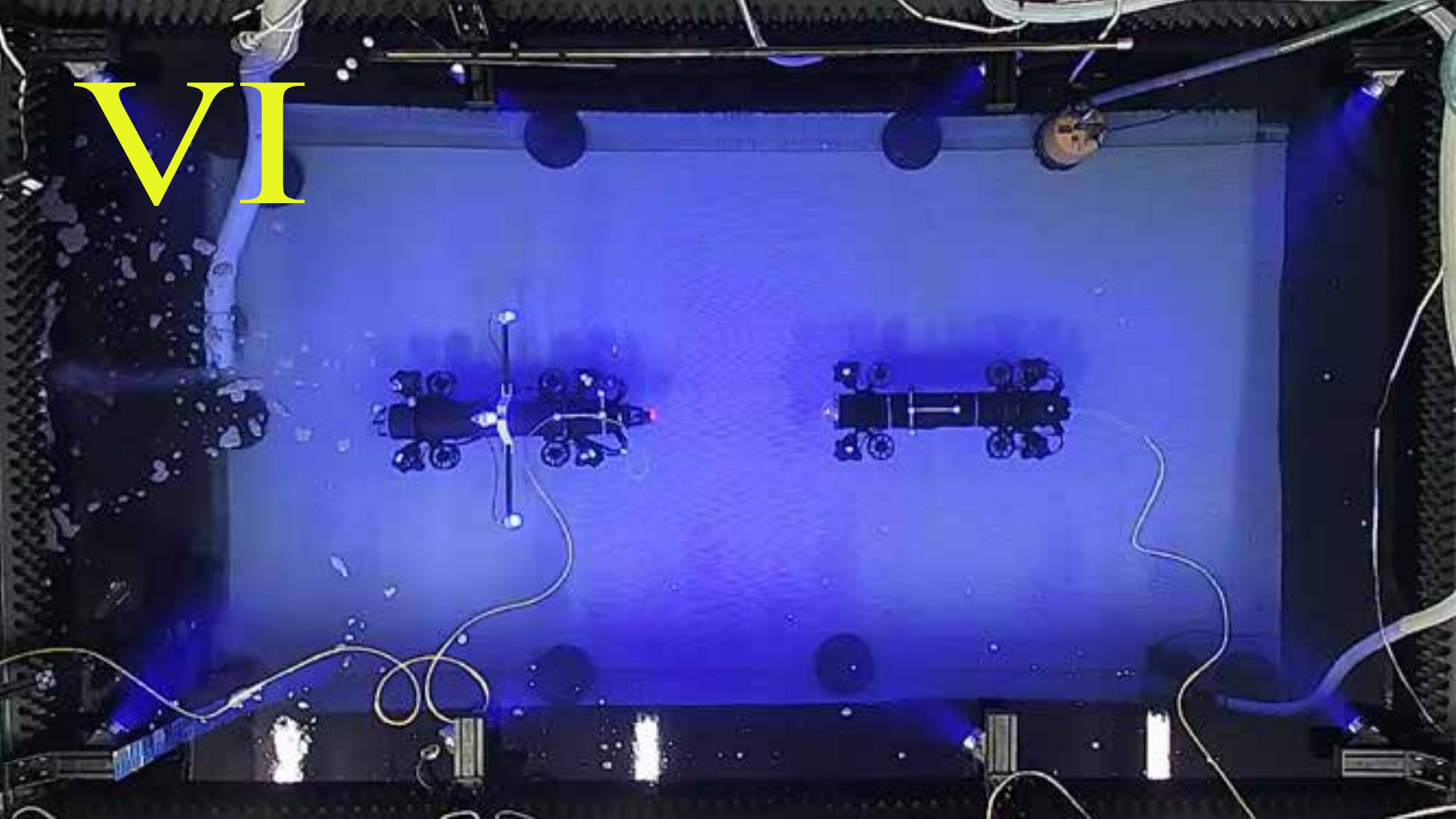}
        \end{minipage}
    }

    \caption{Representative third-person snapshots of FATHOM tracking during challenging (a) T1 longitudinal double shuttle, (b) T2 lateral double shuttle, (c) T3 yaw double sweep, and (d) T4 four-corner coupled motion. Each row shows the initial state, four intermediate stages, and the final state.}
    \label{fig:real_tracking_snapshots}
\end{figure*}

The target $\bm{y}^{*}=[1.5,0,0,0,0,0]^{\top}$ in~\eqref{eq:desired_relative_state} specifies relative position, follower roll and pitch, and relative yaw. Four maneuvers are evaluated. T1 (Longitudinal Double Shuttle), T2 (Lateral Double Shuttle), and T3 (Yaw Double Sweep) excite individual translation or yaw axes, with reference offsets
\begin{equation}
\begin{aligned}
\delta_j:\quad 0&\rightarrow +A_j\rightarrow -A_j\rightarrow +A_j\\
&\rightarrow -A_j\rightarrow 0,\qquad j\in\{x,y,\psi\},
\end{aligned}
\end{equation}
where $A_j$ denotes the corresponding motion amplitude. T4 (Four-Corner Coupled) follows a planar waypoint sequence
\begin{equation}
\begin{aligned}
(0,0)&\rightarrow(A_x,A_y)\rightarrow(-A_x,A_y)\\
&\rightarrow(-A_x,-A_y)\rightarrow(A_x,-A_y)
\rightarrow(0,0),
\end{aligned}
\end{equation}
with coordinated depth and yaw variations. Transitions use quintic minimum-jerk interpolation and waypoint holds. Challenging regimes retain the nominal waypoints but shorten transition and hold durations. Three trials per controller, maneuver, and regime yield 24 trials per controller. Fig.~\ref{fig:real_tracking_snapshots} shows representative challenging FATHOM sequences.

Six controllers were evaluated. \textbf{Mocap-PID} uses privileged motion-capture relative-pose feedback and serves as an expert reference for assessing the gap to externally observed control. Three pose-based autonomous baselines share the AMR-Pose estimator~\cite{sha2026amrpose}, a strong 6-DoF perception frontend designed for sparse and partially corrupted LED observations. \textbf{Pose-PID} uses conventional PID feedback, while \textbf{Pose-ISMC} employs integral sliding-mode control as a robust feedback baseline. \textbf{Pose-DP} conditions a diffusion policy directly on the estimated six-dimensional task state, permitting comparison between explicit-pose and predictive-belief conditioning. \textbf{DAP-only} uses the same belief-conditioned diffusion prior as FATHOM but executes one generated candidate without predictive selection ($K=1$). \textbf{FATHOM} combines predictive belief, multi-candidate DAP generation, and world-model predictive selection with $K=16$, selected from the observed performance--computation tradeoff.

Performance was assessed using motion-capture ground truth. The primary metric is the six-dimensional task-state NRMSE, computed with the fixed normalization defined in Sec.~\ref{subsec:data_processing}. We also examine trial completion, representative state responses, and tracking and recovery under naturally occurring partial marker visibility. A trial is considered successful if the prescribed maneuver is completed without losing tracking.

\subsection{Comparison with Pose-Based Autonomous Control}
\label{subsec:pose_baseline_comparison}

Table~\ref{tab:real_tracking_comparison} summarizes the six-controller comparison. Under nominal motion, FATHOM attains a mean NRMSE of $0.084$, improving upon Pose-PID and Pose-ISMC by $31.1\%$ and $27.0\%$, respectively, while matching the privileged Mocap-PID mean. On nominal T1 and T4, FATHOM also yields slightly lower errors than Mocap-PID ($0.073$ versus $0.075$ and $0.094$ versus $0.101$). Under challenging motion, its mean NRMSE of $0.164$ is $43.3\%$ and $34.7\%$ lower than those of Pose-PID and Pose-ISMC. On the coupled T4 maneuver, FATHOM achieves $0.179$ versus $0.372$ and $0.314$, respectively. Pose-PID and Pose-ISMC fail in three and two of their three challenging T4 trials, with overall challenging completion rates of $50.0\%$ and $66.7\%$. FATHOM completes all 12 challenging trials.

\begin{figure}[t]
    \centering
    \includegraphics[width=0.99\columnwidth]{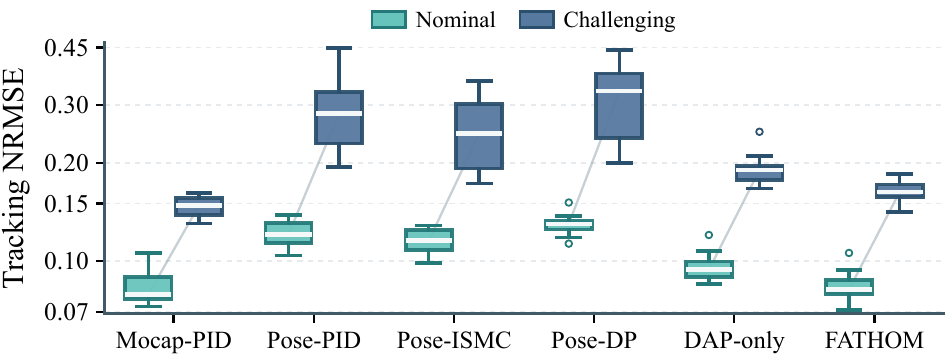}
    \caption{Trial-level tracking NRMSE across six controllers under nominal and challenging motion. Each box summarizes 12 trials from four maneuvers and three repetitions. Mocap-PID uses privileged motion-capture feedback.}
    \label{fig:tracking_nrmse_boxplot}
\end{figure}

Fig.~\ref{fig:tracking_nrmse_boxplot} confirms greater trial-level consistency. Under challenging motion, FATHOM's median NRMSE and interquartile range are $0.163$ and $0.014$, compared with $0.283$ and $0.100$ for Pose-PID and $0.245$ and $0.110$ for Pose-ISMC. Its narrower distribution indicates less variation across maneuvers and repetitions, showing that the gain is not driven by a few favorable trials.

\begin{figure*}[t]
    \centering
    \includegraphics[width=0.99\textwidth]{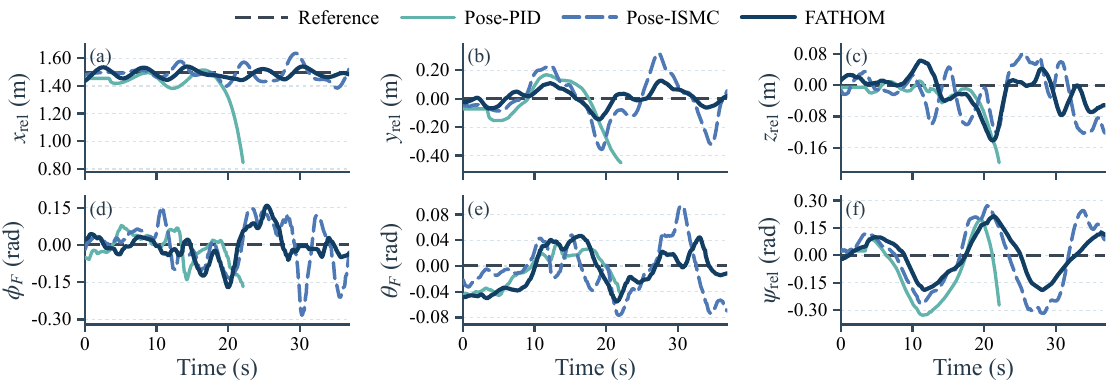}
    \caption{Representative T3 challenging-motion responses. Relative position, follower roll and pitch, and relative yaw are shown for Pose-PID, Pose-ISMC, and FATHOM with the constant reference. The Pose-PID trial terminates following tracking failure.}
    \label{fig:t3_challenging_timeseries}
\end{figure*}

\begin{figure*}[t]
    \centering

    \subfloat[Pose-DP: Lateral Double Shuttle\label{fig:failure_posedp_t2}]{
        \begin{minipage}{0.98\textwidth}
            \centering
            \includegraphics[width=0.160\linewidth]{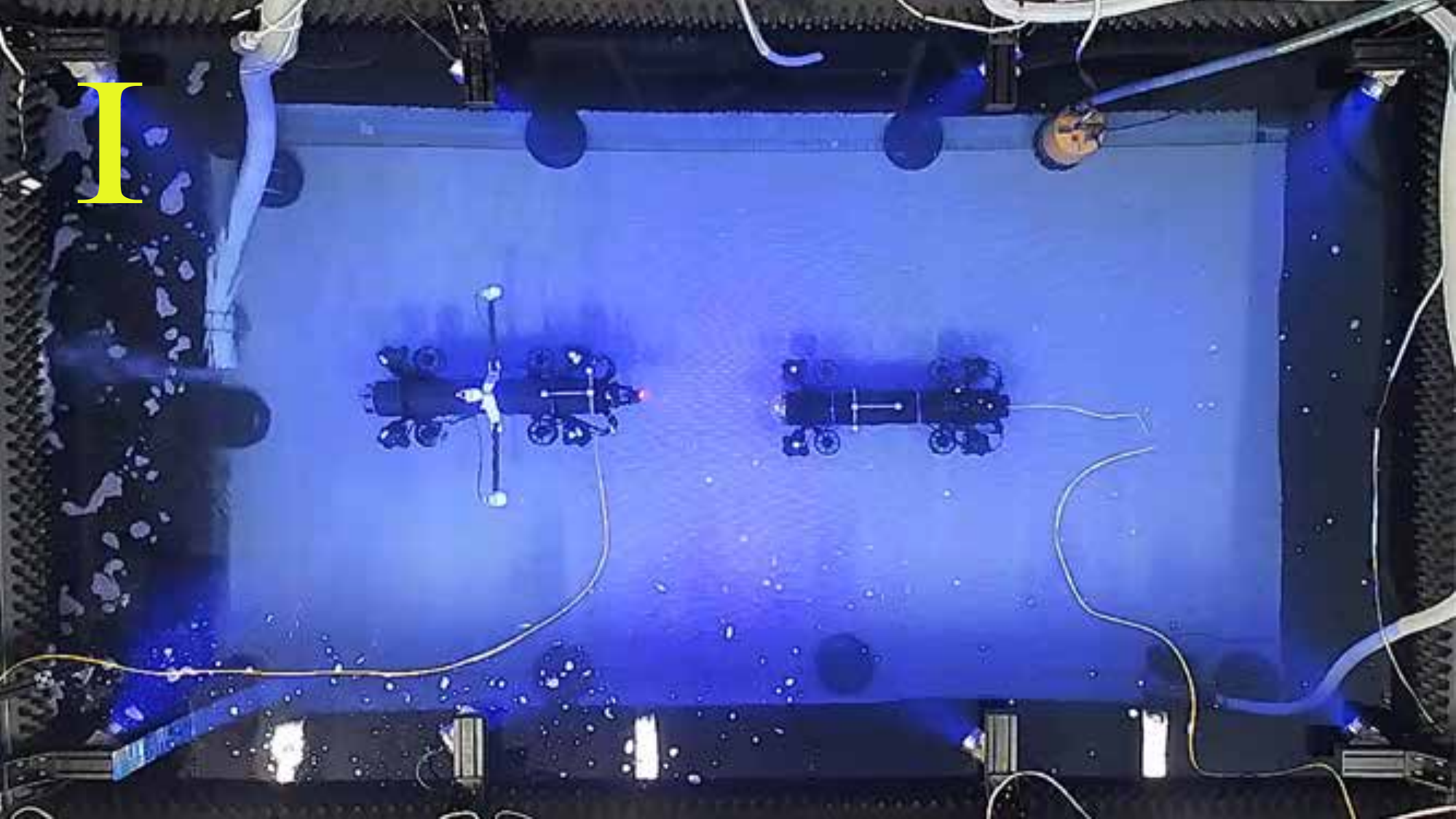}
            \hfill
            \includegraphics[width=0.160\linewidth]{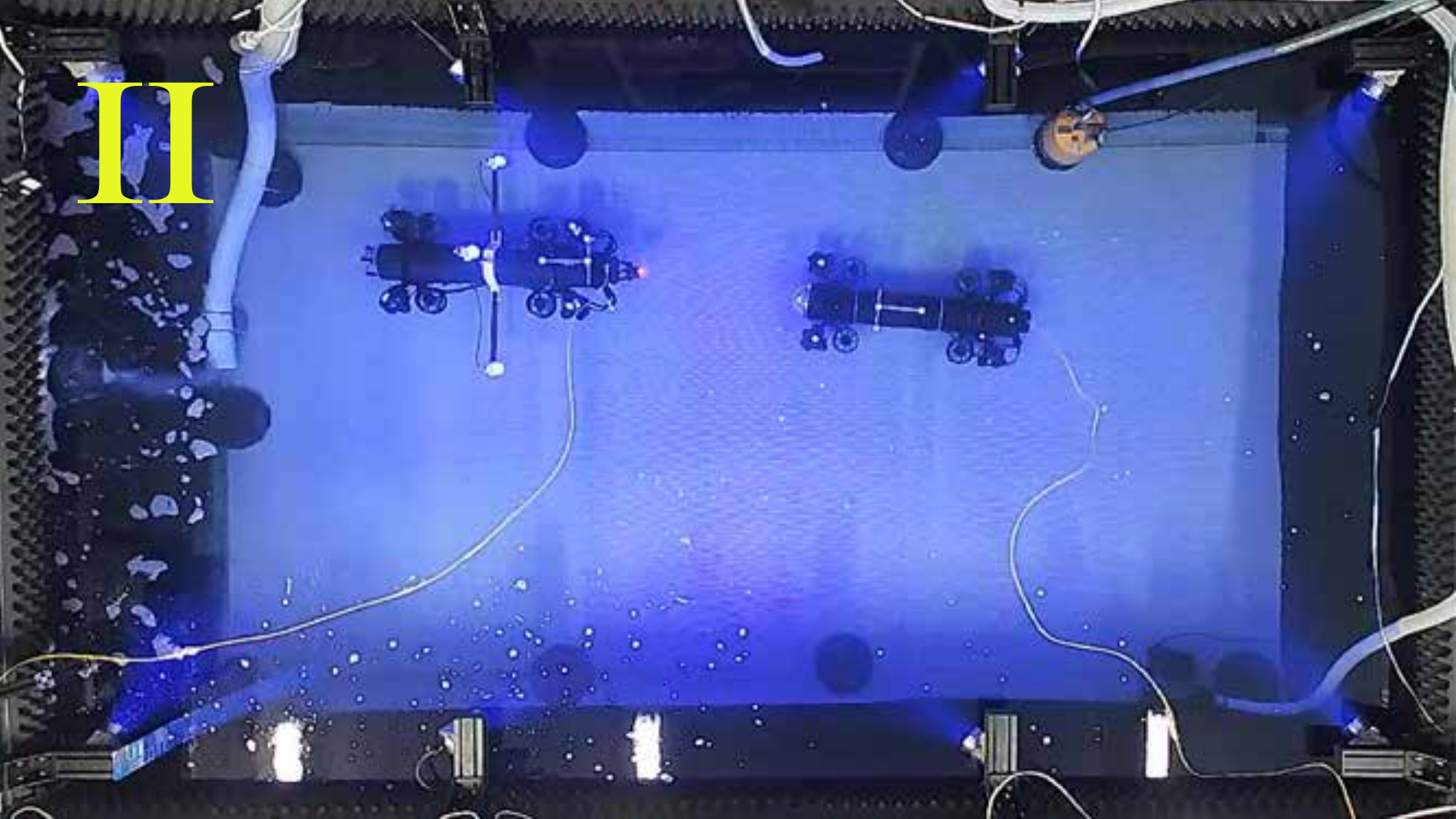}
            \hfill
            \includegraphics[width=0.160\linewidth]{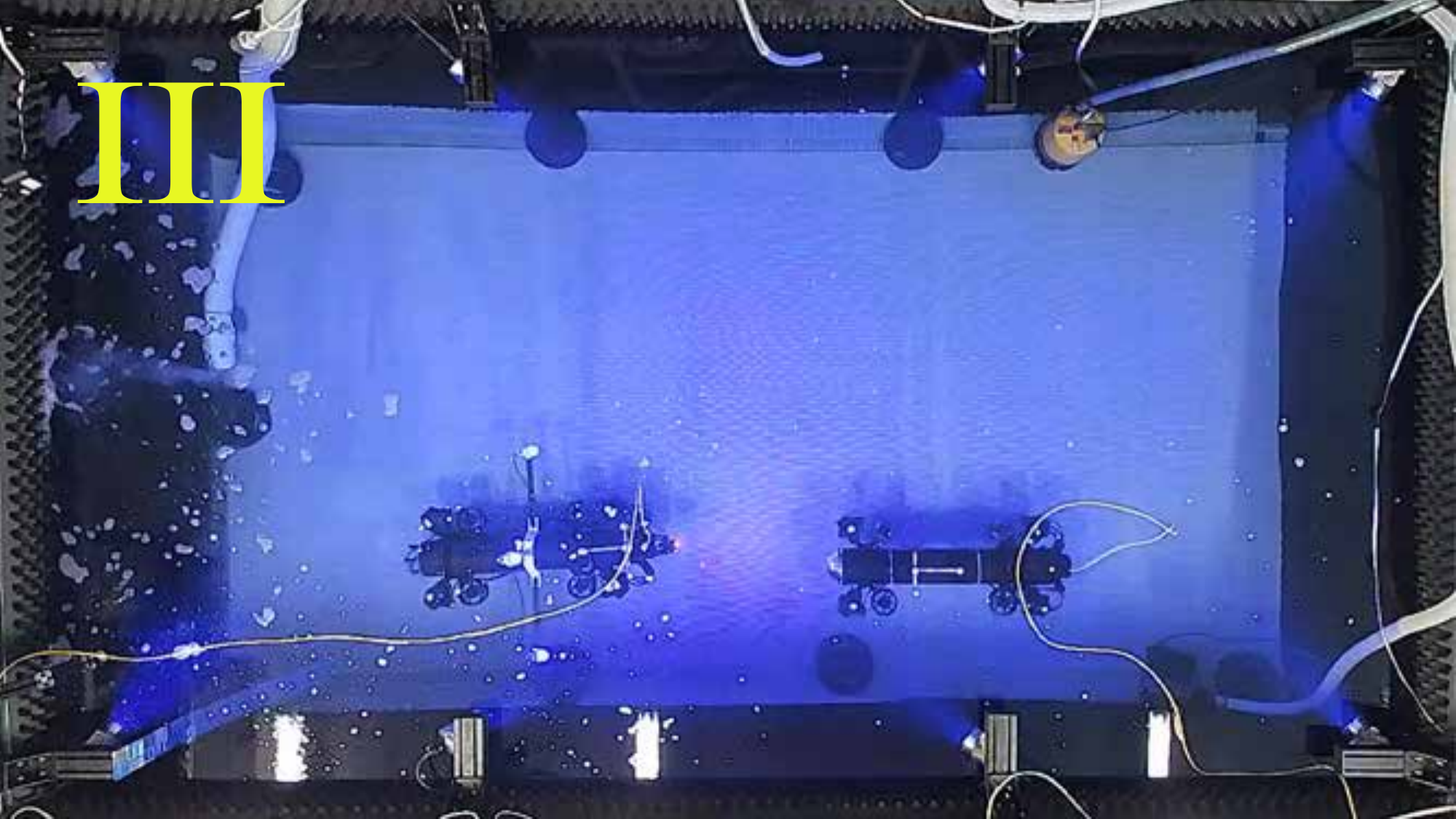}
            \hfill
            \includegraphics[width=0.160\linewidth]{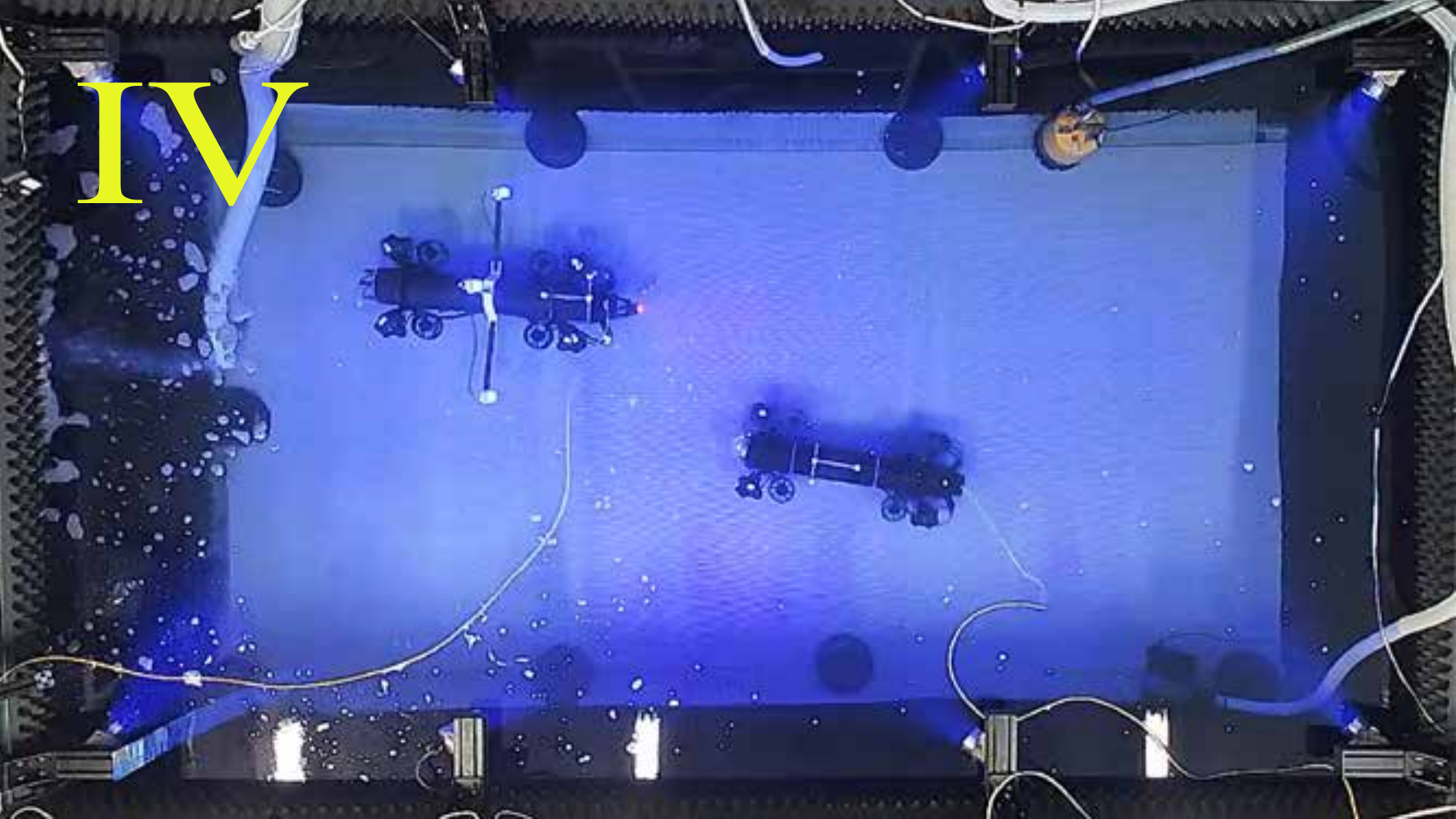}
            \hfill
            \includegraphics[width=0.160\linewidth]{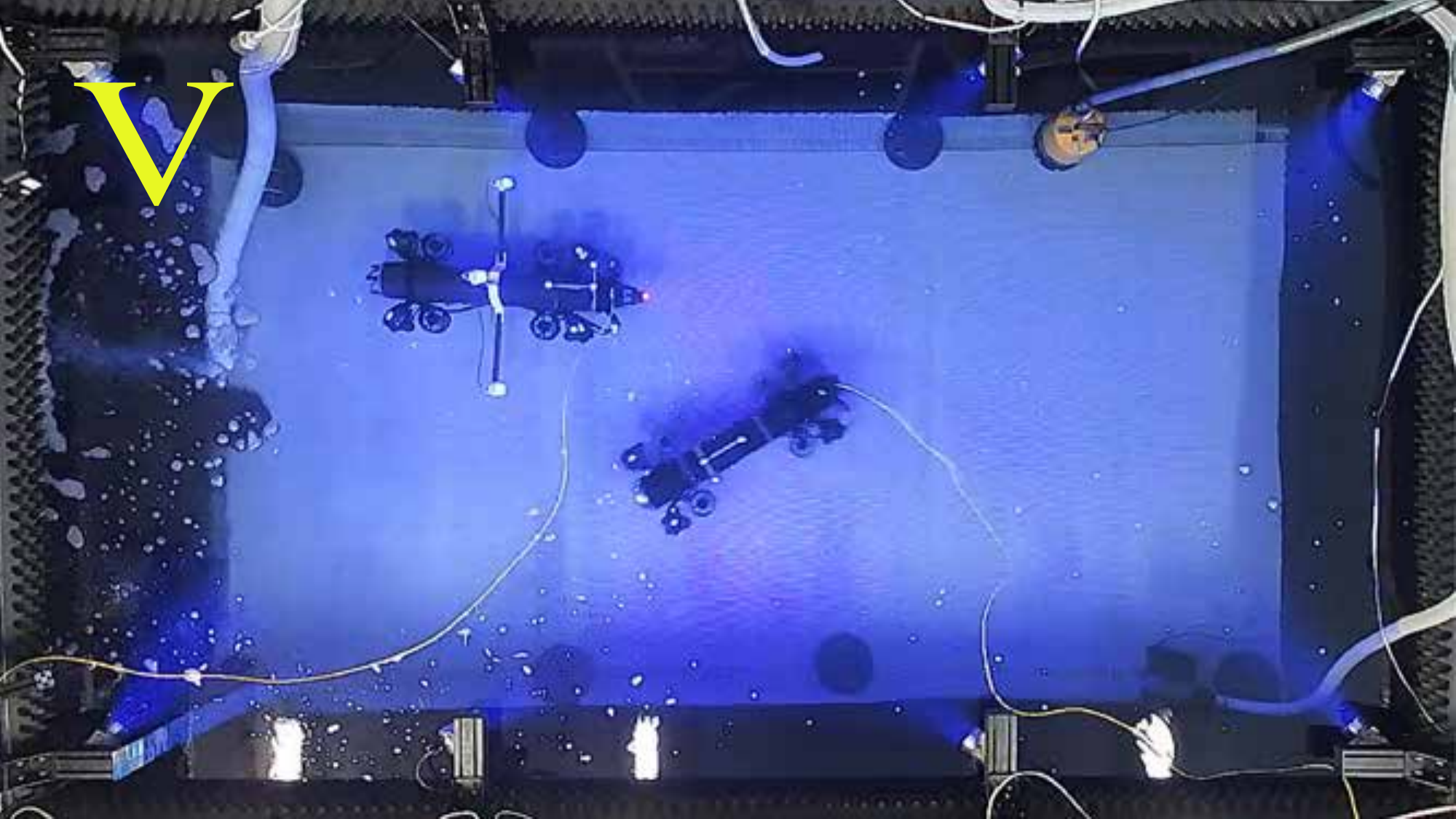}
            \hfill
            \includegraphics[width=0.160\linewidth]{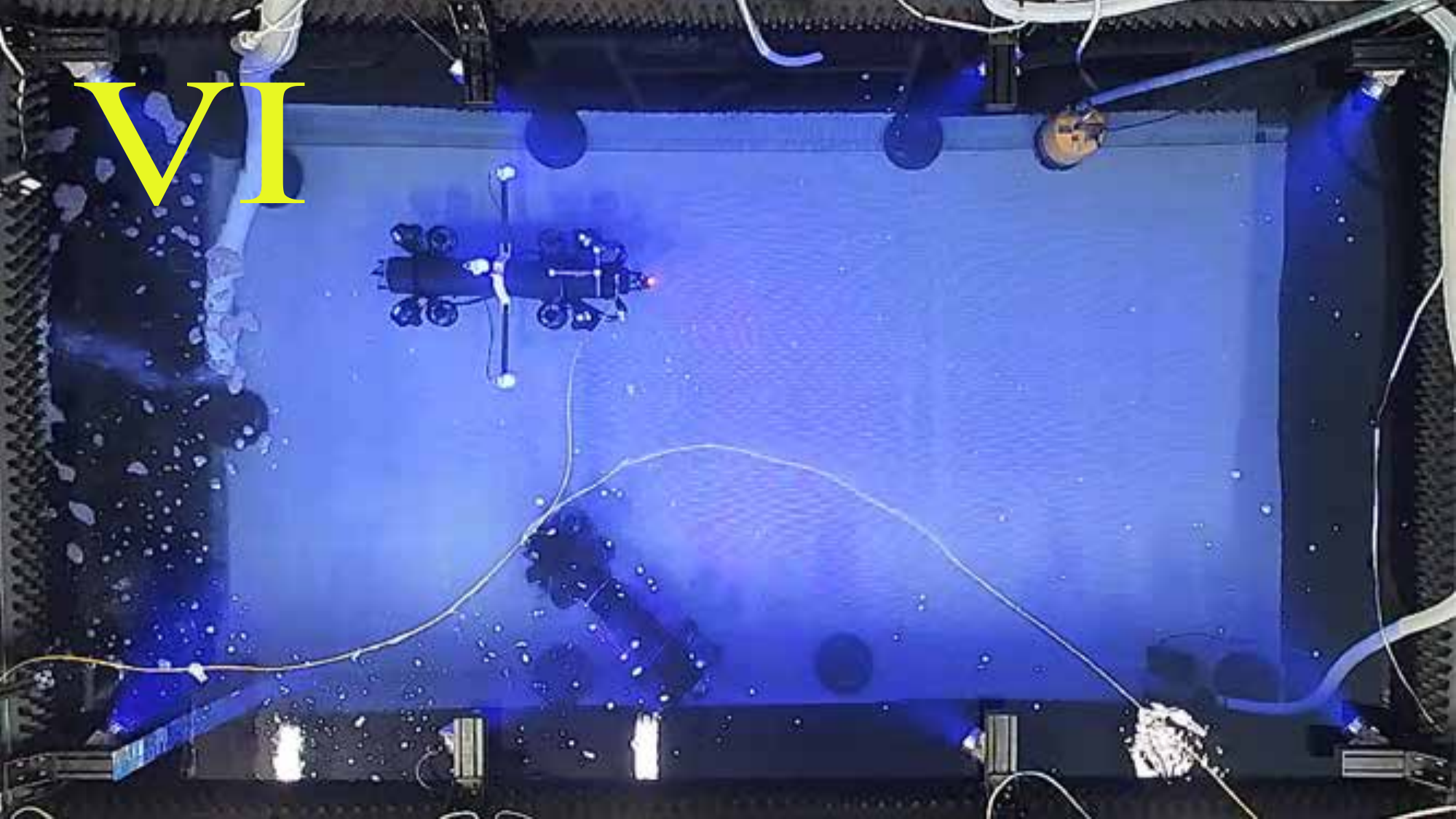}
        \end{minipage}
    }

    \vspace{-0.6em}

    \subfloat[Pose-PID: Yaw Double Sweep\label{fig:failure_posepid_t3}]{
        \begin{minipage}{0.98\textwidth}
            \centering
            \includegraphics[width=0.160\linewidth]{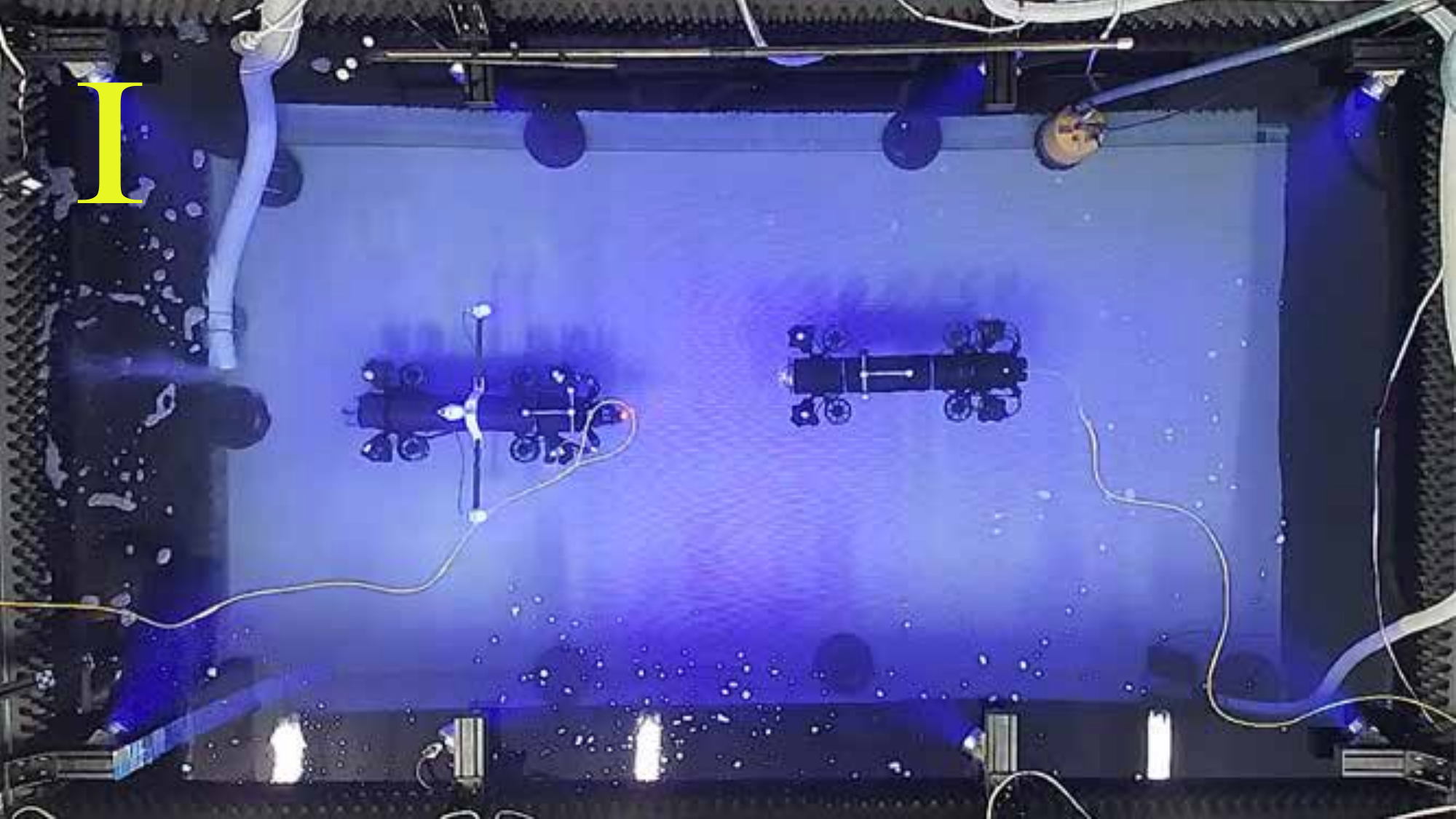}
            \hfill
            \includegraphics[width=0.160\linewidth]{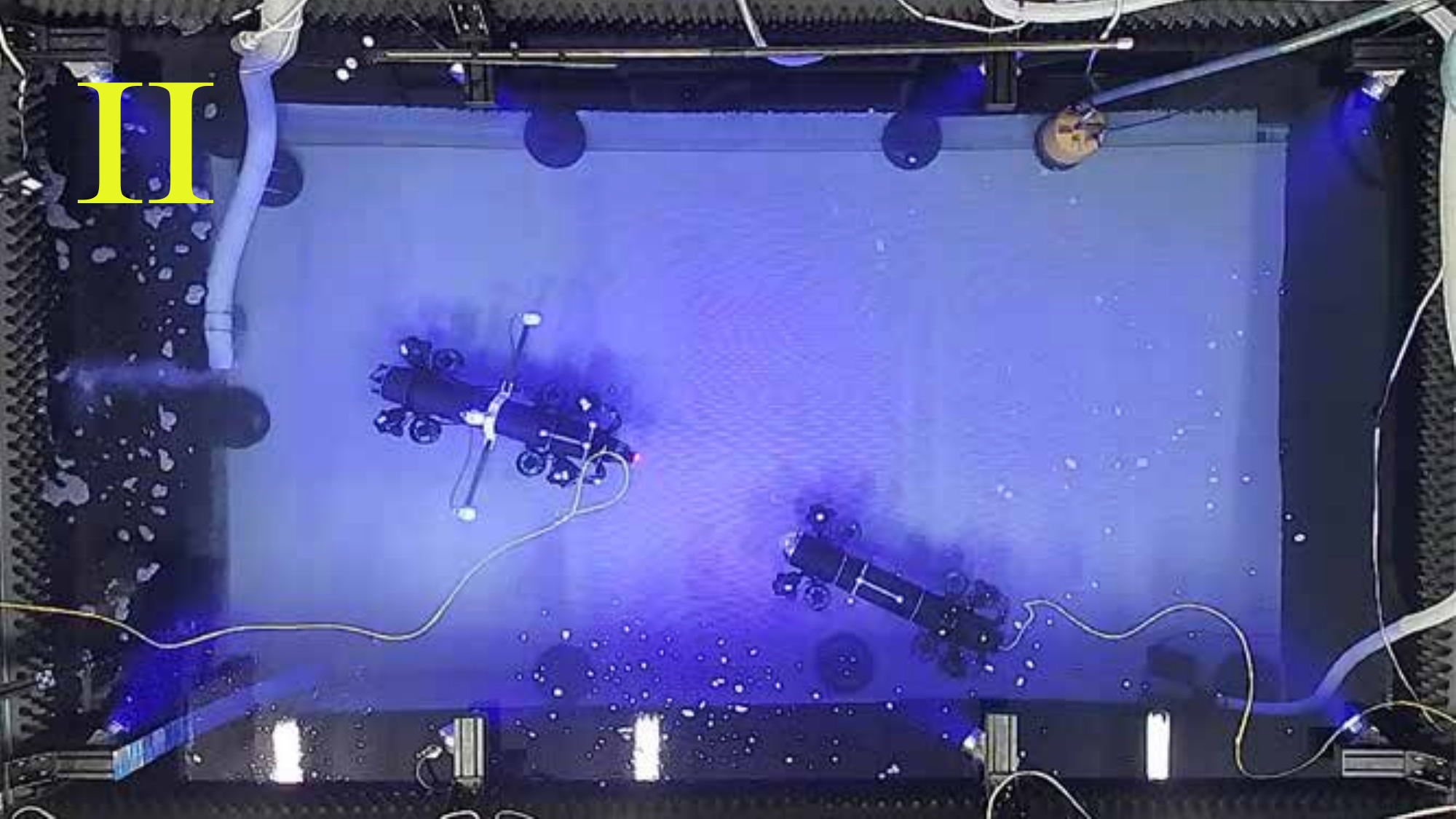}
            \hfill
            \includegraphics[width=0.160\linewidth]{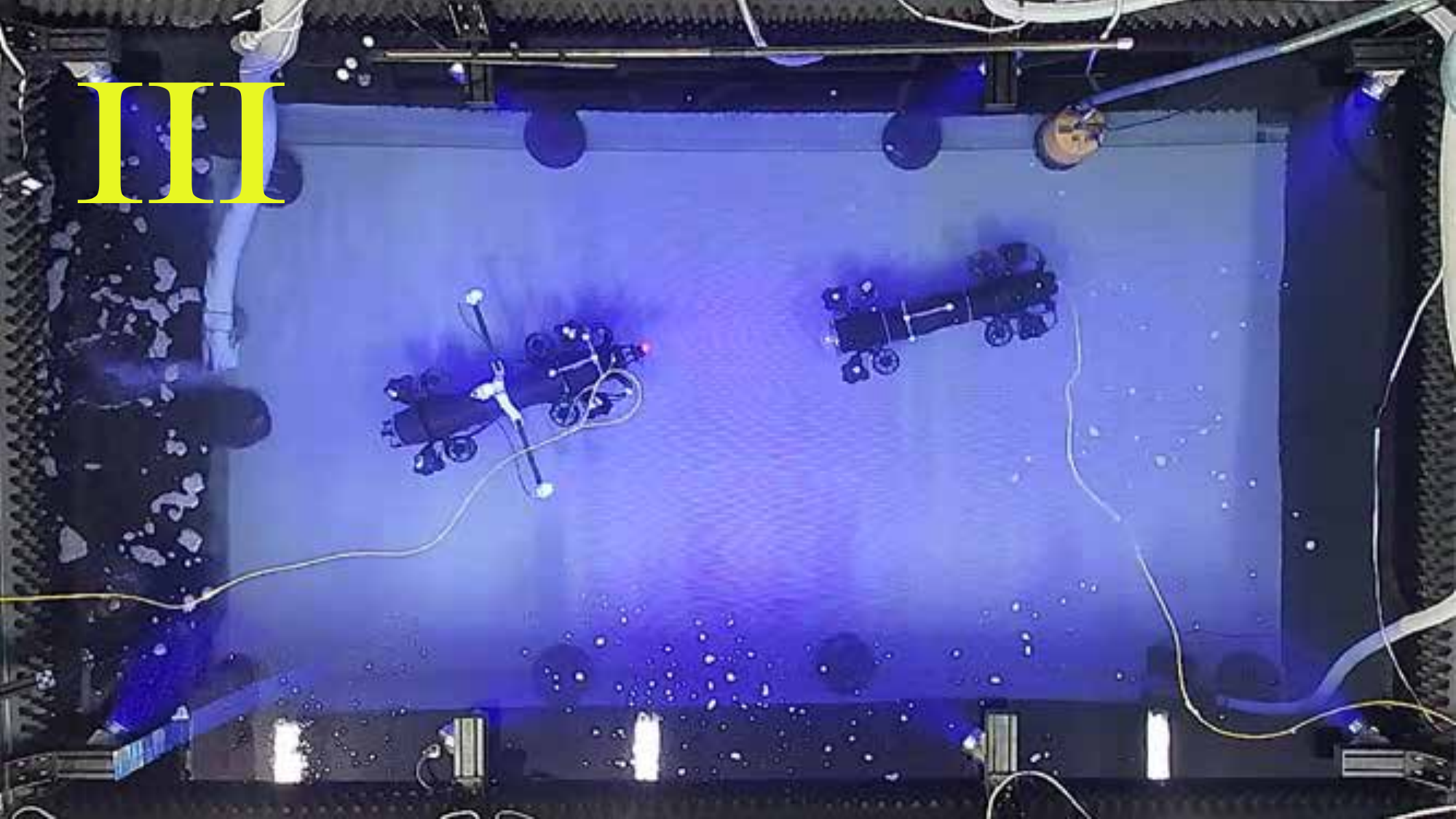}
            \hfill
            \includegraphics[width=0.160\linewidth]{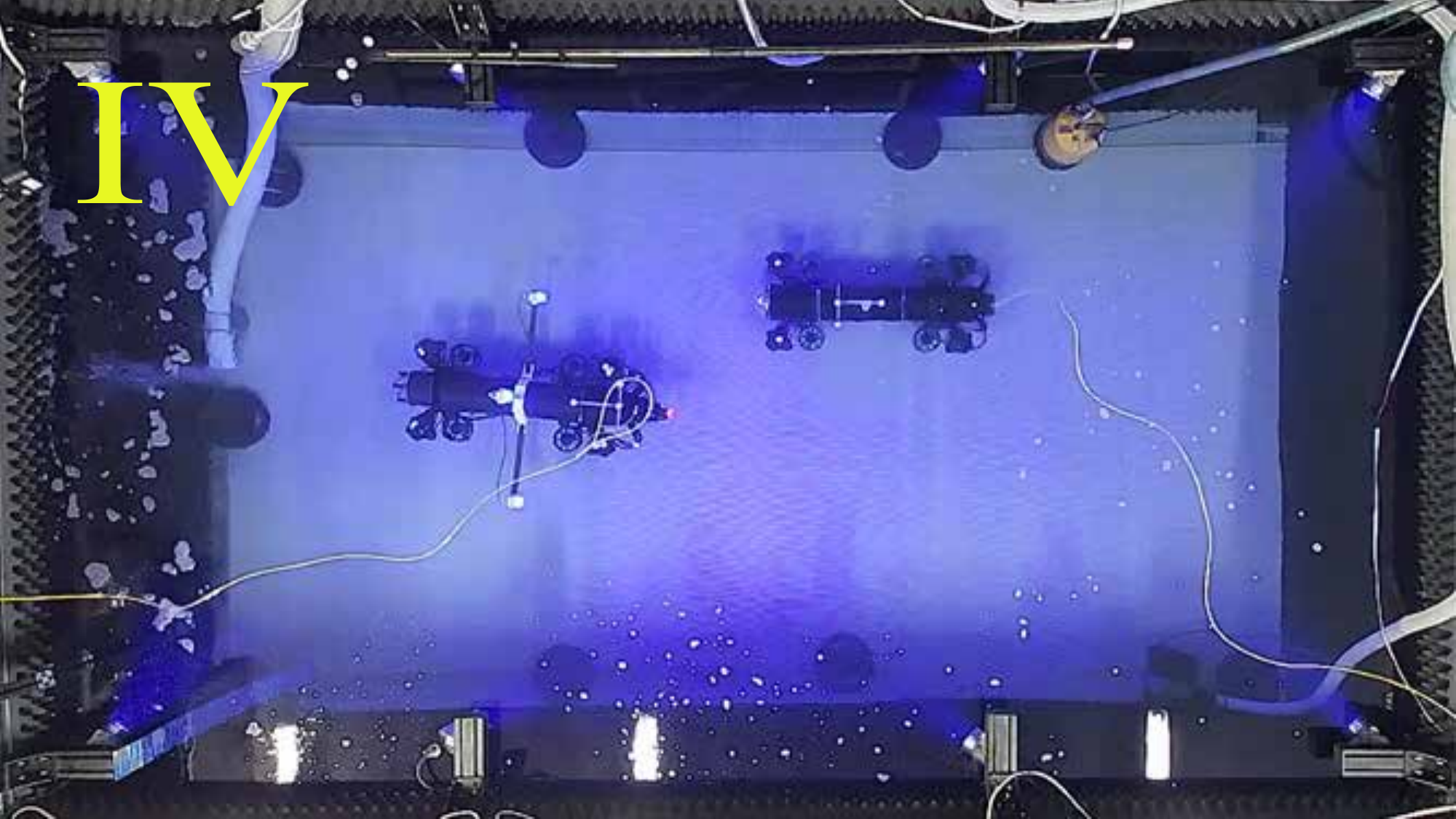}
            \hfill
            \includegraphics[width=0.160\linewidth]{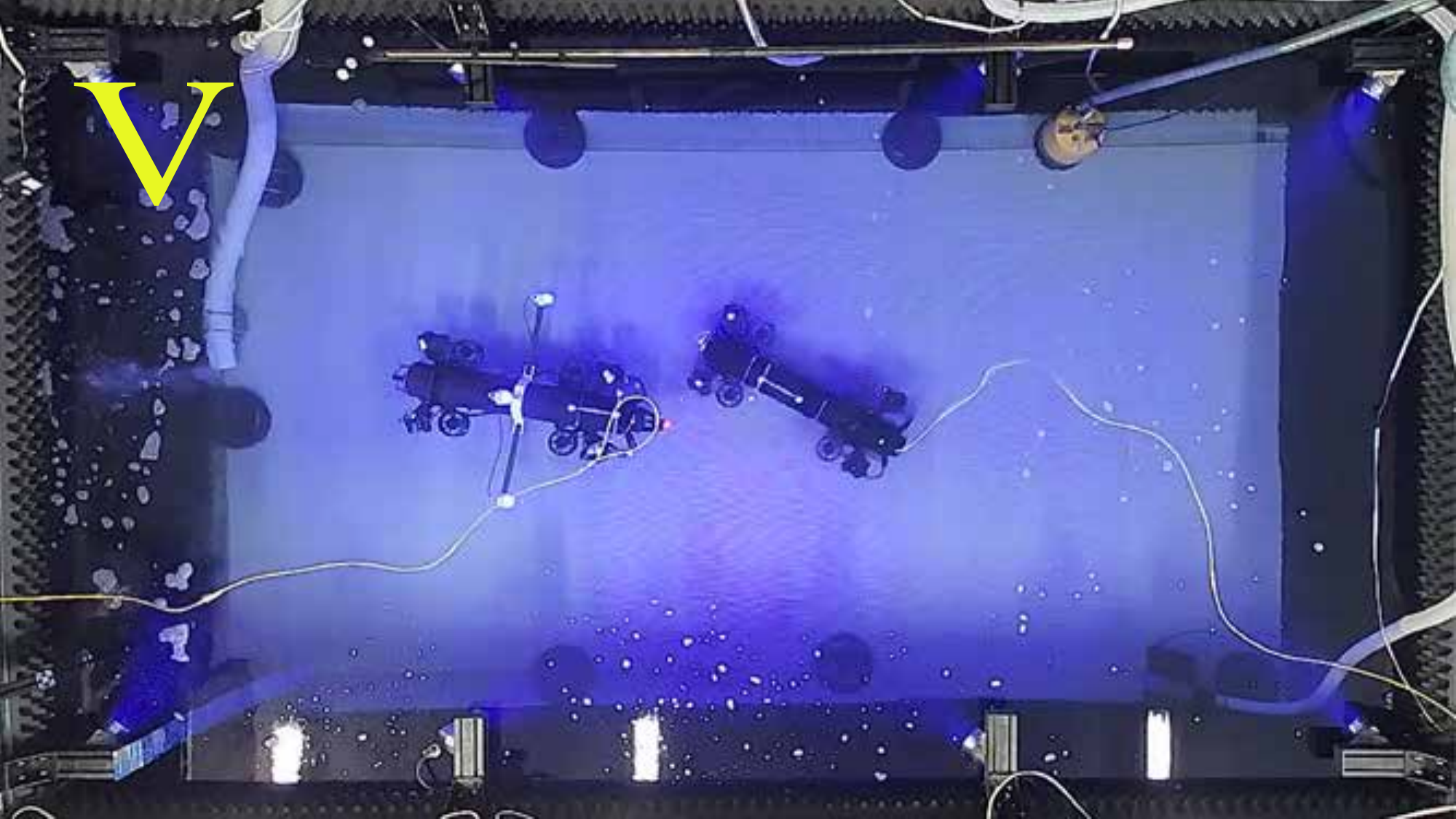}
            \hfill
            \includegraphics[width=0.160\linewidth]{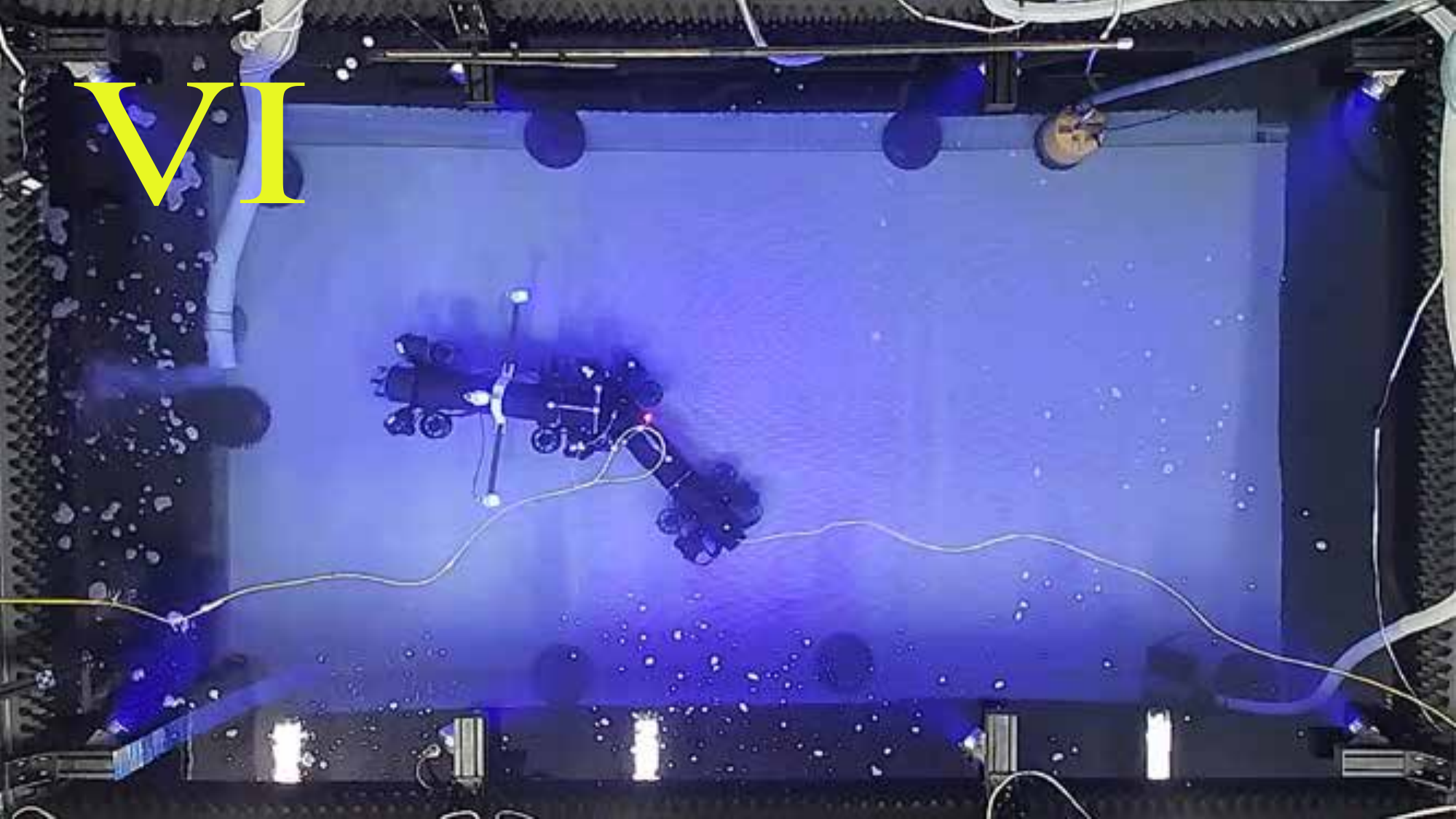}
        \end{minipage}
    }

    \caption{Third-person snapshots of failed challenging-motion trials. (a) Pose-DP on T2 lateral double shuttle and (b) Pose-PID on T3 yaw double sweep. Each row shows the initial state, four intermediate stages, and the terminal state.}
    \label{fig:real_tracking_failures}
\end{figure*}

The representative T3 challenging responses in Fig.~\ref{fig:t3_challenging_timeseries} illustrate the corresponding closed-loop behavior. Pose-PID loses longitudinal tracking near $t=22$~s, as $x_{\mathrm{rel}}$ decreases from the $1.5$~m reference toward $0.85$~m before early termination. Pose-ISMC completes the maneuver but exhibits larger oscillations, particularly in relative yaw, with excursions approaching $0.3$~rad. In contrast, FATHOM keeps $x_{\mathrm{rel}}$ approximately within $1.4$--$1.55$~m and limits yaw excursions to around $0.2$~rad while completing the full trajectory. The disparity is also visible in Fig.~\ref{fig:real_tracking_failures}, where Pose-PID loses formation during the challenging T3 maneuver, in contrast to the completed FATHOM sequences in Fig.~\ref{fig:real_tracking_snapshots}.

Rapid motion alters marker visibility and geometry, while reflections and background lights introduce ambiguous detections. Errors in instantaneous pose estimates may induce inappropriate feedback corrections despite robust control design. FATHOM instead maintains visual--inertial temporal context and predicts action consequences, reducing dependence on individual pose reconstructions. The following analysis examines the representation and selection effects.

\begin{table*}[t]
\centering
\caption{Real-AUV tracking results for four maneuvers under nominal and challenging motion. Task-wise NRMSE is averaged over three trials; aggregate NRMSE and completion rates summarize 12 trials per regime and 24 overall. Superscripts give failed-trial counts out of three. Mocap-PID uses privileged motion-capture feedback.}
\label{tab:real_tracking_comparison}

\renewcommand{\arraystretch}{1.16}
\setlength{\tabcolsep}{1.45pt}
\footnotesize

\newcommand{\fail}[1]{\textsuperscript{\tiny #1}}

\begin{tabular*}{\textwidth}{
@{\extracolsep{\fill}}
l
cccccccc
ccc
ccc
@{}
}
\hline

&
\multicolumn{8}{c}{Tracking NRMSE}
&
\multicolumn{3}{c}{Mean NRMSE}
&
\multicolumn{3}{c}{Success Rate (\%)}
\\

\cline{2-9}
\cline{10-12}
\cline{13-15}

\multirow{2}{*}{Method}
& \multicolumn{2}{c}{T1}
& \multicolumn{2}{c}{T2}
& \multicolumn{2}{c}{T3}
& \multicolumn{2}{c}{T4}
& \multirow{2}{*}{Nom.}
& \multirow{2}{*}{Chall.}
& \multirow{2}{*}{All}
& \multirow{2}{*}{Nom.}
& \multirow{2}{*}{Chall.}
& \multirow{2}{*}{All}
\\

\cline{2-3}
\cline{4-5}
\cline{6-7}
\cline{8-9}

&
Nom. & Chall.
& Nom. & Chall.
& Nom. & Chall.
& Nom. & Chall.
& & & & & &
\\

\hline

Mocap-PID
& 0.075 & 0.134
& 0.083 & 0.140
& 0.077 & 0.155
& 0.101 & 0.157
& 0.084 & 0.147 & 0.115
& 100.0 & 100.0 & 100.0
\\

Pose-PID
& 0.116 & 0.209
& 0.126 & 0.272\fail{1}
& 0.111 & 0.302\fail{2}
& 0.134 & 0.372\fail{3}
& 0.122 & 0.289 & 0.205
& 100.0 & 50.0 & 75.0
\\

Pose-ISMC
& 0.110 & 0.184
& 0.121 & 0.244\fail{1}
& 0.103 & 0.261\fail{1}
& 0.127 & 0.314\fail{2}
& 0.115 & 0.251 & 0.183
& 100.0 & 66.7 & 83.3
\\

Pose-DP
& 0.128 & 0.205
& 0.133 & 0.315\fail{2}
& 0.119 & 0.333\fail{2}
& 0.138 & 0.408\fail{3}
& 0.129 & 0.315 & 0.222
& 100.0 & 41.7 & 70.8
\\

DAP-only
& 0.090 & 0.189
& 0.104 & 0.173
& 0.090 & 0.188
& 0.103 & 0.217
& 0.097 & 0.192 & 0.144
& 100.0 & 100.0 & 100.0
\\

\textbf{FATHOM}
& \textbf{0.073} & \textbf{0.155}
& \textbf{0.086} & \textbf{0.155}
& \textbf{0.082} & \textbf{0.166}
& \textbf{0.094} & \textbf{0.179}
& \textbf{0.084} & \textbf{0.164} & \textbf{0.124}
& \textbf{100.0} & \textbf{100.0} & \textbf{100.0}
\\

\hline
\end{tabular*}
\end{table*}

\subsection{Representation and Predictive Decision Analysis}
\label{subsec:representation_decision_real}

Table~\ref{tab:real_tracking_comparison} compares FATHOM with Pose-DP to assess the complete belief-conditioned predictive framework against direct pose-conditioned diffusion control. FATHOM reduces challenging-motion NRMSE from $0.315$ to $0.164$ ($47.9\%$), including reductions of $50.8\%$, $50.2\%$, and $56.1\%$ on T2--T4. Pose-DP fails in seven of 12 challenging trials, whereas FATHOM completes all 12. Although Pose-DP uses the strong AMR-Pose frontend, its policy is conditioned on a current six-dimensional estimate that may be perturbed by marker loss or false associations. FATHOM instead incorporates sparse visual observations, IMU measurements, and action history into a recurrent predictive belief. The improvement reflects temporal, task-predictive information rather than the belief dimension alone. This comparison evaluates the complete framework; the effect of predictive selection is isolated by DAP-only below.

DAP-only uses the same belief and diffusion prior as FATHOM but executes a generated trajectory without world-model selection. Adding WMPS reduces nominal NRMSE from $0.097$ to $0.084$ ($13.4\%$) and challenging NRMSE from $0.192$ to $0.164$ ($14.6\%$). Across the challenging maneuvers, the error reduction ranges from $10.4\%$ on T2 to $18.0\%$ on T1. Both controllers complete all trials, indicating that predictive selection primarily improves tracking accuracy rather than completion in these conditions. WMPS ranks expert-like candidates by their anticipated tracking consequences instead of accepting the first diffusion sample.

\begin{figure}[t]
    \centering
    \includegraphics[width=0.98\columnwidth]{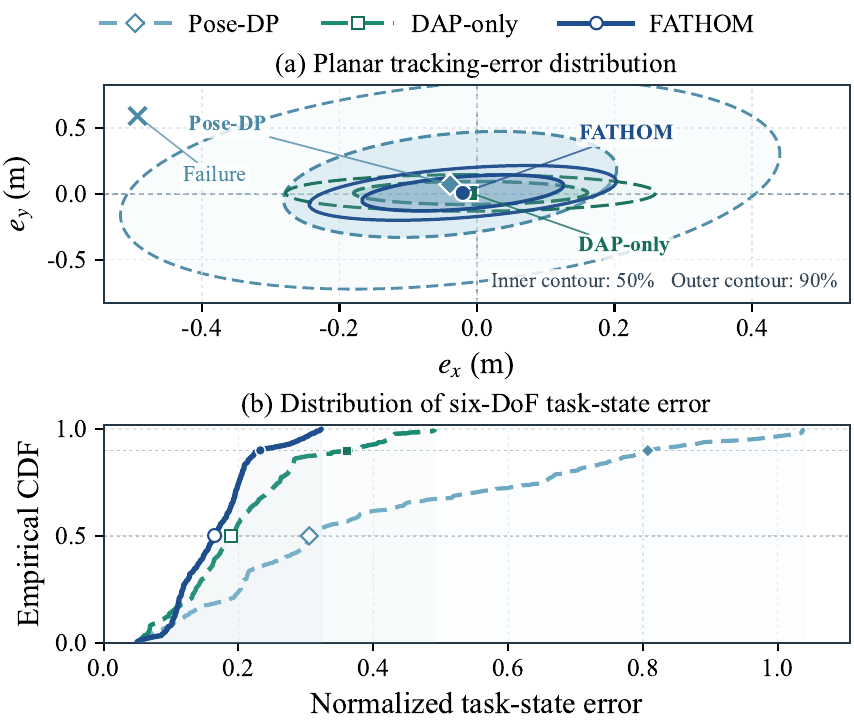}
    \caption{Tracking-error distributions on the challenging T4 maneuver for Pose-DP, DAP-only, and FATHOM. (a) Planar error contours enclosing 50\% and 90\% of samples, with mean markers and the Pose-DP terminal failure marked by a cross. (b) Empirical CDF of frame-wise normalized six-dimensional task-state error. Open and filled markers denote the median and 90th percentile.}
    \label{fig:t4_error_distribution}
\end{figure}

The challenging-motion boxplots in Fig.~\ref{fig:tracking_nrmse_boxplot} give median NRMSEs of $0.332$, $0.190$, and $0.163$ for Pose-DP, DAP-only, and FATHOM, respectively, with interquartile ranges of $0.137$, $0.018$, and $0.014$. Fig.~\ref{fig:t4_error_distribution} further illustrates the difference on a challenging T4 trial. FATHOM concentrates planar errors near the target, whereas Pose-DP exhibits a broader distribution before failure. The median normalized task errors for Pose-DP, DAP-only, and FATHOM are approximately $0.30$, $0.19$, and $0.16$, with 90th percentiles of $0.81$, $0.36$, and $0.23$. The Pose-DP distribution includes only frames recorded before termination. Compared with DAP-only, FATHOM shows a larger reduction at the 90th percentile than at the median, suggesting that WMPS helps suppress large transient errors in this trial. Together with the trial-level distributions in Fig.~\ref{fig:tracking_nrmse_boxplot} and failure snapshots in Fig.~\ref{fig:real_tracking_failures}, these results show improved consistency and fewer large tracking deviations.

We use $K=16$ because larger candidate sets did not consistently improve real-AUV performance. Although Table~\ref{tab:kh_sensitivity} shows diminishing model-predicted cost gains with increasing $K$, these gains are measured in model space and need not translate into lower realized tracking error. Imperfect predictions may also misrank additional candidates. This motivates a practical balance between candidate diversity and selection reliability, consistent with Section~\ref{subsec:selection_theory}.

\subsection{Tracking under Natural Partial Marker Visibility}
\label{subsec:partial_visibility}

Rapid leader motion causes marker loss through occlusion and field-of-view changes, while reflections complicate sparse detection association. Table~\ref{tab:partial_visibility} evaluates frames with $N_{\mathrm{vis}}\leq3$ and recovery over annotated visibility-loss events. FATHOM achieves a partial-visibility NRMSE of $0.163$, improving upon Pose-PID, Pose-ISMC, and Pose-DP by $51.1\%$, $34.8\%$, and $57.7\%$, respectively. Its P95 error is $0.246$, versus $0.416$--$0.699$ for the pose-based baselines. The pose-based recovery rates range from $81.2\%$ to $88.5\%$, while FATHOM and DAP-only recover in every recorded event, but FATHOM reduces DAP-only's NRMSE by $20.9\%$ and mean recovery time from $2.65$ to $2.17$~s ($18.1\%$).

\begin{figure*}[t]
    \centering

    \subfloat[Pose-PID: T3 Challenging\label{fig:visibility_posepid_t3}]{
        \begin{minipage}{0.98\textwidth}
            \centering
            \includegraphics[width=0.192\linewidth]{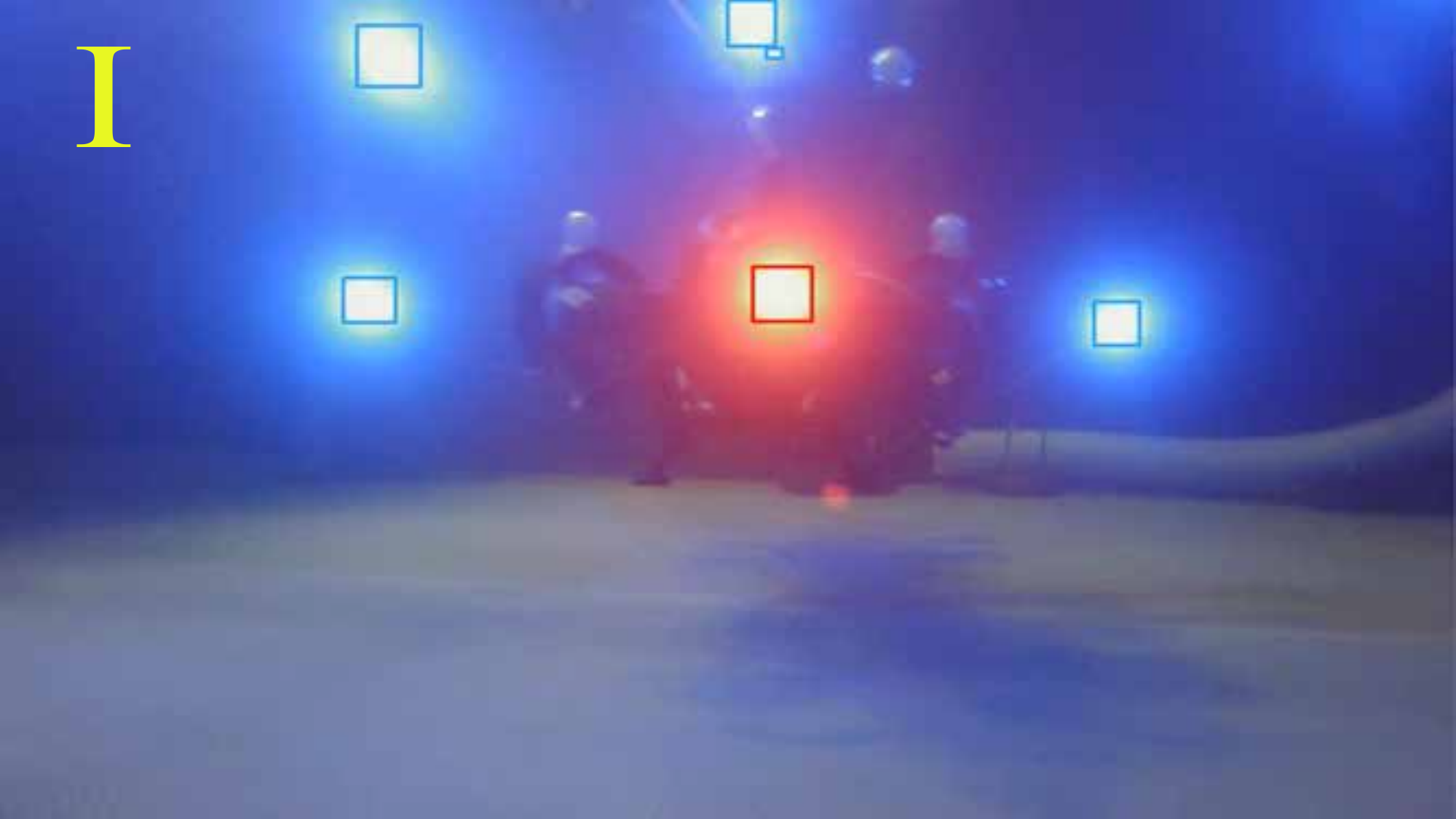}
            \hfill
            \includegraphics[width=0.192\linewidth]{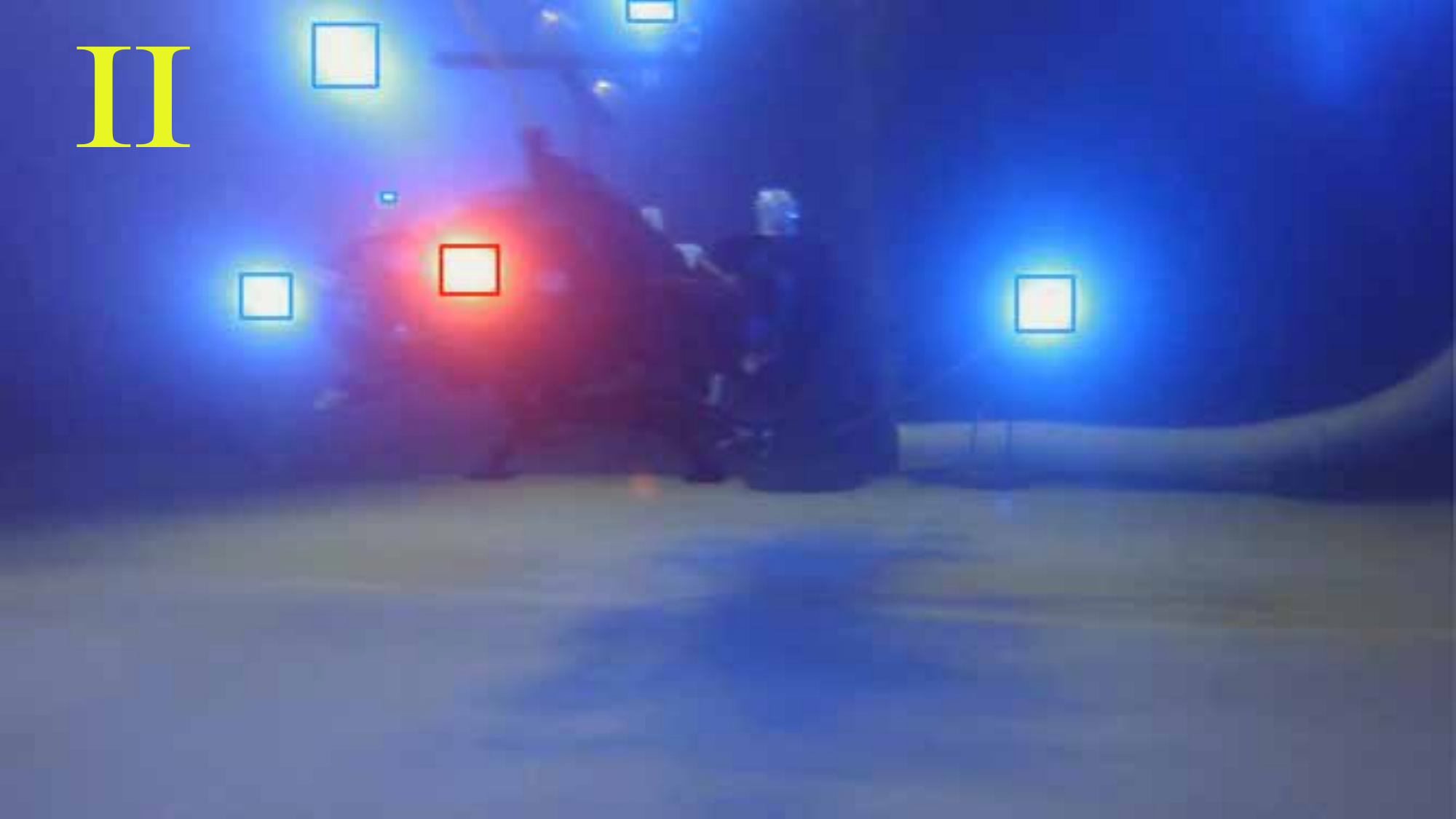}
            \hfill
            \includegraphics[width=0.192\linewidth]{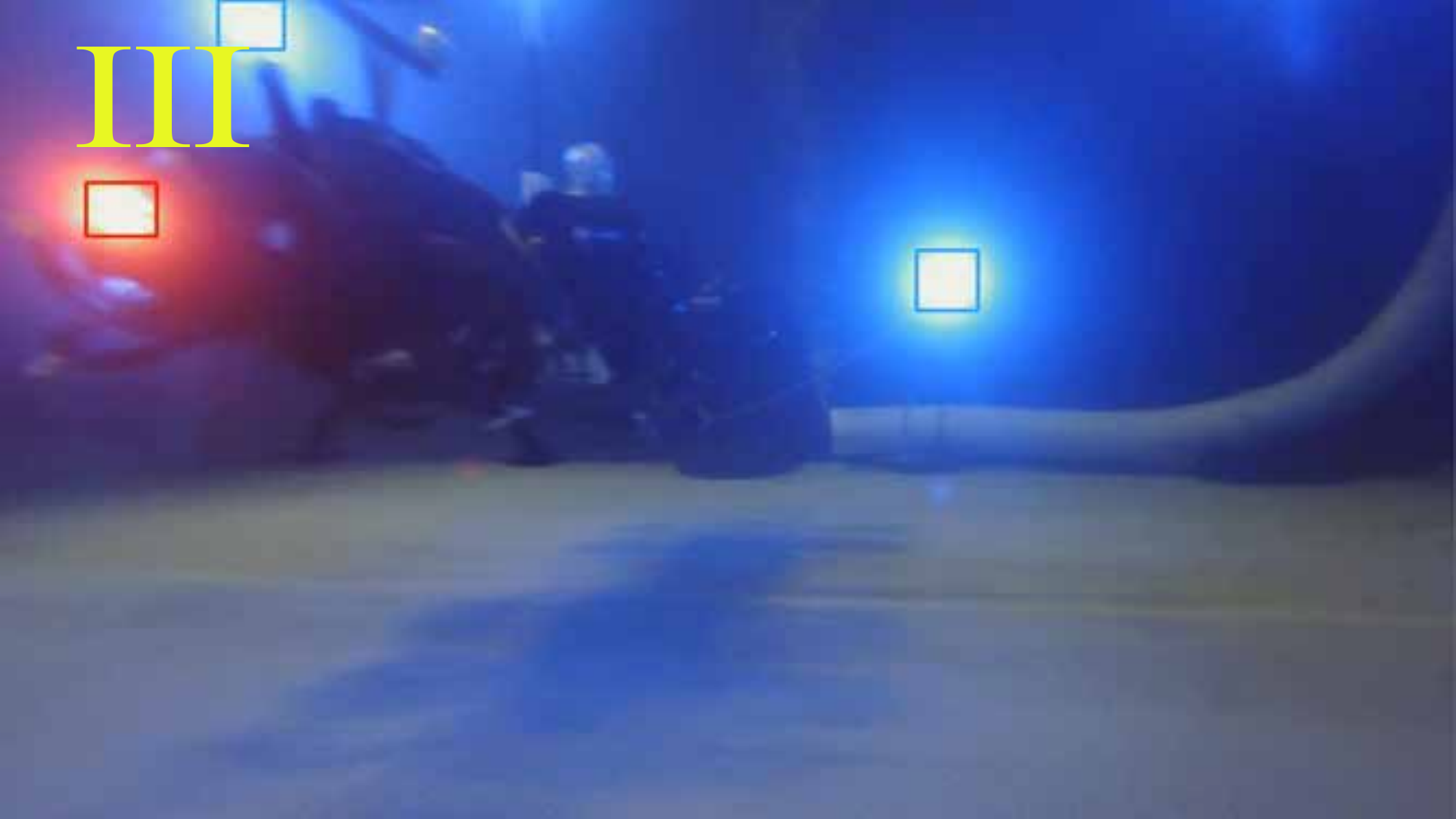}
            \hfill
            \includegraphics[width=0.192\linewidth]{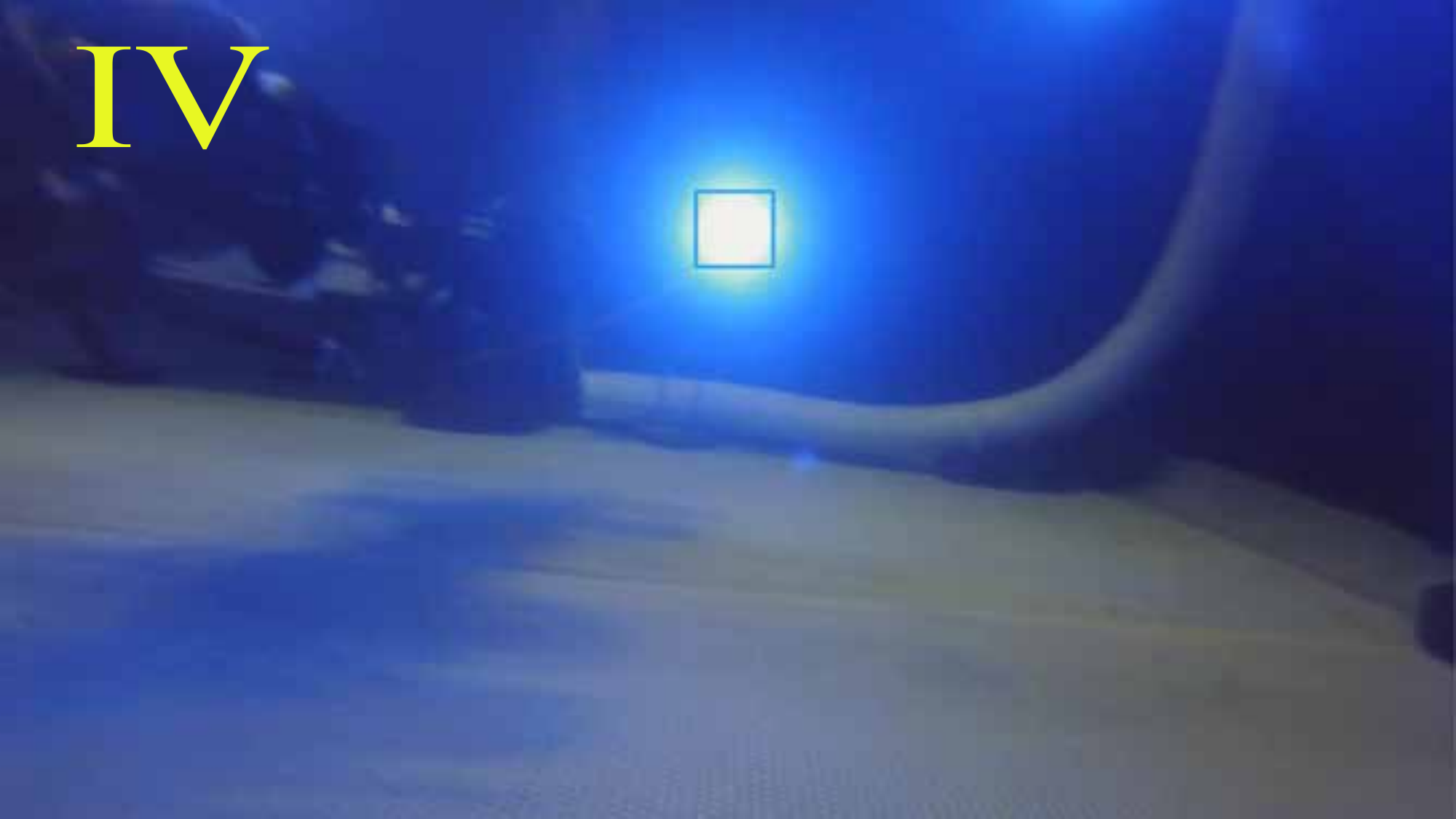}
            \hfill
            \includegraphics[width=0.192\linewidth]{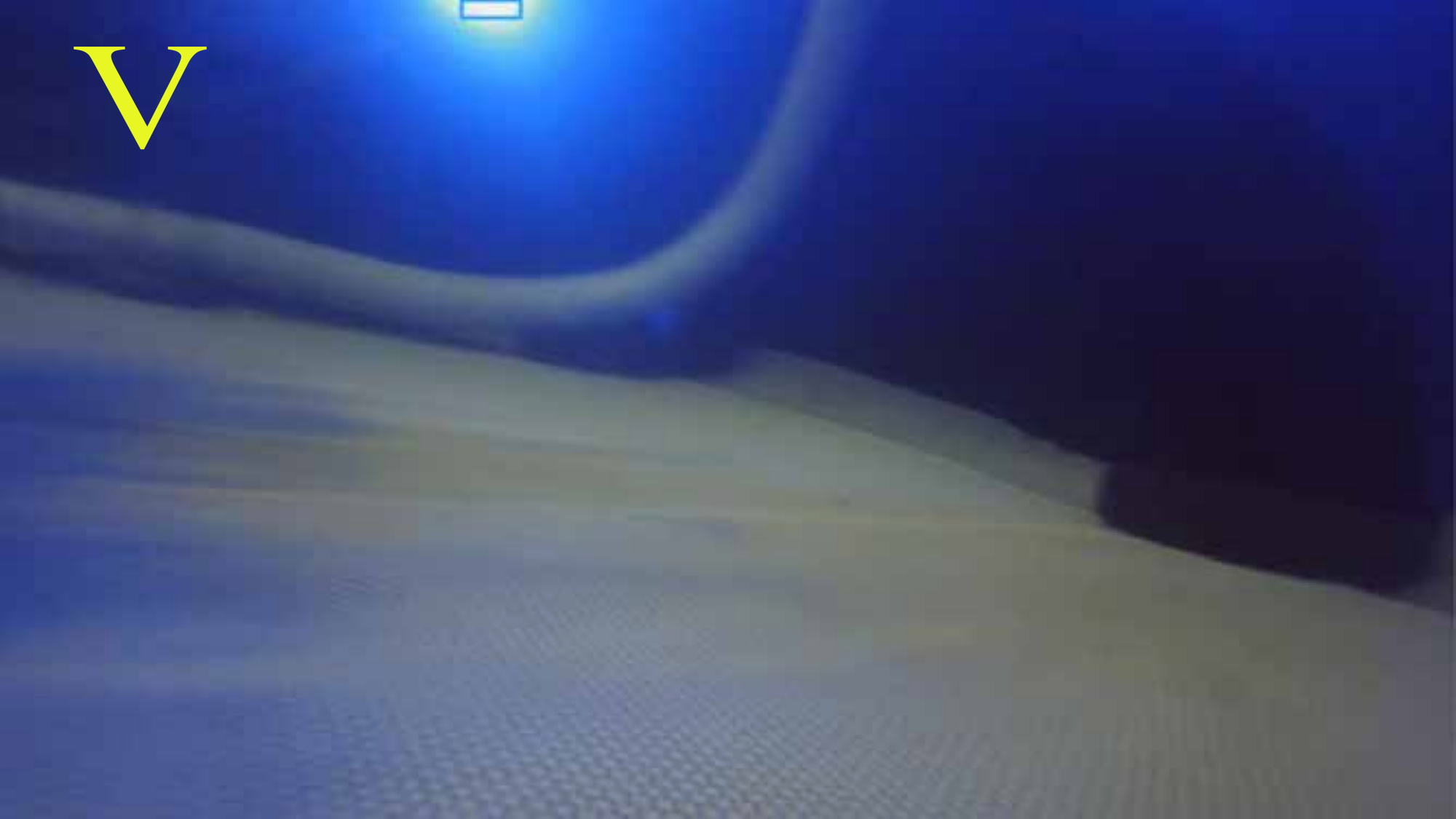}
        \end{minipage}
    }

    \vspace{-0.6em}

    \subfloat[FATHOM: T3 Challenging\label{fig:visibility_fathom_t3}]{
        \begin{minipage}{0.98\textwidth}
            \centering
            \includegraphics[width=0.192\linewidth]{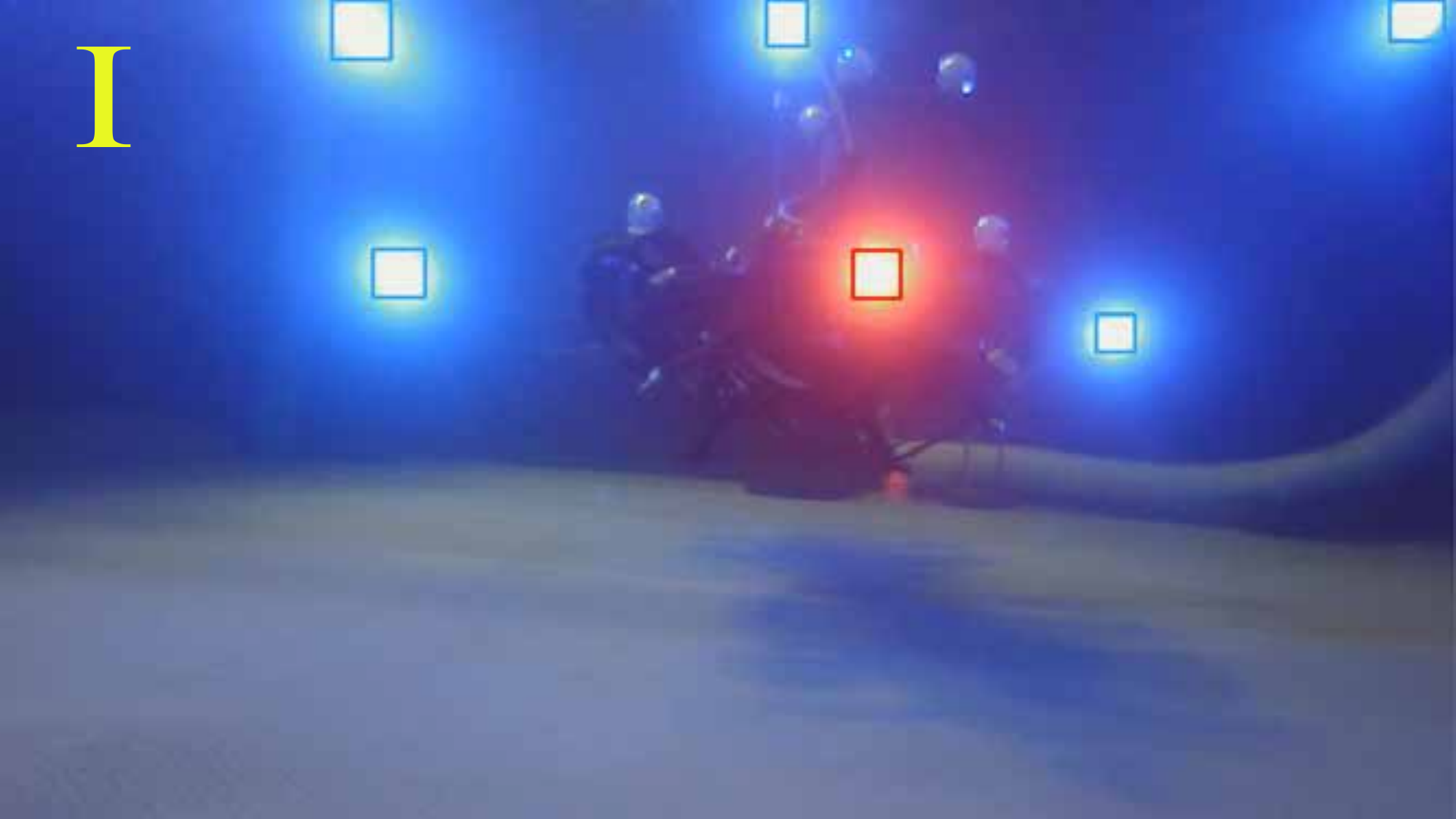}
            \hfill
            \includegraphics[width=0.192\linewidth]{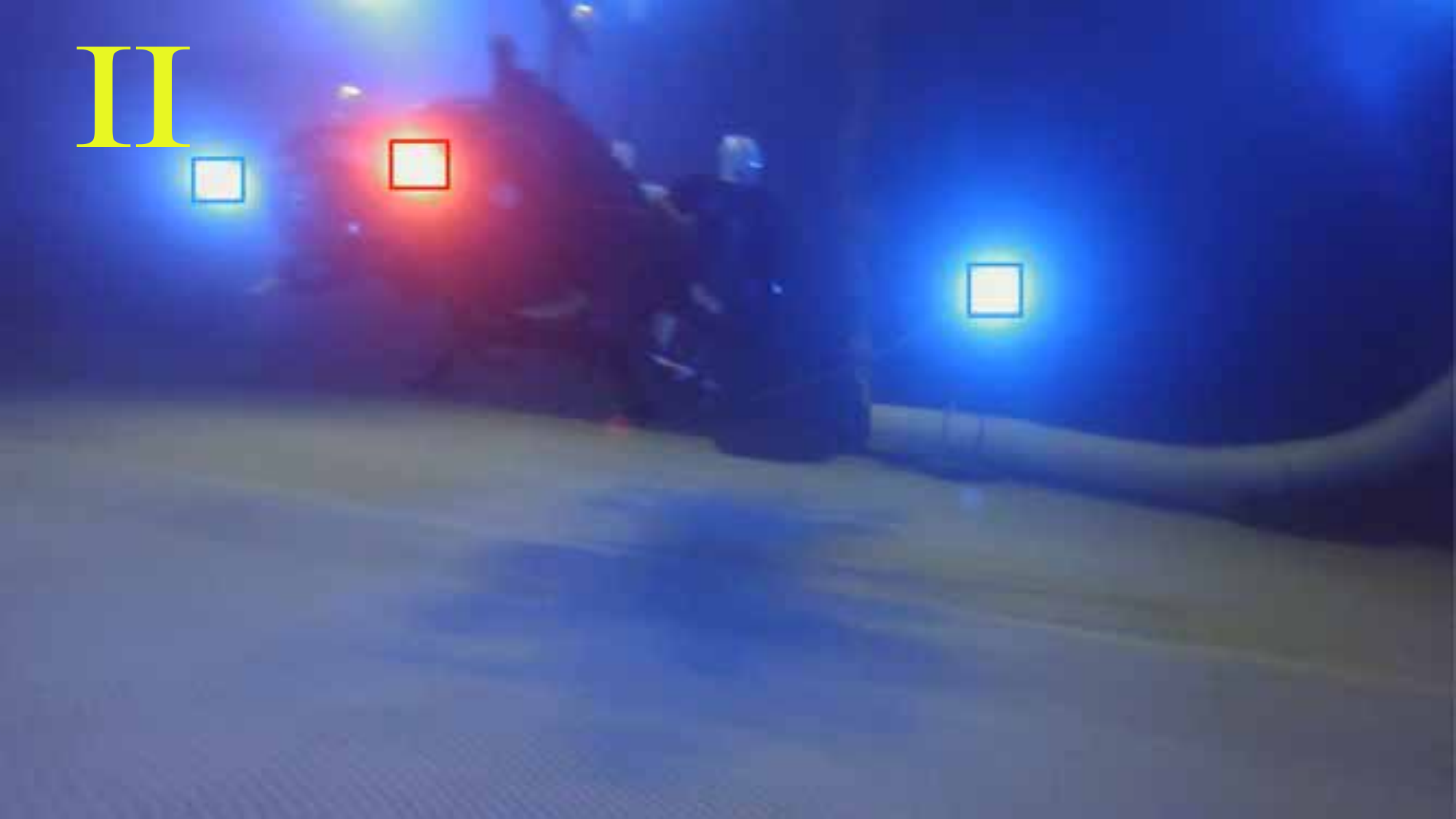}
            \hfill
            \includegraphics[width=0.192\linewidth]{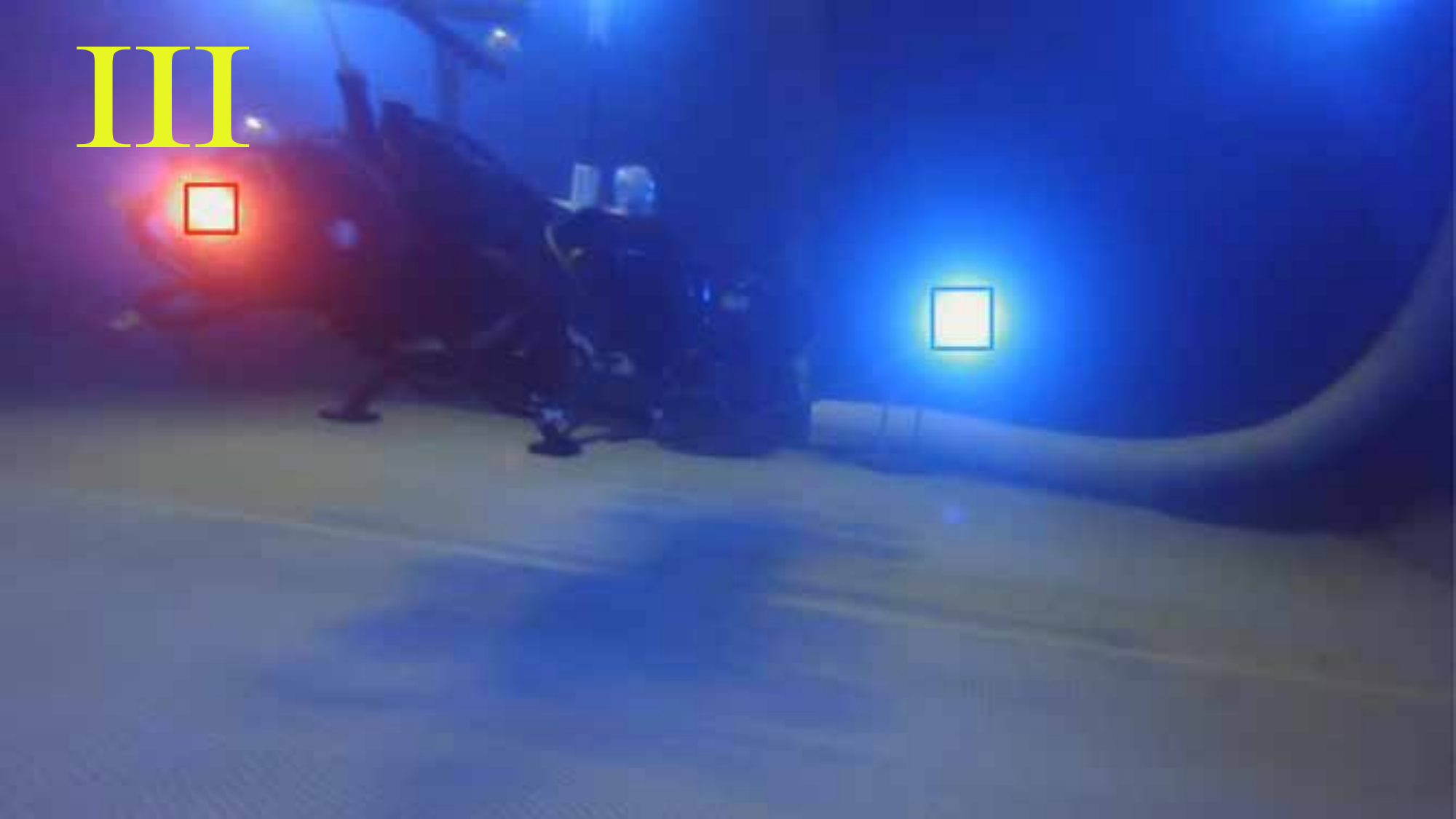}
            \hfill
            \includegraphics[width=0.192\linewidth]{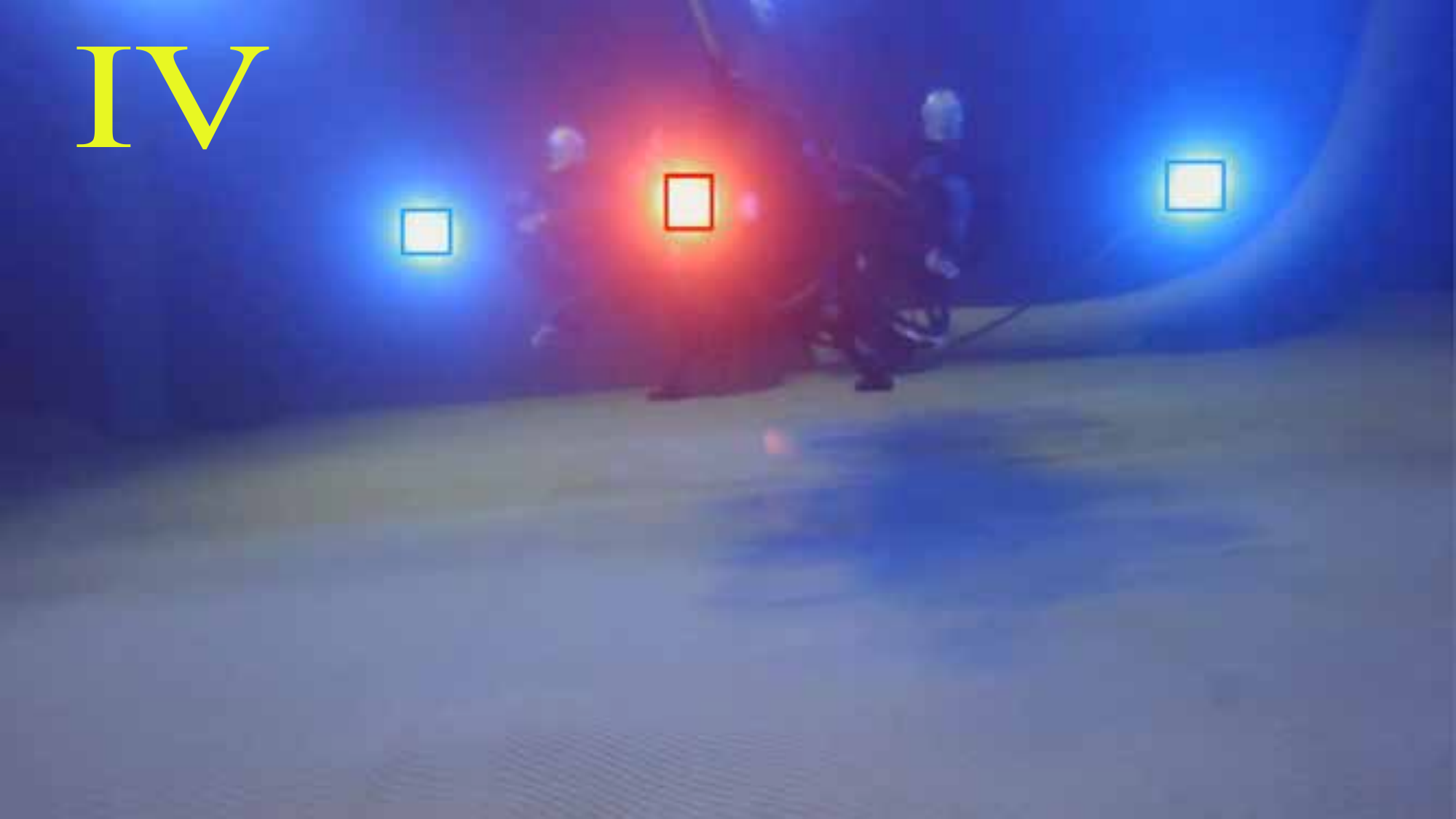}
            \hfill
            \includegraphics[width=0.192\linewidth]{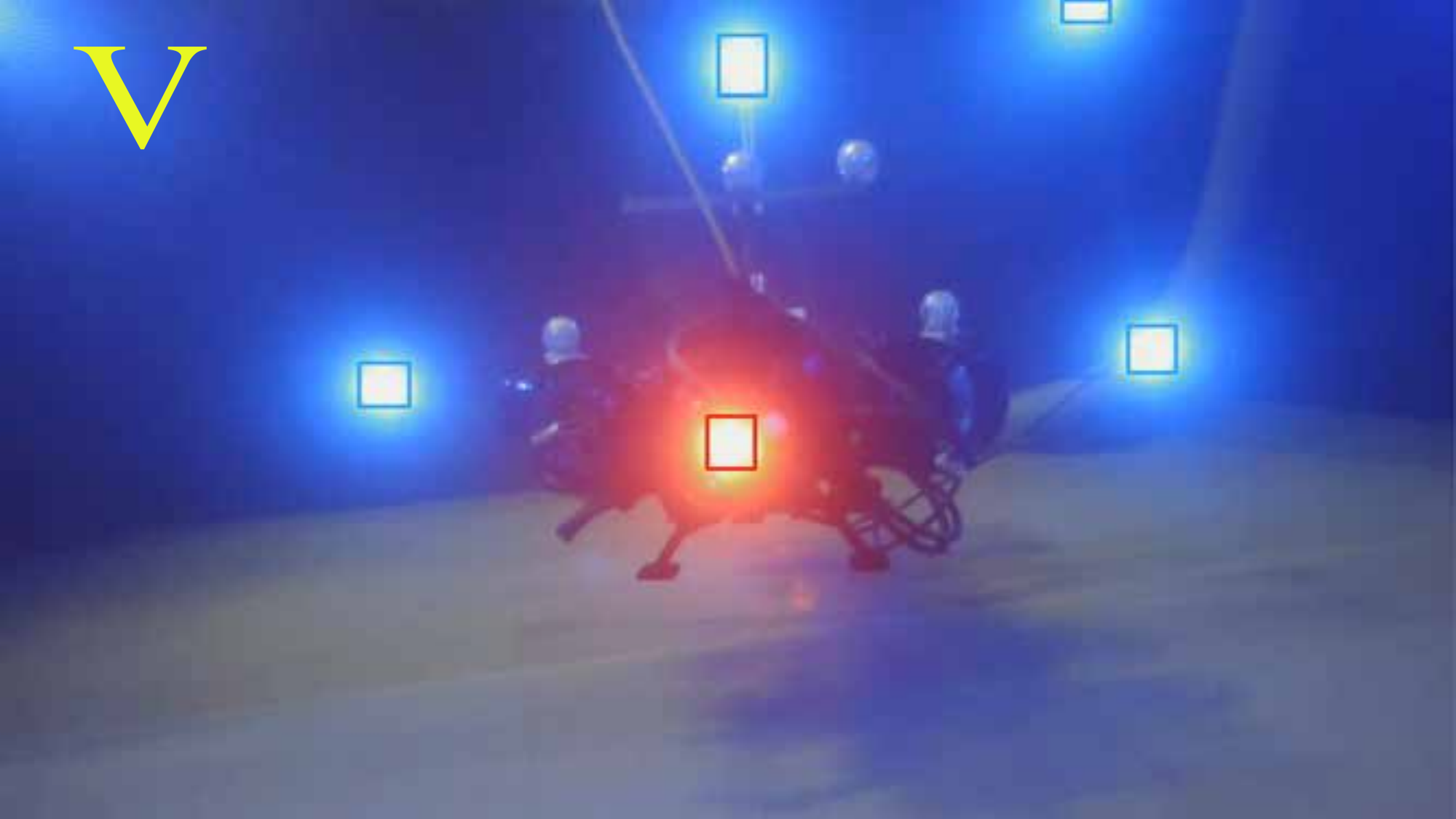}
        \end{minipage}
    }

    \vspace{-0.6em}

    \subfloat[Pose-DP: T4 Challenging\label{fig:visibility_posedp_t4}]{
        \begin{minipage}{0.98\textwidth}
            \centering
            \includegraphics[width=0.192\linewidth]{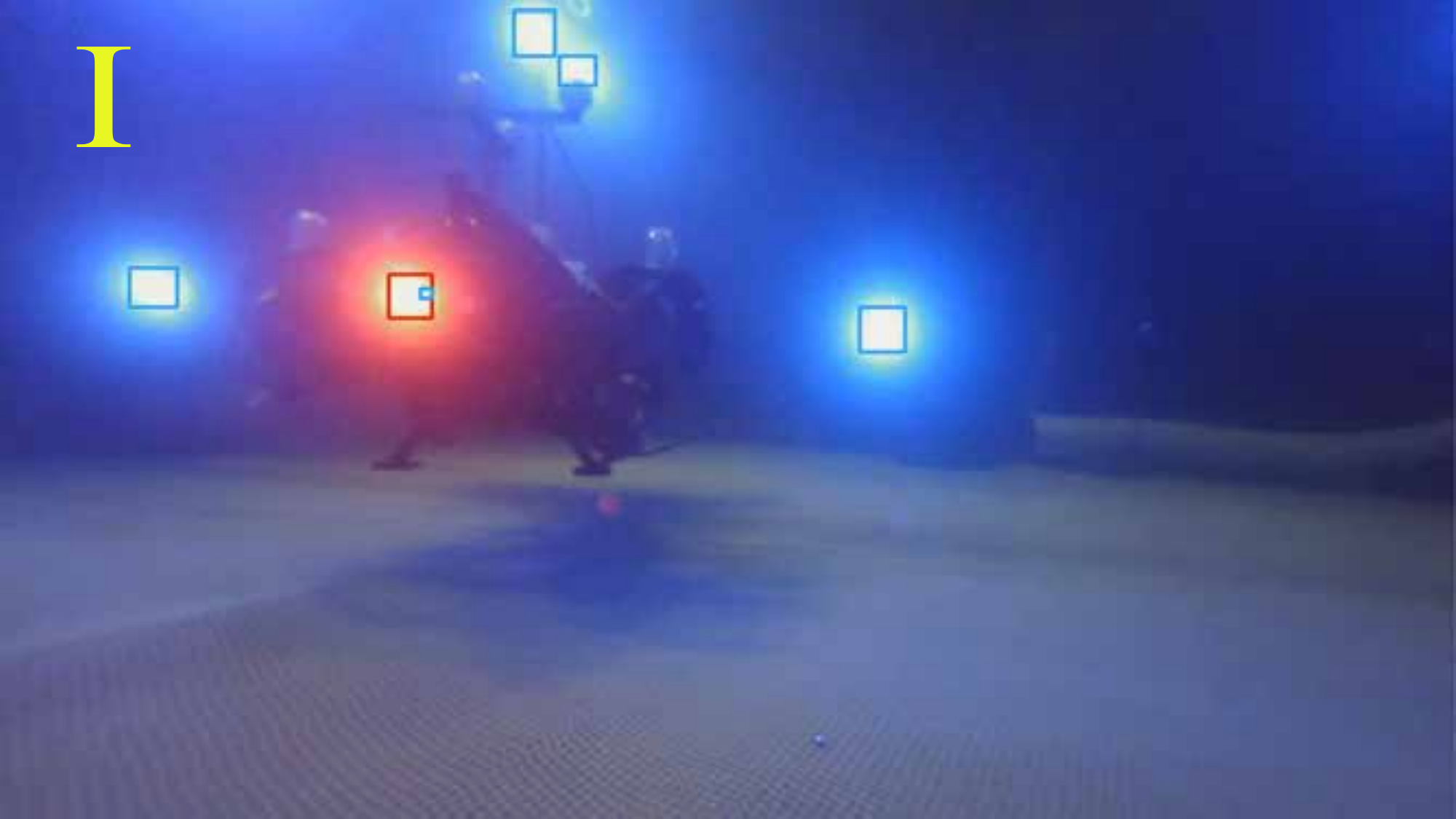}
            \hfill
            \includegraphics[width=0.192\linewidth]{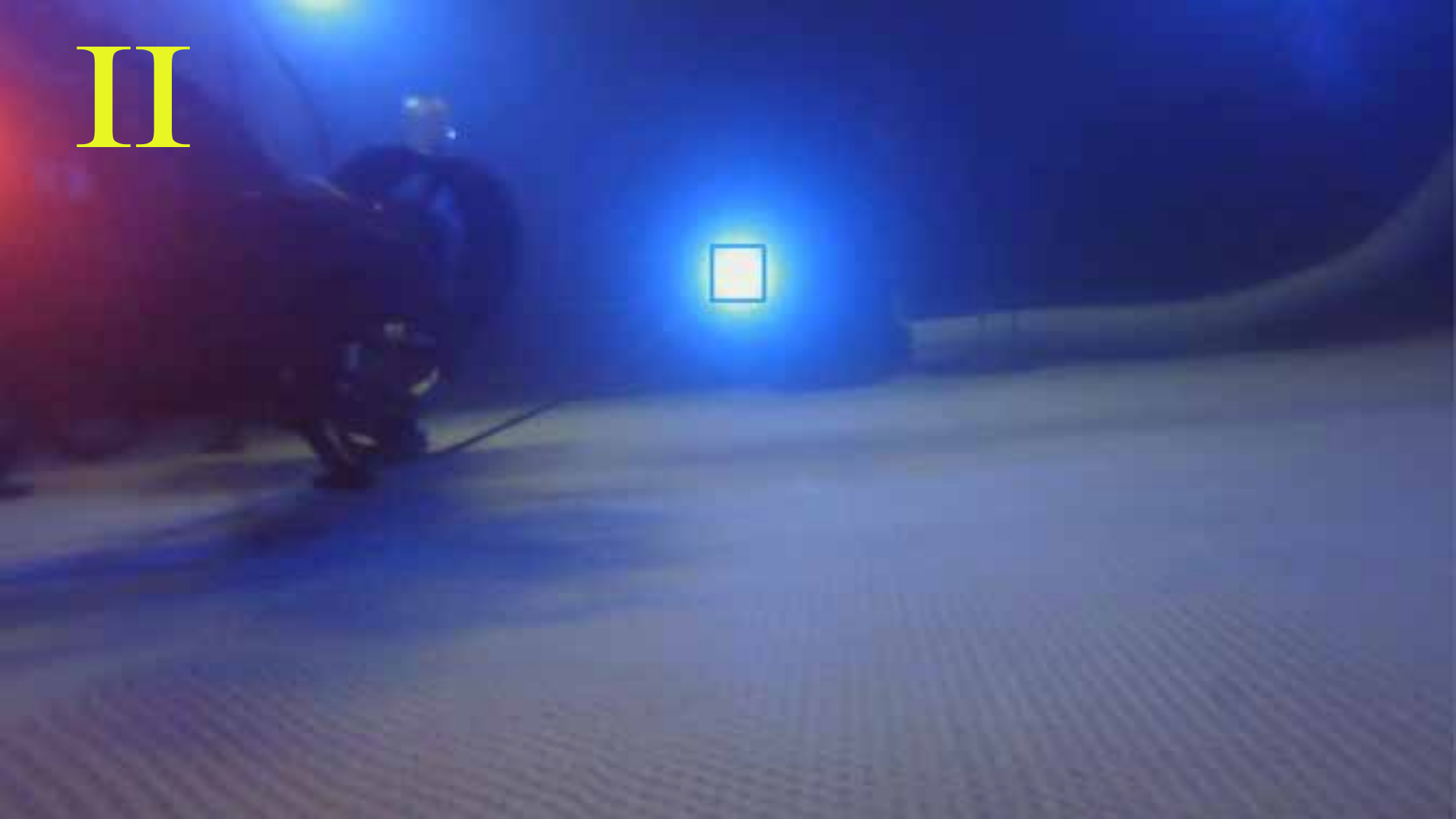}
            \hfill
            \includegraphics[width=0.192\linewidth]{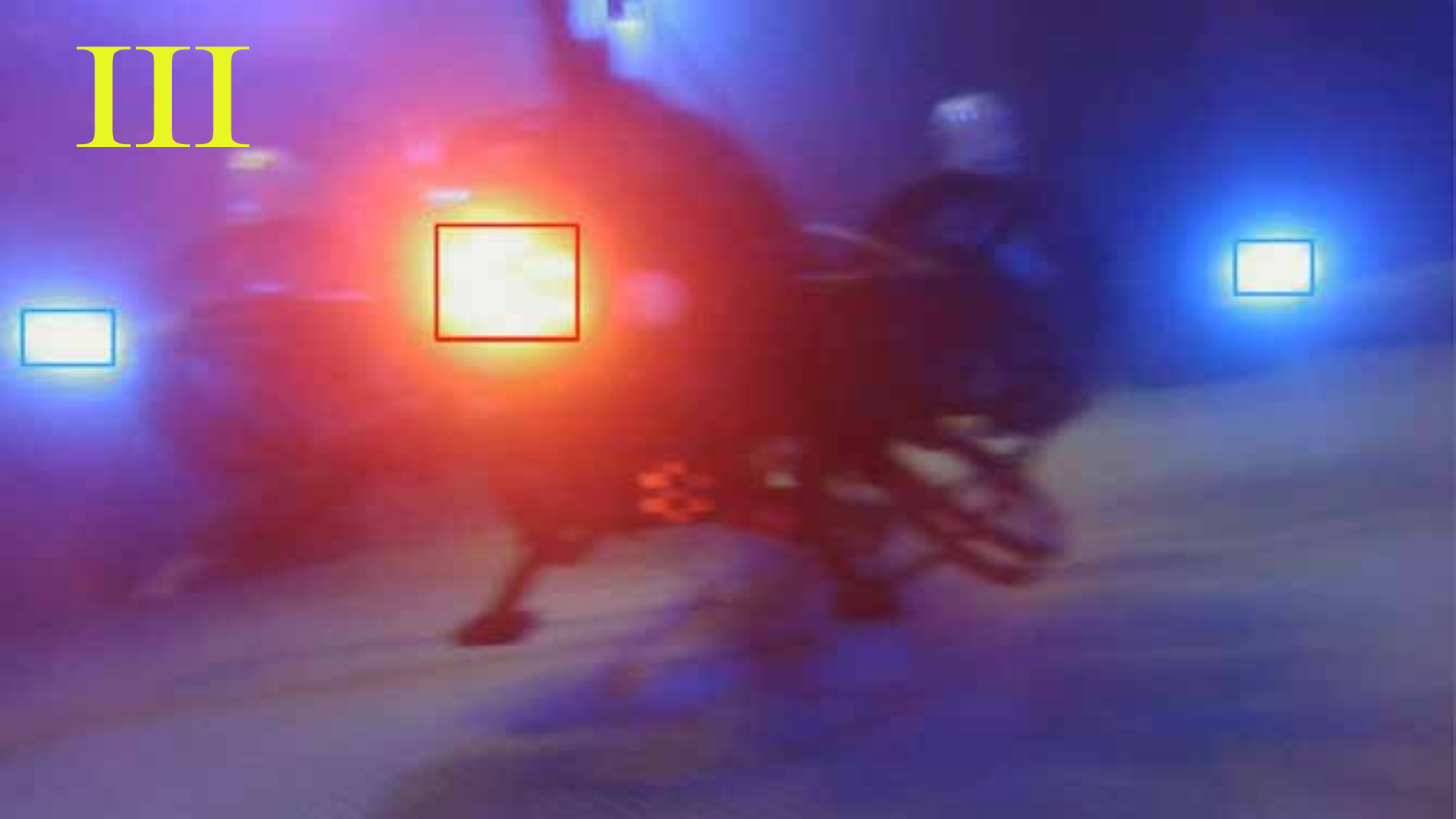}
            \hfill
            \includegraphics[width=0.192\linewidth]{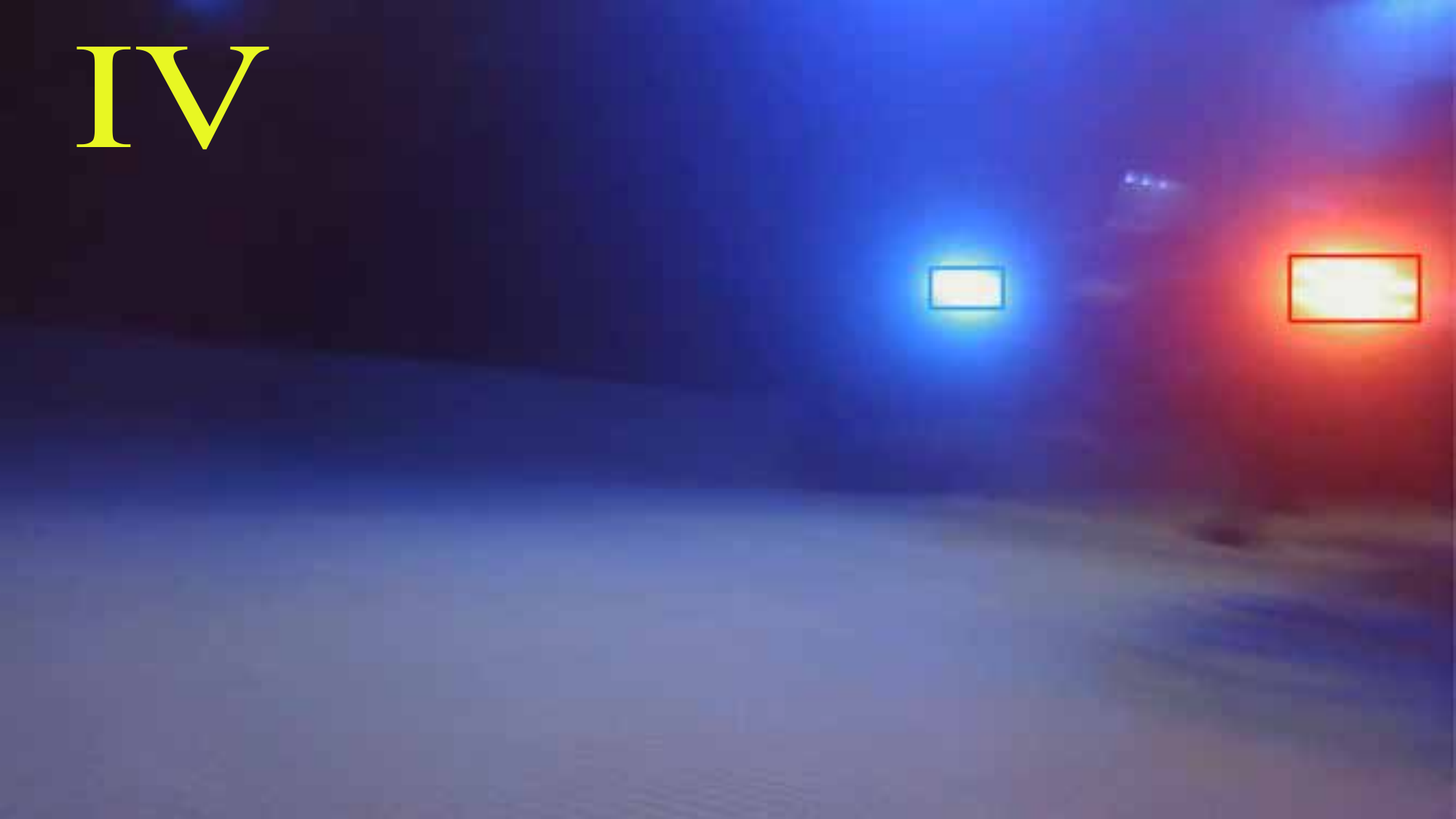}
            \hfill
            \includegraphics[width=0.192\linewidth]{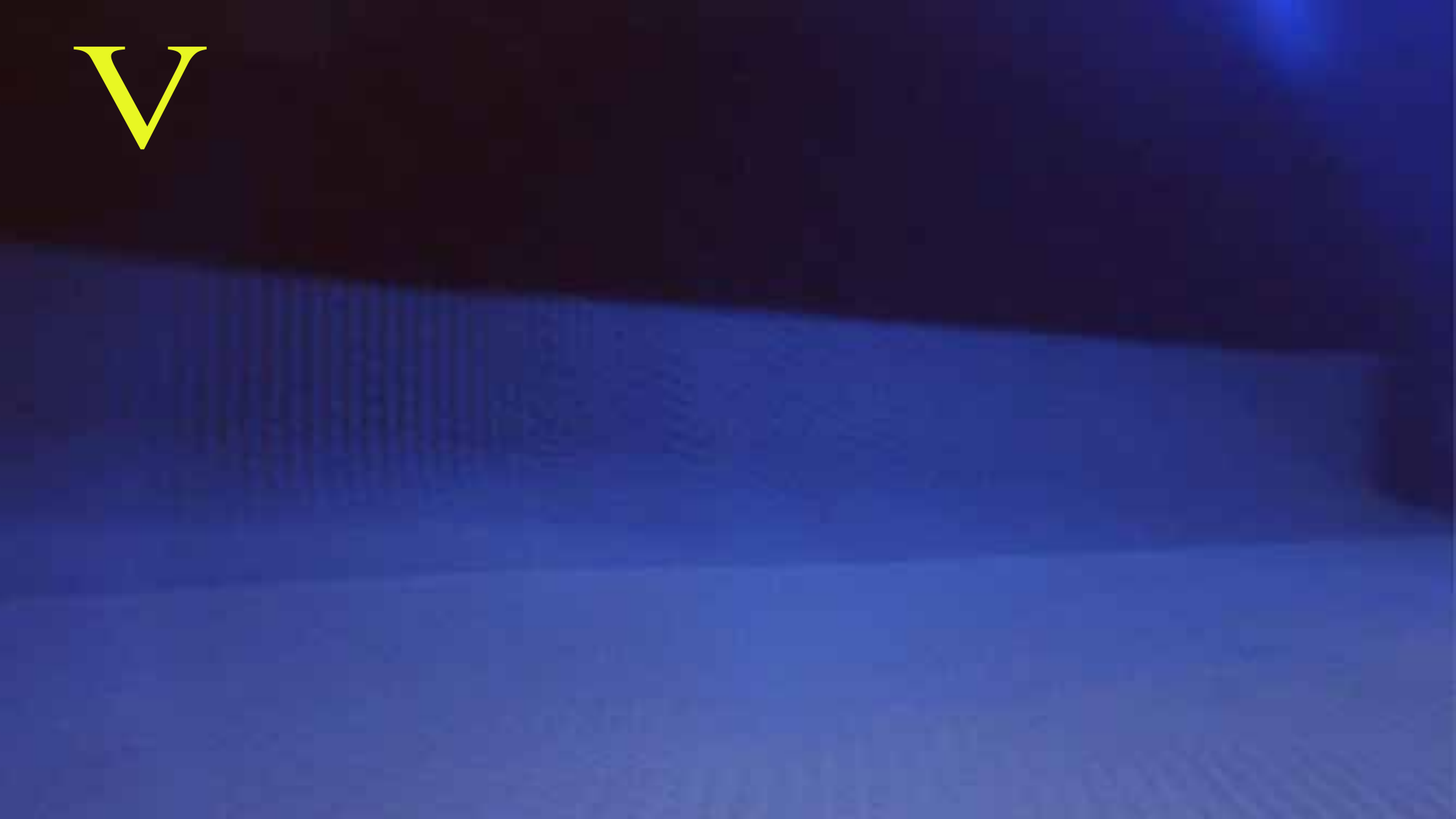}
        \end{minipage}
    }

    \vspace{-0.6em}

    \subfloat[FATHOM: T4 Challenging\label{fig:visibility_fathom_t4}]{
        \begin{minipage}{0.98\textwidth}
            \centering
            \includegraphics[width=0.192\linewidth]{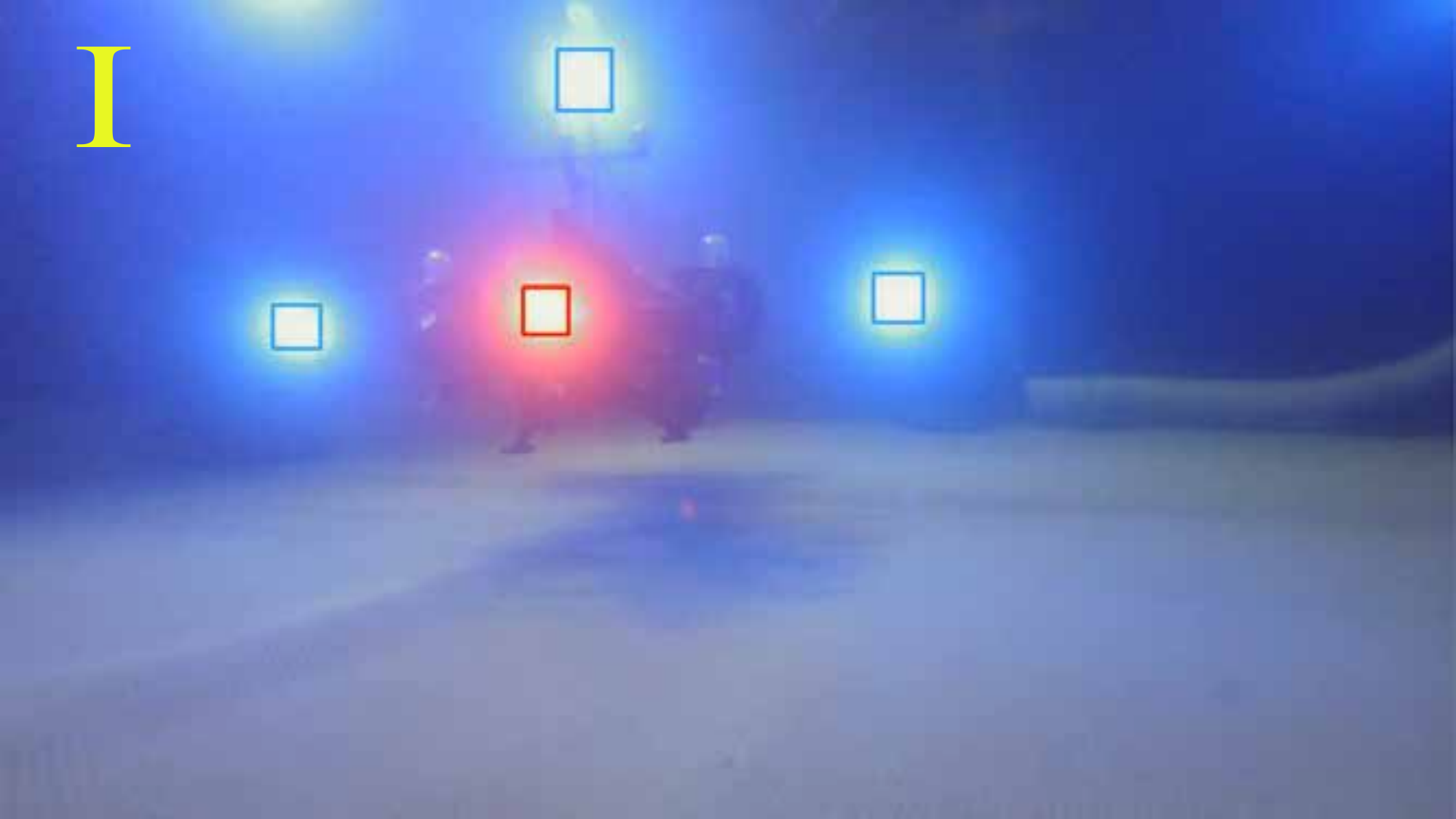}
            \hfill
            \includegraphics[width=0.192\linewidth]{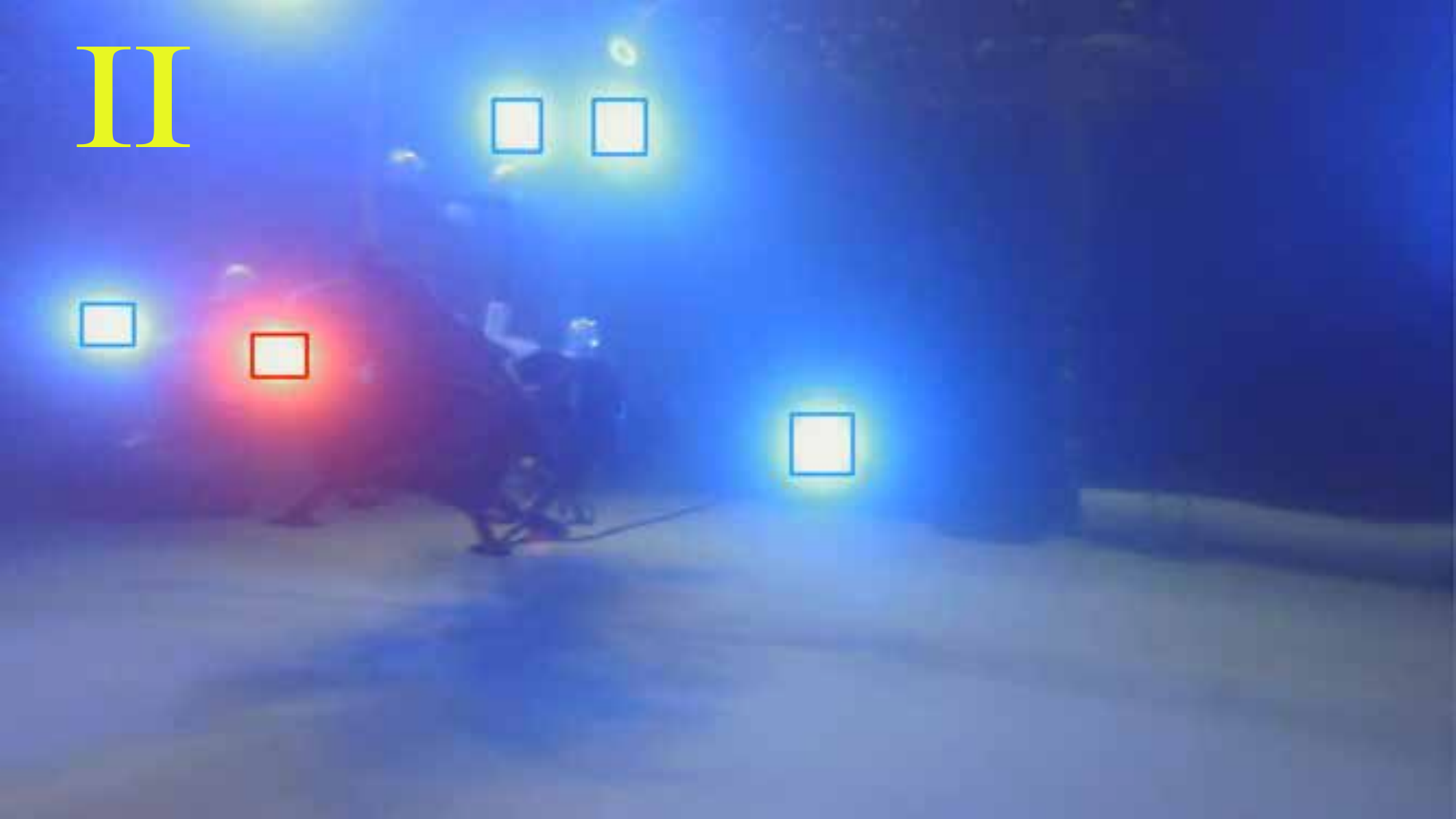}
            \hfill
            \includegraphics[width=0.192\linewidth]{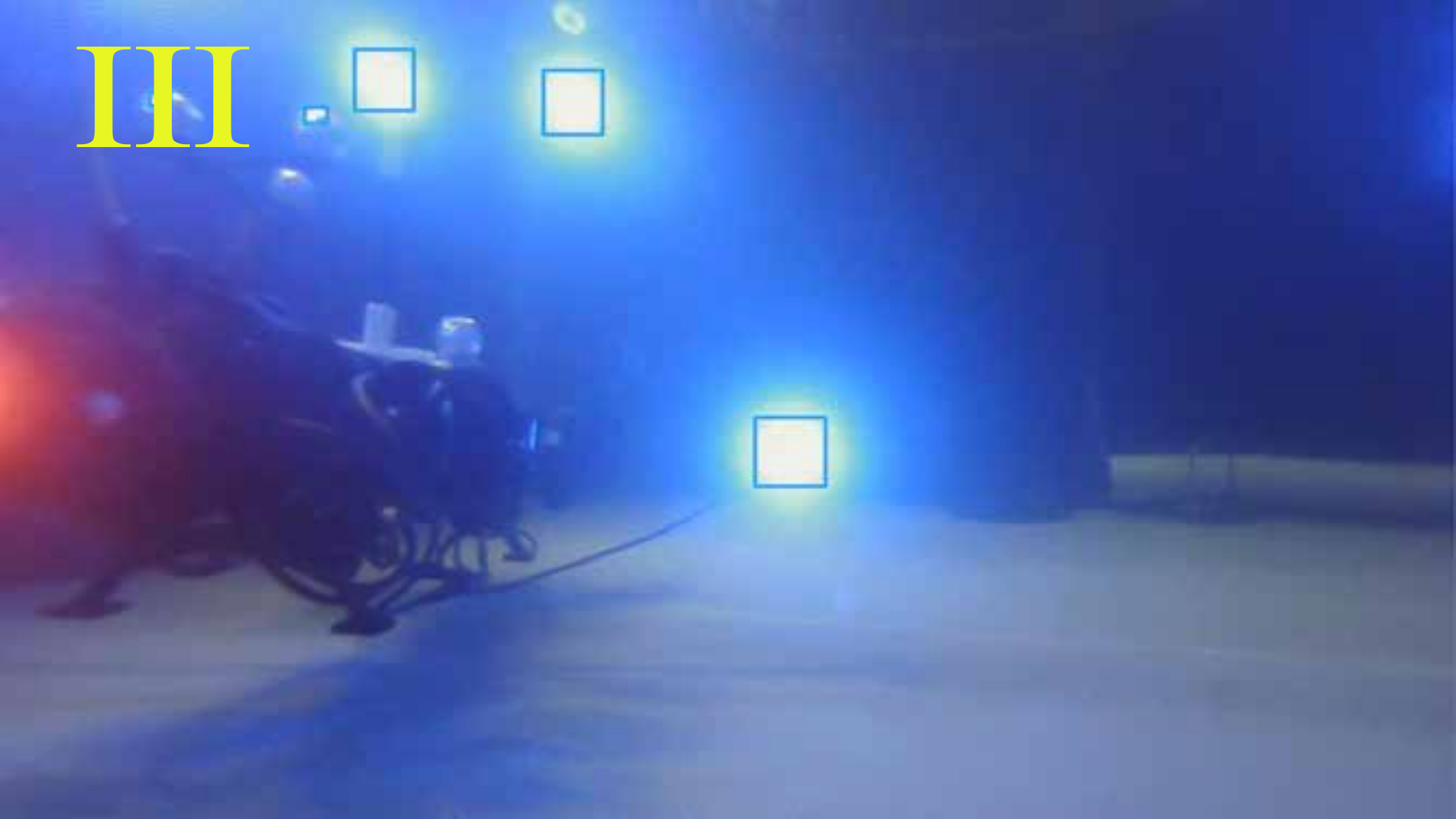}
            \hfill
            \includegraphics[width=0.192\linewidth]{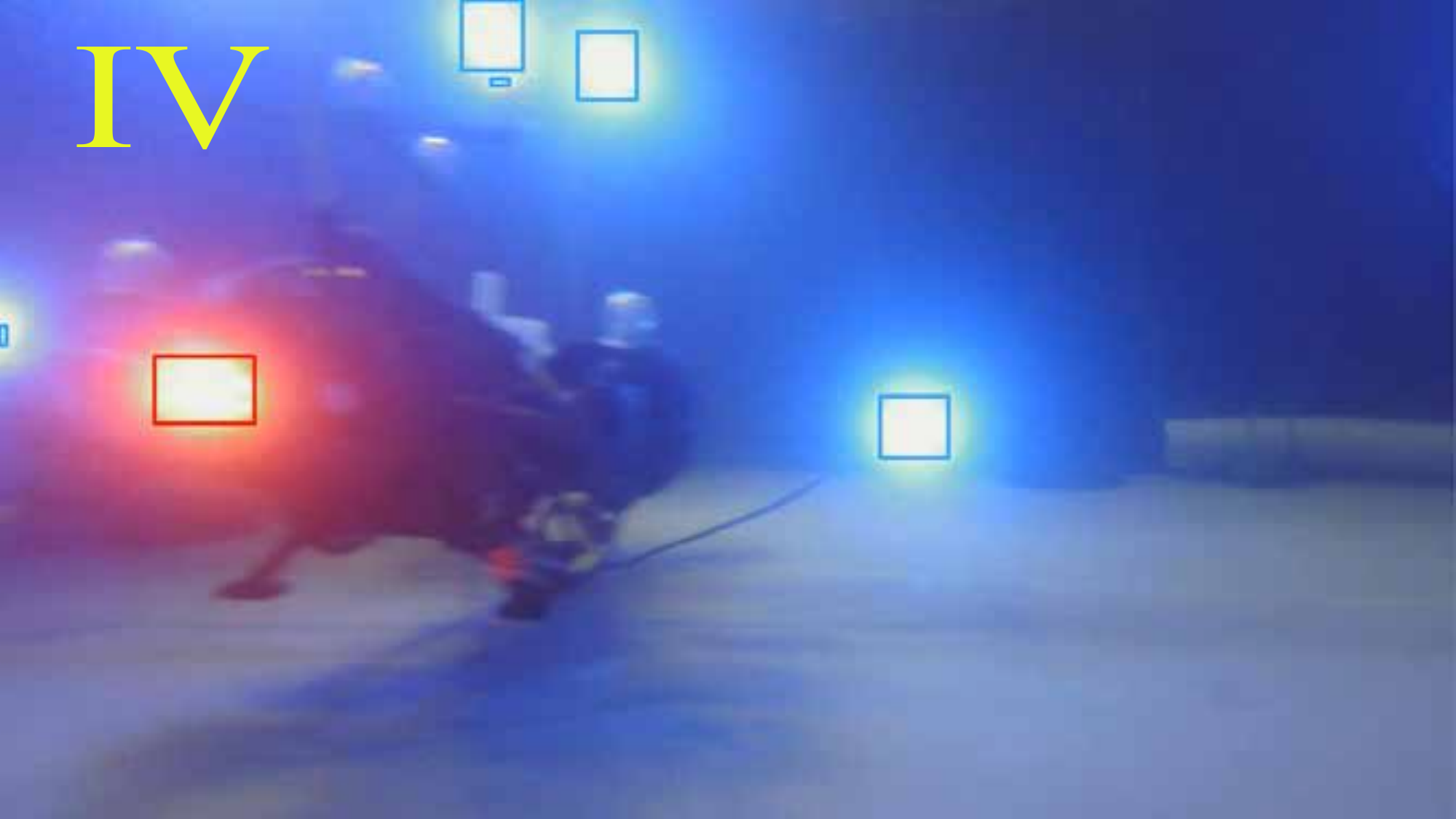}
            \hfill
            \includegraphics[width=0.192\linewidth]{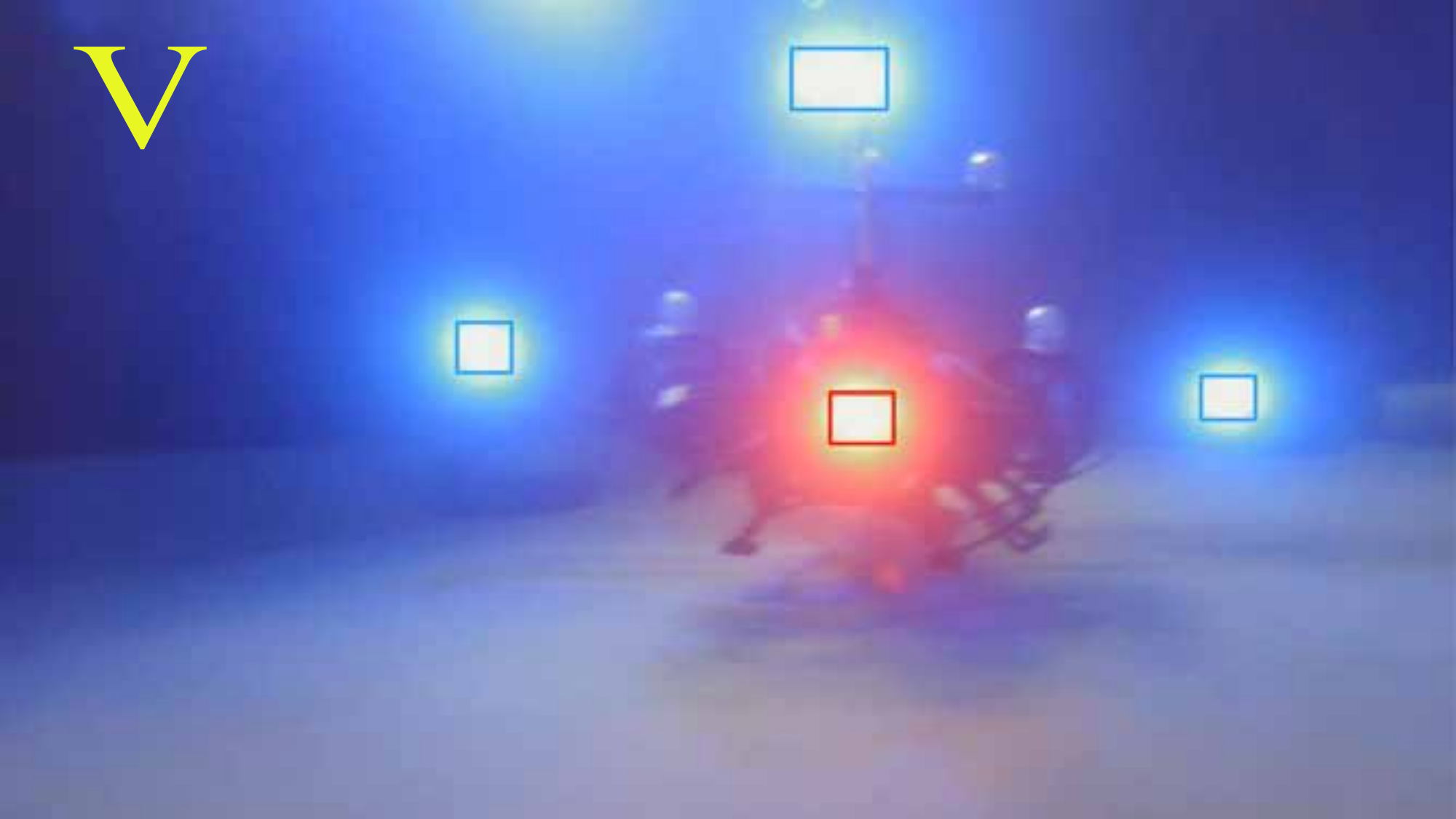}
        \end{minipage}
    }

    \caption{First-person observations during challenging T3 and T4 maneuvers. (a) Pose-PID loses the target in T3, whereas (b) FATHOM restores marker visibility. (c) Pose-DP loses the target in T4, whereas (d) FATHOM reacquires the sparse marker pattern and completes the maneuver.}
    \label{fig:partial_visibility_examples}
\end{figure*}

Fig.~\ref{fig:partial_visibility_examples} shows onboard observations during tracking loss and recovery. In T3, the red marker moves toward the left image boundary and disappears under Pose-PID, with bright background lights still present. FATHOM instead restores the displaced marker pattern toward the image center and continues tracking. In T4, Pose-DP loses the target beyond the right image boundary following large position and scale changes. FATHOM reacquires partially lost markers near the left image boundary, restores a more centered observation, and completes the maneuver. These sequences connect degraded visual evidence and subsequent recovery to actual closed-loop outcomes, rather than merely reporting marker counts.

FATHOM does not directly correct the LED detector. Instead, its recurrent visual--inertial belief retains temporal information through incomplete observations, while DAP and WMPS generate and select coordinated actions using predicted tracking costs. Regulating relative motion can restore favorable camera geometry and marker visibility without correcting the detector itself. This illustrates the perception--control interaction, as improved tracking also provides more informative subsequent observations. The reduced partial-visibility error and recovery time support the complementary roles of belief maintenance and predictive selection.

\begin{table}[t]
\centering
\caption{Robustness under natural partial marker visibility ($N_{\mathrm{vis}}\leq3$). NRMSE and P95 Error use pooled partial-visibility frames, whereas Recovery Rate and mean Recovery Time use annotated events, with time measured over successful recoveries.}
\label{tab:partial_visibility}

\renewcommand{\arraystretch}{1.12}
\setlength{\tabcolsep}{2.0pt}
\footnotesize

\begin{tabular*}{0.98\columnwidth}{
@{\extracolsep{\fill}}
lcccc
@{}
}
\hline

Method
& \multicolumn{1}{c}{Partial-Vis.}
& \multicolumn{1}{c}{P95}
& \multicolumn{1}{c}{Recovery}
& \multicolumn{1}{c}{Recovery}
\\

& NRMSE
& Error
& Rate (\%)
& Time (s)
\\

\hline

Pose-PID
& 0.333
& 0.699
& 85.7
& 3.10
\\

Pose-ISMC
& 0.250
& 0.416
& 88.5
& 2.97
\\

Pose-DP
& 0.385
& 0.661
& 81.2
& 3.46
\\

DAP-only
& 0.206
& 0.309
& 100.0
& 2.65
\\

\textbf{FATHOM}
& \textbf{0.163}
& \textbf{0.246}
& \textbf{100.0}
& \textbf{2.17}
\\

\hline
\end{tabular*}
\end{table}

\subsection{Real-Time Deployment Performance}
\label{subsec:realtime}

FATHOM runs onboard an NVIDIA Jetson AGX Orin (64 GB) at a nominal 10-Hz control rate with $K=16$ and a 10-step horizon. Table~\ref{tab:runtime} reports an inference-and-selection latency of $73.83\pm7.70$~ms on average and $85.10$~ms at the 95th percentile. DAP sampling dominates computation ($56.69$~ms), followed by predictive selection ($12.04$~ms) and belief updating ($4.09$~ms). The measured pipeline fits within the nominal 100-ms control period at the 95th percentile.

\begin{table}[t]
\centering
\caption{Onboard FATHOM computation time (ms) for belief updating, diffusion sampling, and predictive selection.}
\label{tab:runtime}

\renewcommand{\arraystretch}{1.12}
\setlength{\tabcolsep}{1.2pt}
\footnotesize

\begin{tabular*}{0.98\columnwidth}{
@{\extracolsep{\fill}}
lcccc
@{}
}
\hline

Statistic
& \multicolumn{1}{c}{Belief}
& \multicolumn{1}{c}{DAP}
& \multicolumn{1}{c}{Predictive}
& \multicolumn{1}{c}{Total}
\\

& Update
& Sampling
& Selection
&
\\

\hline

Mean $\pm$ Std. (ms)
& 4.09 $\pm$ 0.89
& 56.69 $\pm$ 6.22
& 12.04 $\pm$ 2.00
& \textbf{73.83 $\pm$ 7.70}
\\

P95 (ms)
& 5.83
& 66.44
& 15.28
& \textbf{85.10}
\\

\hline
\end{tabular*}
\end{table}

\subsection{Summary of Experimental Findings}
\label{subsec:experiment_summary}

The real-AUV experiments validate the complementary roles of predictive belief and world-model-guided action selection. Across 24 trials per controller, FATHOM achieves an overall NRMSE of $0.124$ and completes all 12 challenging trials. Relative to DAP-only, WMPS reduces challenging-motion error by $14.6\%$, while the error distributions show improved consistency and fewer large deviations. Under natural partial marker visibility, FATHOM achieves an NRMSE of $0.163$ and recovers in all annotated events, with a mean recovery time of $2.17$~s. The onboard inference-and-selection pipeline averages $73.83$~ms, supporting nominal 10-Hz operation. Together, the results show that temporal belief inference and predictive action selection enable reliable tracking without external state feedback.

\section{Conclusion}

This paper proposed FATHOM, a world-model-guided diffusion framework for sparse-marker AUV tracking under partial observations. SSME and MT-RSSM integrate sparse visual observations, inertial measurements, and action history into a predictive tracking belief. Conditioned on this belief, DAP generates expert-like thruster trajectories, while WMPS predicts their tracking consequences and selects the most favorable candidate. Offline evaluations validated the learned predictive representation and action selection. Real-AUV experiments demonstrated superior tracking performance over pose-based and diffusion-only baselines, achieving an overall NRMSE of $0.124$ and completing all challenging trials. FATHOM also maintained robust tracking under natural partial marker visibility and supported real-time onboard operation without motion-capture feedback.

Future work will investigate disturbance-aware prediction and control to improve robustness against hydrodynamic disturbances and extend FATHOM to open-water environments with more variable sensing and dynamics.

\bibliographystyle{IEEEtran}

\bibliography{IEEEabrv,paper}

\raggedbottom


\end{document}